\documentclass[review]{elsarticle}
\graphicspath{ {./figures/} }
\usepackage{hyperref}
\usepackage{float}
\usepackage{verbatim} 
\usepackage{apalike}
\usepackage[raggedrightboxes]{ragged2e}
\usepackage{svg}
\usepackage{svg-extract}
\usepackage{array}
\usepackage{setspace}
\usepackage{graphicx}
\usepackage{caption}
\usepackage{longtable}
\restylefloat{figure}
\restylefloat{table}
\usepackage{colortbl}
\usepackage{amsmath}
\usepackage{caption}

\journal{Expert Systems with Applications}
\biboptions{authoryear}

\begin{document}
\begin{frontmatter}
\begin{titlepage}
\begin{center}
\vspace*{1cm}

\textbf{A Review of Physics-Informed Machine Learning Methods with Applications to Condition Monitoring and Anomaly Detection}

\vspace{1.5cm}
Yuandi Wu$^a$ (wuy187@mcmaster.ca), Brett Sicard$^a$ (sicardb@mcmaster.ca), Stephen Andrew Gadsden$^a$ (gadsden@mcmaster.ca)\\
\hspace{10pt}

\begin{flushleft}
\small  
$^a$ Intelligent and Cognitive Engineering Laboratory, McMaster University, Hamilton, Ontario, Canada \\

\vspace{1cm}
\textbf{Corresponding Author:} \\
S. Andrew Gadsden \\
1280 Main Street West, Hamilton, ON L8S 4L8, Canada \\
Tel: (905) 525-9140 \\
Email: gadsden@mcmaster.ca

\end{flushleft}        
\end{center}
\end{titlepage}

\thispagestyle{empty}

\listoftables
\vspace{5pt}
\listoffigures

\newpage
\pagenumbering{arabic}

\title{A Review of Physics-Informed Machine Learning Methods with Applications to Condition Monitoring and Anomaly Detection}

\author[label1]{Yuandi Wu}
\ead{wuy187@mcmaster.ca}

\author[label1]{Brett Sicard}
\ead{sicardb@mcmaster.ca}

\author[label1]{Stephen Andrew Gadsden*}
\ead{gadsden@mcmaster.ca}

\cortext[cor1]{Corresponding author.}
\address[label1]{McMaster University, 1280 Main Street West, Hamilton, ON L8S 4L8, Canada }
\begin{singlespacing}
     
\section*{Highlights}
    \begin{enumerate}
        \item Surveyed recently published literature on physics-informed machine learning.
        \item Establishes a bridge between physics-based modeling and data-driven approaches.
        \item Emphasizes the importance of integrating domain knowledge into the PIML framework.
        \item Showcases real-world case studies and industrial applications.
        \item Provides a foundational survey for future research work in the field.
    \end{enumerate}
 
\begin{abstract}
Condition monitoring plays a vital role in ensuring the reliability and optimal performance of various engineering systems. Traditional methods for condition monitoring rely on physics-based models and statistical analysis techniques. However, these approaches often face challenges in dealing with complex systems and the limited availability of accurate physical models. In recent years, physics-informed machine learning (PIML) has emerged as a promising approach for condition monitoring, combining the strengths of physics-based modeling and data-driven machine learning. This study presents a comprehensive overview of PIML techniques in the context of condition monitoring. The central concept driving PIML is the incorporation of known physical laws and constraints into machine learning algorithms, enabling them to learn from available data while remaining consistent with physical principles. Through fusing domain knowledge with data-driven learning, PIML methods offer enhanced accuracy and interpretability in comparison to purely data-driven approaches. In this comprehensive survey, detailed examinations are performed with regard to the methodology by which known physical principles are integrated within machine learning frameworks, as well as their suitability for specific tasks within condition monitoring. Incorporation of physical knowledge into the ML model may be realized in a variety of methods, with each having its unique advantages and drawbacks. The distinct advantages and limitations of each methodology for the integration of physics within data-driven models are detailed, considering factors such as computational efficiency, model interpretability, and generalizability to different systems in condition monitoring and fault detection. Several case studies and works of literature utilizing this emerging concept are presented to demonstrate the efficacy of PIML in condition monitoring applications. From the literature reviewed, the versatility and potential of PIML in condition monitoring may be demonstrated. Novel PIML methods offer an innovative solution for addressing the complexities of condition monitoring and associated challenges. This comprehensive survey helps form the foundation for future work in the field. As the technology continues to advance, PIML is expected to play a crucial role in enhancing maintenance strategies, system reliability, and overall operational efficiency in engineering systems.

\end{abstract}

\begin{keyword}
Machine learning \sep deep learning \sep physics-informed machine learning \sep condition monitoring
\sep anomaly detection
\end{keyword}
 \end{singlespacing}
\end{frontmatter}

\begin{singlespacing}
\section{Introduction}
\label{introduction}

Throughout the last decade, Machine learning (ML) algorithms have witnessed rapid development in a variety of industries for their efficacy and ability to extrapolate patterns from data. Purely through available data, ML models are capable of accurately representing the relation between a given set of inputs and outputs with minimal human interference. This property made ML models ideal for the representation of complex systems in which the relation and parameters governing behavior are not easily obtained. Despite their many advantages, however, ML models are not without their drawbacks.\\

In general, ML algorithms are a data-driven process that seeks to derive the relationship between a given input and its respective output. This process is generally performed in accordance with some defined optimization algorithm, in which predictions made by the model are evaluated and continuously adjusted to better represent the data given. As can be expected, the performance of ML models are heavily reliant on the data by which they are optimized upon. Indeed, restrictions to data quality and availability are amongst the main concerns when choosing to work with ML \citep{l2017machine}. For many engineering applications, the collection of sufficient quantities of data to build a reliable model may be challenging, costly, and/or not feasible due to time and resource constraints. A considerable amount of clean, representative, and non-sparse data is required to properly formulate the model \citep{l2017machine}. Insufficient quantities and/or non-representative data often lead to a skewed representation of system behavior that is inconsistent with the true underlying physical relationship, ultimately resulting in misleading conclusions. Furthermore, ML models are considered to be "black box" models, in which intermediary information between input and output is not relevant nor required in producing a correlation between some input and output. That is to say, the underlying mechanism of a system is often not considered in the development of these models, and while effective in representing a system, may not further contribute to our understanding of said system \citep{rudin2019stop}. \\

With respect to the representation of systems based on prior knowledge, physics-based modeling has also been traditionally employed. However, models developed purely on the understanding of the system see limited use in modeling real-world systems, due to the many challenges to its applicability. First and foremost, physical models are often computationally expensive to model \cite{jia2019physics}. Due to the computational complexity of most real-world physical systems, and the variety of governing equations involved for each specific physical agent or phenomenon, the cost required to fully model said systems is considerable. Furthermore, physical models often represent an imperfect interpretation of the system, due to missing or incomplete understanding of the system.\\

Naturally, researchers have come to the realization that the combination of physical and data-driven models was the next step in the prediction and modeling of system behavior. This paradigm of PIML was initially conceptualized by \citep{karpatne2017theory} formally in their study of theory-guided data science, in which they outlined various avenues of integration between domain knowledge and data-driven solutions. Through this unification, new physics-informed models are capable of benefiting from both physics-based and data-driven methods concurrently. Since their publication, a plethora of studies regarding the PIML paradigm has been conducted. Various authors, most notably \cite{raissi2019physics}, further advanced the integration between theory and data science with the introduction of Physics-Informed Neural Networks (PINNs), whereby physical laws in the form of governing equations are encoded within neural networks. The neural network architecture and properties made it especially suitable in their use case, for approximating the solutions of Partial Differential Equations (PDEs). \cite{raissi2019physics} made use of the neural network architecture in their demonstration of a systematic methodology for solving non-linear partial differential equations. \cite{karniadakis2021physics} reviewed popular methodologies by which the integration of physics and data-driven techniques takes place, as well as presented their insights on limitations and potential applications of the technique. \cite{meng2022physics} also surveyed a variety of work in the area of PIML, and presented a summary of core motivations behind their development, popular physical governing equations employed in various applications, as well as methods of integration. From literature, it is evident that despite their novelty, applications of PIML have been prominent in a variety of fields. \\

For this survey, the applications of PIML methods within the context of condition monitoring in various engineering applications are examined. Condition monitoring is an essential aspect of the engineering industry as it plays a vital role in ensuring the reliability, safety, and efficiency of assets. Implementations of PIML in this area involve the continuous monitoring of various parameters such as vibration, temperature, pressure, and other critical factors that can indicate the health state of the asset monitored. Through continuous sampling of these parameters, engineers may identify potential problems before they occur, and take corrective actions to prevent costly and unplanned downtime, equipment failure, or even catastrophic accidents \cite{surucu2023condition}. Recent developments in PIML and information capabilities have led to a wide variety of innovative methodologies for the integration of physical knowledge for applications in condition monitoring. In the survey by \cite{xu2022review}, the authors have already outlined extensively, the specific applications of PIML with condition monitoring. As such, rather than focusing on the specific applications, this survey aims to provide readers with an overview of recent methodologies of integration between the integration of physics-based knowledge with machine learning methods. The overall objective of this paper is thus to provide readers with a foundation for comprehending its specific applications, and a deeper understanding of the underlying principles and mechanisms of PIML. 

As will be discussed in the body of this survey, PIML learning approaches offer distinct advantages over conventional machine learning techniques due to their ability to incorporate fundamental physical laws and principles into the learning process. PIML effectively combines the interpretive capabilities of machine learning algorithms with the foundational understanding of physics, leveraging prior knowledge to guide the learning process. Often, this learning process results in a more accurate and interpretable model. Furthermore, PIML methods benefits from reduced reliance on vast amounts of labeled training data, as physics-based guidelines for optimization can constrain the solution space and provide insights, even in data-scarce scenarios. In all, physics-informed methods enable better generalization, robustness, and interpretability, making them superior to conventional ML approaches in many scientific and engineering applications. Furthermore, these methods offer better explainability to the end user in the context of explainable artificial intelligence (xAI), which is a growing consideration for the wide-adoption of AI techniques.

The literature survey is organized as follows: Section \ref{review methodology} provides an outline of the search methodology in determining articles for review. Section \ref{PIML} provides a detailed explanation of the methodologies by which physics may be integrated into data-driven solutions. Furthermore, the section also details the background of popular architectures within the machine learning community, as well as how authors in various fields seek to incorporate prior physical knowledge within these models. Section \ref{Discussion} provides a summary and interpretations of recent trends, with a focus on discussion pertaining to the advantages and limitations of the methodologies surveyed. Finally, the survey is concluded and summarized in Section \ref{Conclusion}.

\section{Literature Review Methodology} \label{review methodology}
\begin{figure}[h!]
\begin{center}
\includegraphics[width=\textwidth]{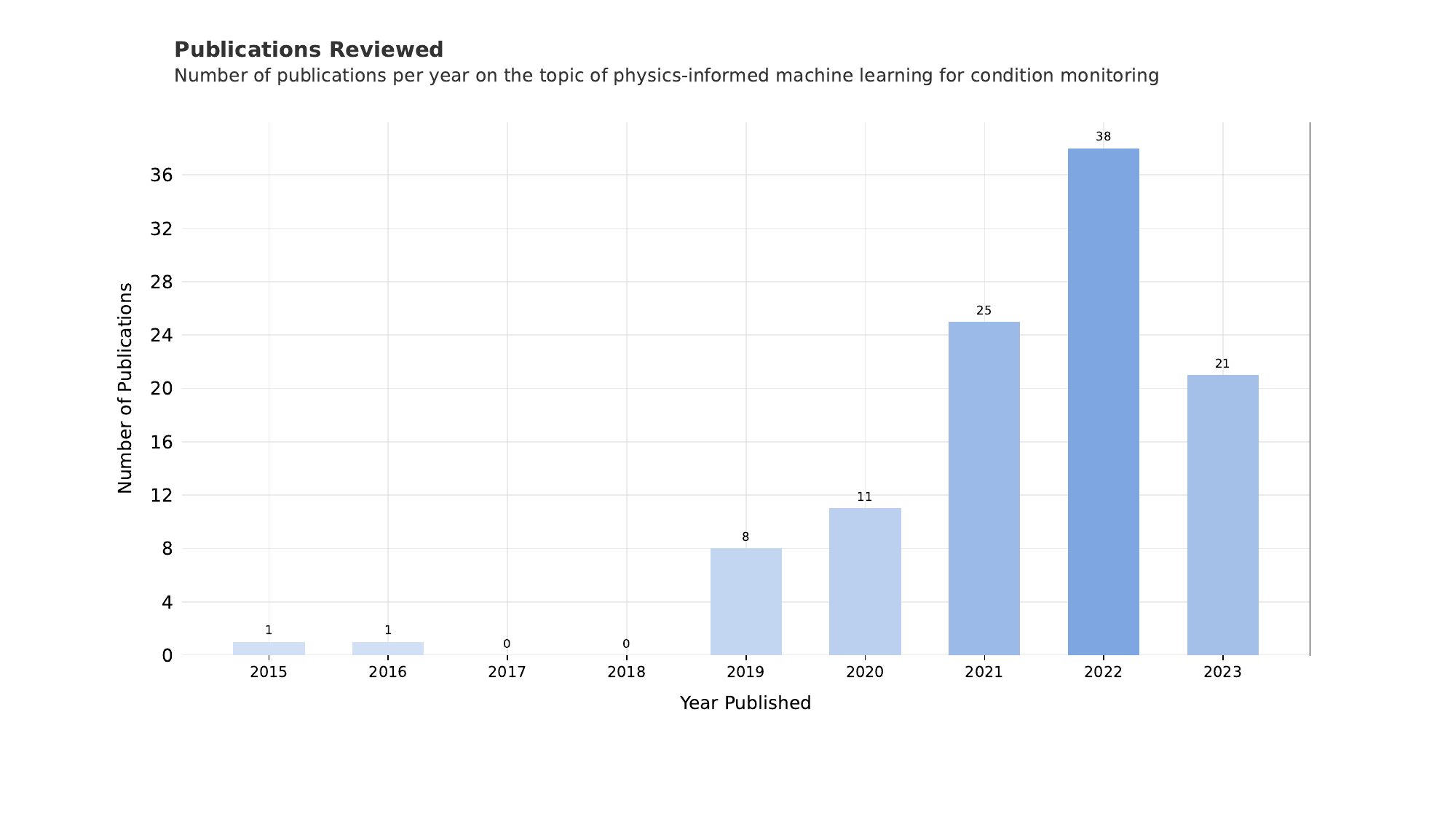}
\caption{Tallied number of literary works discussed in this review, with respect to their year of publication. Note: Literature works reviewed in 2023 were limited up until the time of writing of this survey (June 2023).}\label{tally}
\end{center}
\end{figure}

This survey reviewed recent developments for integration between physics-based modeling and ML with applications in condition monitoring and anomaly detection. A total of 105 papers published were selected after screening. Literary works covered are tallied with respect to their year of publication, and presented in the visual format in figure \ref{tally}. From the figure, it is evident that the concept of PIML has been rapidly gaining popularity within the research community. In this survey, search methodologies involve filter keywords such as “physics-informed”, “physics guided”, “physics-based”, “Machine learning”, “condition monitoring”, "fault detection", "anomaly detection", and et cetera. Searches were performed on platforms such as Google Scholar, IEEE Xplore, Science-Direct, and ACM Digital Library. Results were filtered based on relevancy, year, and citations.

\section{Physics-Informed Machine Learning} \label{PIML}
This section details the background of PIML models, as well as introduces several methods by which physical meaning may be embedded within data-driven solutions. Implementation of PIML varies greatly depending on the field of application, and a diverse set of implementation methodologies exists. In general, integration between physics-based modeling with ML is typically accomplished through the following frameworks and may be summarized:
\begin{enumerate}
    \item  Physics Embedded in Feature Space
    \item  Data-Enhanced Refinement of Physical Models
    \item  Physics-Informed Regularization
    \item  Physics-Guided Design of Architectures
\end{enumerate}
Details of these will be discussed in the following sections.

\subsection{Physics Embedded in Feature Space} \label{31}
Perhaps the most straightforward method of integration between physical principles with ML methods is the development of the feature space of an ML model with physical modeling. Augmentations or alterations to the feature space do not directly affect model architecture, and the resultant model is still considered to be a black-box model, that is, a model capable of producing relevant results without revealing information regarding the mechanisms by which the results are derived \cite{karniadakis2021physics}. By leveraging the fundamental understanding of the underlying physics, however, these methods shape the feature space of a machine learning algorithm in a manner consistent with the physical laws. This integration offers several advantages over traditional machine learning approaches and leads to a more robust, and data-efficient framework. Through this integration, ML algorithms may be designed to exploit prior knowledge of physical relationships to be more accurately and efficiently applied to a variety of engineering applications. As described by \cite{karniadakis2021physics}, this form of integration primarily concerns with the introduction of observational biases to enhance the performance of ML models. Here, observational biases refer to the specific measurements or features that embody the underlying physics or prior knowledge about the system under consideration. Through the incorporation of prior knowledge, the introduction of observational biases through various input augmentation procedures serves as a guide in constraining algorithm predictions to be physically plausible. Various studies have demonstrated that algorithms are more capable of identifying relevant features in comparison to purely data-driven methods, leading to improved modeling capabilities and reduced data limitations \citep{leturiondo2017validation,gitzel2021using,deng2022physics}.  Within the context of applications in condition monitoring, it is often critical to have engineered features within the ML model that are sensitive to changes in the condition of the asset and is capable of properly differentiating nominal operational conditions from fault conditions. Several approaches to this incorporation may be seen in literature. For example, physically generated parameters and variables may be employed as additional inputs within the feature space. The addition of physics-informed features may be done either directly in the form of an additional augmented dataset parsed through the ML pipeline, or indirectly through methods such as transfer learning, whereby the features from a physics-informed source domain are captured via the ML algorithm and re-purposed. Subsequent subsections will discuss these methods of feature manipulation, with examples.

\subsubsection{Physics-Guided Input Feature Augmentation}
The field of machine learning has experienced tremendous growth in recent years, and this growth has been fueled in part by the availability of large datasets for the expressive and representative training of ML models \citep{l2017machine}. However, in the context of complex engineering tasks, collecting and labeling large quantities of data may be expensive, time-consuming, and in some cases, impractical or impossible. Moreover, due to the black-box nature of ML models, it is difficult to adjust the behaviors of the ML model purely from adjustments to datasets, even when information regarding the system is known beforehand.\\

A prevailing solution in literature has been to use synthetically generated features from system models to supplement or replace real-world data, with the main advantage being that it allows for the creation of large datasets with a high degree of variability, while simultaneously adhering to governing physical principles. This property is valued in many such engineering applications, where small quantities of observational data available may not accurately reflect the full range of operating conditions of a system or piece of equipment \citep{hopwood2022physics, gardner2021overcoming}. For example, observational data regarding specific fault conditions are rare and impractical to curate in many such applications. Furthermore, due to the rarity and impracticality of inducing specific system faults, available datasets are often imbalanced and severely skewed \citep{hopwood2022physics}. This poses significant issues for available ML algorithms and their performance, as standard classifiers tend to overly focus on larger classes. As such, the synthesis of physically relevant features or data represents an effective methodology for obtaining clean, balanced datasets in these scenarios.\\

Limitations that this approach encounters are often with respect to the accurate replication of the complexity of real-world operating conditions, and the risk that the generated data will not accurately reflect the behavior of the equipment or system in question due to incomplete or false prior physics knowledge \cite{serre2019deep}. Despite this, many authors have nevertheless elected to resolve this issue through the generation of physically consistent synthetic features or data through known physics regarding the system. In this fashion, the generative model forms or supplements existing feature space with tailored observational biases. The overall objective of this integration is the detection of potential issues with a higher degree of accuracy, with lesser requirements with respect to real-world data collection, and improved overall adherence to the expected behavior of the system with respect to physical principles. A summary of recent works implementing this framework is provided in Table \ref{table:1}.

\begin{singlespacing}
\begin{tiny}
\begin{longtable}[h!]{p{0.25\textwidth}  p{0.11\textwidth} p{0.25\textwidth} p{0.25\textwidth}} \hline \hline \\
    \\
    \textbf{Article Title} & \textbf{Citation} & \textbf{Description} & \textbf{Application} \\ 
    \hline \hline \\[0.05cm]
    \endhead

    \hline
    \endfoot

    \caption{Literature compiled for feature formation or augmentation using physics-based or physics-informed means, for use in machine learning algorithms.}
    \label{table:1}
    \endlastfoot
    
    \RaggedRight{Motor bearing fault detection using spectral kurtosis-based feature extraction coupled with k-nearest neighbor distance analysis} & \centering{\cite{tian2015motor}} & \RaggedRight{Feature engineering with spectral kurtosis, with classification using the k-nearest neighbor algorithm} & \RaggedRight{Machinery fault diagnosis with bearings} \\ \hline \\
    
    \RaggedRight{Hybrid Model-Based and Data-Driven Fault Detection and Diagnostics for Commercial Buildings} & \centering{\cite{frank2016hybrid}} & \RaggedRight{Feature engineering using first principles and empirical analysis, classification with variety of classical machine learning algorithms} &  \RaggedRight{Anomalous behavior detection in building energy consumption} \\ \hline \\ 

    \RaggedRight{Physics-guided logistic classification for tool life modeling and process parameter optimization in machining} & \cite{karandikar2021physics} & \RaggedRight{Taylor tool life relation with cutting speed applied to form input feature space through logarithmic transformations, in conjunction with a linear logistic classifier} & \RaggedRight{Remaining useful life estimation and state of health monitoring for machine tools} \\ \hline \\ 

    \RaggedRight{A physics-informed machine learning approach for notch fatigue evaluation of alloys used in aerospace} & \centering{\cite{hao2023physics}} & \RaggedRight{Physic-driven parameters for augmenting input feature space, regression using Support Vector Regression, Random Forest, and XGBoost} & \RaggedRight{Fatigue life estimations in poly-crystalline alloys} \\ \hline \\ 

    \RaggedRight{Structural Health Monitoring using deep learning with optimal finite element model generated data} & \centering{\cite{seventekidis2020structural}} & \RaggedRight{Finite element simulation generated structural data, using classification with convolutional neural networks} & \RaggedRight{Structural health monitoring} \\ \hline \\

    \RaggedRight{A hybrid physics-assisted machine-learning-based damage detection using Lamb wave} & \centering{\cite{rai2021hybrid}} & \RaggedRight{Finite element models to form input feature space comprised of damage specific features, for training a neural network} & \RaggedRight{Structural health monitoring}\\ \hline \\

    \RaggedRight{A personalized diagnosis method to detect faults in gears using numerical simulation and extreme learning machine} & \centering{\cite{liu2020personalized}} & \RaggedRight{Finite element simulation generated fault data, for use with Extreme Machine Learning classification} & \RaggedRight{Machine Condition Monitoring for gearboxes} \\ \hline \\ 

    \RaggedRight{Physics-informed machine learning model for battery state of health prognostics using partial charging segments} & \centering{\cite{kohtz2022physics}} & \RaggedRight{Finite element simulation of dominant degradation mode, Gaussian process regression for learning relation between voltage curve and solid electrolyte buildup} & \RaggedRight{State of health monitoring and remaining useful life estimation for lithium-ion batteries}\\ \hline \\ 

    \RaggedRight{Physics-informed machine learning assisted uncertainty quantification for the corrosion of dissimilar material joints} & \centering{\cite{bansal2022physics}} & \RaggedRight{Finite element corrosion model to simulate the corrosion process, generated data employed to train a Gaussian Process model} & \RaggedRight{Structural health monitoring for corrosion damage estimation} \\ \hline \\
     
    \RaggedRight{Hybrid deep fault detection and isolation: Combining deep neural networks and system performance models} & \centering{\cite{chao2019hybrid}} & \RaggedRight{Calibration-Based system performance models, informed feature selection for variational autoencoder and artificial neural network classification} & \RaggedRight{Machinery fault diagnosis in turbine engines} \\ \hline \\ 

    \RaggedRight{Fusing physics-based and deep learning models for prognostics} & \centering{\cite{chao2022fusing}} & \RaggedRight{Parameter estimation with physics-based models, classification with artificial neural networks} & \RaggedRight{Machinery fault diagnosis and remaining useful life estimation in turbine engines}\\ \hline \\ 

    \RaggedRight{Physics-informed neural networks for electrode-level state estimation in lithium-ion batteries} & \centering{\cite{li2021physics}} & \RaggedRight{Electrochemical-thermal model for the generation of synthetic data, for use with an artificial neural network for estimation of electrochemical state at different spatial positions} & \RaggedRight{Remaining useful life estimation and state of health monitoring of lithium-ion batteries} \\ [1.3cm] 

    \RaggedRight{A physics-informed machine learning model for porosity analysis in laser powder bed fusion additive manufacturing} & \centering{\cite{liu2021physics}} & \RaggedRight{Feature engineering with derivation of physical effects using machine operating parameters, for use as feature space in a support vector regressor} & \RaggedRight{Monitoring for porosity buildup in components during the additive manufacturing process}\\ \hline \\ 

    \RaggedRight{Physics-informed Cyber-Attack Detection in Wind Farms} & \centering{\cite{alotibi2022physics}} & \RaggedRight{Physics-based power inequalities as an indicator of deviations from nominal operations, classification with the isolation forest algorithm} & \RaggedRight{Anomalous behavior detection and monitoring of cyber-physical assets} \\ \hline \\

    \RaggedRight{Physics-Based Method for Generating Fully Synthetic IV Curve Training Datasets for Machine Learning Classification of PV Failures} & \centering{\cite{hopwood2022physics}} & \RaggedRight{Avalanche breakdown model simulations of string-level current-voltage curves, detection with 1-dimensional convolutional neural network} & \RaggedRight{Fault detection and diagnosis in photovoltaic cells} \\ \hline \\ 

    \RaggedRight{Hybrid model of a physics-based model and machine learning for real-time estimation of unmeasurable parts: Mapping from measurable to unmeasurable variables} & \centering{\cite{kaneko2022hybrid}} & \RaggedRight{Multiple mass-spring-damper models for the generation of labeled time series data, gated recurrent unit recurrent neural network for the prediction of parameters} & \RaggedRight{Estimation of parameters and anomalous behavior detection in offshore drilling systems}\\ \hline \\ 

    \RaggedRight{Physics-informed deep learning for tracker fault detection in photovoltaic power plants} & \centering{\cite{zgraggen2022physics}} & \RaggedRight{Generation of fault data through physics-informed corruption of operational data, classification with a 1-dimensional convolutional neural network} & \RaggedRight{Fault detection and diagnosis in photovoltaic power plants}\\ \hline \\ 

    \RaggedRight{A Combined Machine Learning and Physics-Based Tool for Anomaly Identification in Propulsion Systems} & \centering{\cite{darr2023combined}} & \RaggedRight{Automatic simulation of anomalies in fluid networks with real-time fault detection and classification using long short-term memory recurrent neural network} & \RaggedRight{Anomalous behavior detection in propulsion systems, Automation of Simulation of Anomalies}\\ \hline \\ 
    
    \RaggedRight{Physics-informed long short-term memory networks for response prediction of a wind-excited flexible structure} & \centering{\cite{tsai2023physics}} & \RaggedRight{Data generation through mathematical model optimized aerodynamic and aeroelastic parameters for the response of the structure, with a long short-term memory prediction framework} & Structural Health Monitoring \\ \hline \\ 

    \RaggedRight{A novel scalable method for machine degradation assessment using deep convolutional neural network} & \centering{\cite{li2020novel}} & \RaggedRight{Establishment of health indicators via high-fidelity physics-based methods. Convolutional Neural Network employed to map monitored low-fidelity data to established health indicators} & \RaggedRight{Machinery degradation modeling and remaining useful life estimation} \\ \hline \\

    \RaggedRight{Real-Time Faulted Line Localization and PMU Placement in Power Systems Through Convolutional Neural Networks} & \centering{\cite{li2019real}} & \RaggedRight{Feature engineering based on substitution theory, convolutional neural network based classifier for fault localization} & \RaggedRight{Fault diagnosis and localization in electrical grids} \\ \hline \\

    \RaggedRight{Comparative Study between Physics-Informed CNN and PCA in Induction Motor Broken Bars MCSA Detection} & \centering{\cite{boushaba2022comparative}} & \RaggedRight{Extraction of fault correlated features in the frequency domain through Fourier transforms and processing in the frequency domain for physically relevant features, detection via convolutional neural networks} & Anomalous behavior detection in induction motors\\ \hline \\ 

    \RaggedRight{Physics-informed machine learning for sensor fault detection with flight test data} & \centering{\cite{de2020physics}} & \RaggedRight{Dynamic mode decomposition with control to extract dominant features, classifications with decision tree} & \RaggedRight{Anomalous behavior detection for sensor faults in commercial flight test data} \\ \hline \\

    \RaggedRight{Physics-Informed Machine Learning for Degradation Modeling of an Electro-Hydrostatic Actuator System} & \centering{\cite{ma2023physics}} & \RaggedRight{Features and model hyperparameters selection through failure mechanism of system, classification with a long short-term memory network} & \RaggedRight{State of health monitoring for an electro-hydrostatic actuator system}\\ \hline \\ 

    \RaggedRight{Roll Wear Prediction in Strip Cold Rolling with Physics-Informed Autoencoder and Counterfactual Explanations} & \centering{\cite{jakubowski2022roll}} & \RaggedRight{Generation of new physics-driven features correlated to physical wear of asset, for use in training an autoencoder for wear prediction} & \RaggedRight{Machinery health monitoring for degradation prediction} \\ \hline \\ 

    \RaggedRight{Physics-Informed Feature Space Evaluation for Diagnostic Power Monitoring} & \centering{\cite{green2022physics}} & \RaggedRight{Feature selection through evaluation of relevance through time, feature separability check using geometric overlap with respect to hyper-ellipsoidal regions, evaluated through SVM and neural network.} & \RaggedRight{Condition monitoring and power monitoring om electro-mechanical system} \\ \hline \\

    \hline \hline \\[0.01cm]

\end{longtable}
\end{tiny}
\end{singlespacing}

\begin{figure}[h!]
\begin{center}
\includegraphics[width=\textwidth]{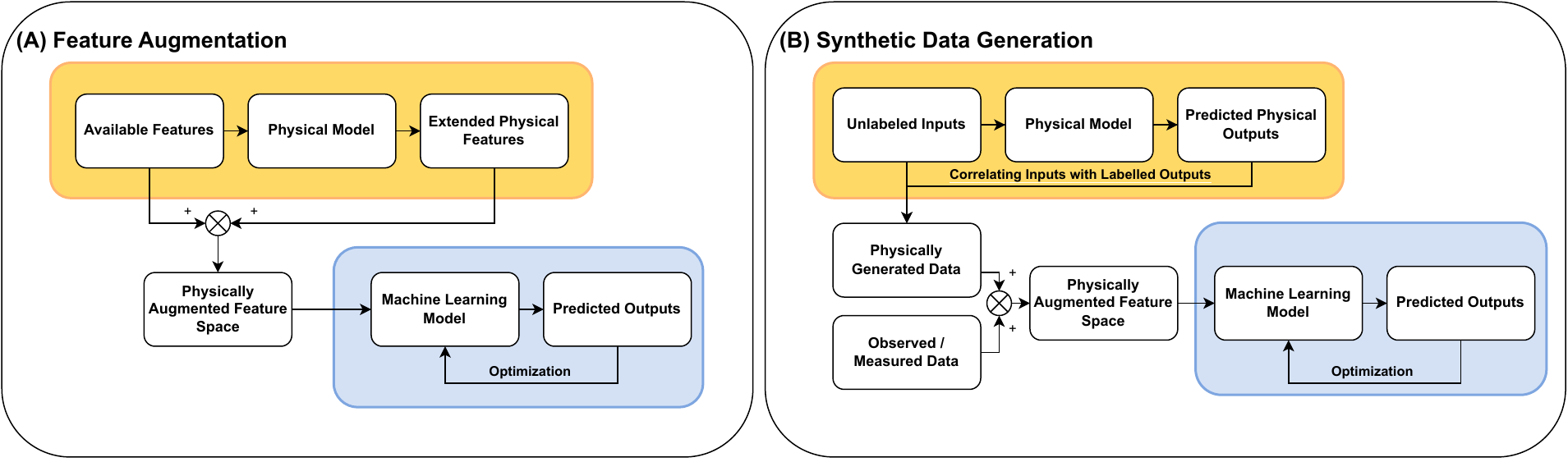}
\end{center}
\caption{General outline for the process of the generation of synthetic data via physics-based methods.}\label{physgen}
\end{figure}

Physics-based models may be used to simulate a wide range of physical systems. Through augmentation of feature space from said models, ML algorithms may be trained to accurately predict the behavior of these systems based on grounded, albeit potentially incomplete physical principles. This approach is preferred due to the ease of generation of large quantities of generally reliable data, as well as its capability to circumvent many practical and ethical concerns \citep{de2021next}. For example, additional features may be extracted or generated through knowledge of the system itself, forming an augmented feature space \ref{physgen}(A). Alternatively, large quantities of labeled data may be obtained from parsing unlabeled inputs through a physical or numerical simulation model, for a physically generated output. Thereafter, the labels and generated outputs may be used in the training process, as illustrated by Figure \ref{physgen}(B). \\

Subtractive feature engineering involves mainly feature selection: a technique commonly employed in ML algorithms to select features that are relevant and meaningful to the problem. Leveraging physics-based constraints, a physics-informed feature selection strategy may aim to identify and retain the most critical features for accurate and interpretable predictions. In addition to the plethora of implementations mentioned above, the action of generating synthetic data has also been semi-automated through deep learning structures known as generative adversarial networks. In these structures, a generator and a discriminator neural network are trained simultaneously via physics-informed regularization to produce physically consistent synthetic data. More information regarding the networks in particular, as well as several examples of implementations in literature may be found in Section \ref{Generative Networks}: Generative Deep Learning Networks.

Prior to the popularization of the physic-informed machine learning paradigm, early studies have already made use of the various aforementioned advantages and properties of physics-guided synthetic data generation to generate physically consistent results on a large scale for use in the training process of data-driven models. Rather than defining the data-driven model from scratch, the \emph{a-priori} parameters or variables defined in or by the physics-based models are used to full effect. For example, early works by \cite{tian2015motor} and \cite{frank2016hybrid} made use of informed data pre-processing techniques and physical models to generate or supplement the input feature spaces of their respective ML models. In the work of \cite{tian2015motor}, the authors explored an informed strategy for feature extraction with applications in the monitoring and diagnosing of bearing faults within electrical motors. Known frequency domain fault features were extracted via spectral kurtosis, and were subsequently utilized to train a semi-supervised K-Nearest Neighbour (kNN) algorithm. \cite{frank2016hybrid} proposed a hybrid model for fault diagnostics and anomaly detection in building energy usage. The authors employed a high-fidelity system model to supplement available data for use in data-driven models. Data generated comprises of the system in both the healthy and faulty state and served to supplement available historical data from a statistical model, and observed data. A variety of classification algorithms, such as the Support Vector Machine (SVM), and Random Forest (RF) are presented to classify anomalous behaviors from data. In more recent times, \cite{karandikar2021physics} proposed a logistic classification scheme to model the degradation of machine tools making use of known physical laws as constraints to the model. In their study, the non-linear physical relationship between cutting speed and tool life is embedded through a logarithmic manipulation of the input parameters. Transforms of input variables such as cutting speed and time are used as the input feature space for a logistic classifier model to ensure physical consistency with the Taylor tool life model, enforcing linearity in the logarithmic space. Following this, the logistic classifier outputs the probability of degradation state in the tool. Similarly, \cite{li2020novel} proposed a deep CNN-based surrogate model for tool wear monitoring. The model employs high-fidelity information from sensors, informed via physics-based methods such as vibration modal analysis or finite element analysis. Physics-based methods are employed to not only optimize the data collection procedure by determining sensor placements but also as a feature engineering mechanism for the construction of health indicators. A machine learning model is subsequently trained to learn the relationship between low-fidelity signals and established health indicators. \cite{hao2023physics} introduced a framework for the estimation of notch fatigue degradation in poly-crystalline alloys through the embedding of various physical parameters in the input feature space. Employing a sensitivity analysis, key parameters governing the behavior are identified: Physics-driven parameters introduced involve un-notched specimen reference life, derived from the Basquin model, stress state and stress ratio at the notch root, from Neuber’s rule, and energy-type damage parameter from the Smith-Watson-Topper model. In all, the Latin hypercubic sampling-based PIML model introduced was shown to have superior generalizability and predictive capabilities. \\

\begin{figure}[h!]
\begin{center}
\includegraphics[width=\textwidth]{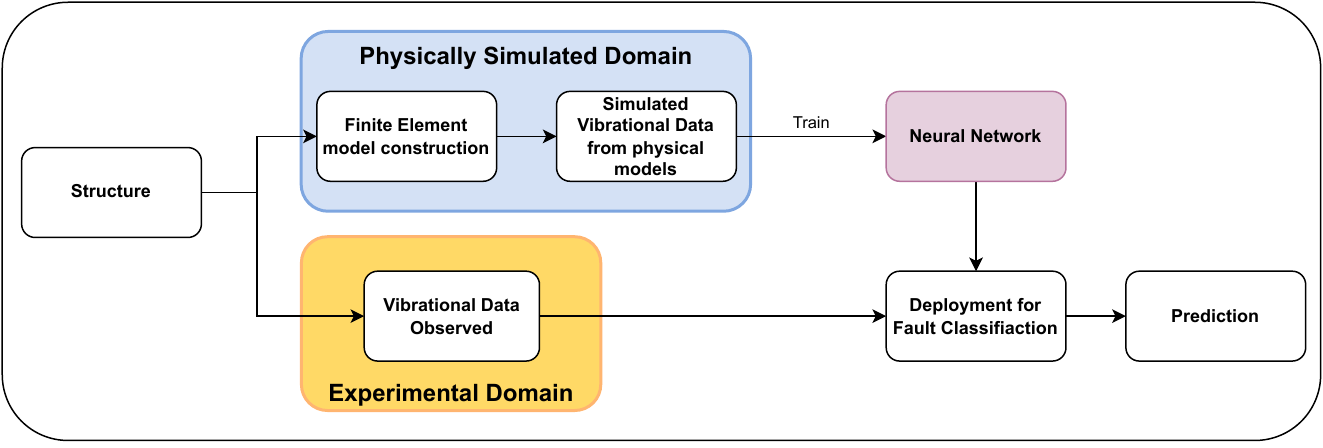}
\end{center}
\caption{Finite Element model of the structure monitored constructed to provide simulated training data for Neural Network, as demonstrated in the work of \cite{seventekidis2020structural}.}\label{FEAgen}
\end{figure}

A common theme in existing literature, for applications involving solid structures such as structural health or machinery health monitoring, is to employ finite element models to generate physical data. With their inherent versatile and robust nature in simulating complex real-world systems, finite element models provide a systematic approach for predicting and analyzing various physical behaviors through the discretization of complex geometries into smaller elements. More specifically, each element is modeled using mathematical equations that describe the physics governing the behavior of the particular element. In this format, governing or constitutive equations representing the physics of the system may be embedded within the feature space of the ML model itself; equations governing physical phenomena such as the laws of conservation of mass, momentum, and energy, as well as material properties and boundary conditions may be represented and loosely enforced. A variety of studies establishes the physical model through finite element simulations in which the physics of the system is incorporated via mathematical formulations. For example, \cite{seventekidis2020structural} utilized the finite element model as a source of simulation data to train an ML model for damage identification problems for structural health monitoring applications, with the procedure employed illustrated in figure \ref{FEAgen}.

The health state classification model is trained solely on labeled structural response vibrational data generated through a finite element model in various loading conditions. The resultant CNN-based classifier was applied to a benchmark linear beam structure with good accuracy in determining damage states. \cite{rai2021hybrid} employed an artificial neural network for damage localization and detection under the lamb wave response in an aluminum sample. In their work, various finite element simulations are employed for the construction of damage-specific features, in a system which the authors have termed the damage parameter database. Subsequently, the database is used as the input layer in the training process of an artificial neural network, whereby parameters are updated using the robust Levenberg–Marquardt algorithm. \cite{liu2020personalized} similarly employed finite element methods for the simulation of fault data. In their work, the authors introduced a gearbox fault diagnostics pipeline whereby finite element methods were employed to numerically simulate fault samples during gearbox operations. Signals obtained are separated into the time and time-frequency domains for use in the generation of fault samples in training an extreme learning machine model. \cite{kohtz2022physics} employed a Gaussian process regression for prognostics and estimating the remaining useful life of a lithium-ion battery. The effect of the dominant degradation process, the build-up of solid electrolyte interface, is modeled from a physical finite element simulation. Subsequently, results from the physical model are used in combination with experimental data to train a co-kriging-based multi-fidelity model. Through the model, an empirical relation between measured voltage curves and the state of health of the lithium-ion battery is derived. \cite{bansal2022physics} studied the effect of galvanic corrosion on joints comprised of differing materials. The authors proposed a framework whereby feature selection is performed based on results from physical simulation. More specifically, a finite element model was employed to simulate material loss due to galvanic corrosion, while taking into account environmental factors. Subsequently, based on the results of a sensitivity analysis, parameters most correlated to material loss are selected as features for use in PIML-based surrogate modeling of the joints. \\

Incorporation of synthetically generated data or features may prove invaluable in systems where data collection remains a limiting factor. PIML models are commonly employed to estimate difficult-to-observe variables in a variety of applications. Leveraging physical constraints, models are capable of providing insights into the behavior of complex systems, even when direct measurements are limited or unavailable. For instance, in the work of \cite{chao2019hybrid}, the authors explore a hybrid approach for fault detection and isolation in engines. In their study, a physical model of an engine is constructed and non-observable process variables are inferred with the Unscented Kalman Filter. Through this process, the authors effectively enhance the feature space of the two data-driven diagnostics models explored, based on Artificial Neural Networks (ANNs) and Variational Auto-Encoders (VAEs) respectively. Using this study as a basis, the authors further expanded on this model with their proposed hybrid framework for prognostics and Remaining Useful Life (RUL) estimation in a fleet of engine systems. In another study by the same author, a physical model of the system was employed to estimate difficult-to-measure parameters of the system relating to component health. In combination with observed data, the estimated parameters are fed in as data to a neural network, forming a physics-augmented feature space \citep{chao2022fusing}. Further examples include the work of \cite{darr2023combined}, who sought to detect and alleviate issues associated with anomalies in propulsion systems during launch. The group proposed a novel data generation scheme that automates the process of physical simulations for the creation of anomalous data. The group utilized an LSTM network for the detection of anomalous behaviors and events. \cite{alotibi2022physics} created a framework for the detection of false data injection attacks on the operation of wind turbines. Monitored parameters such as power output from the physical asset are parsed through a physics-based model, whereby based on the law of kinetic energy, augments the available feature space for ML. A physics-informed Isolation Forest is employed to perform the anomaly detection. The algorithm combines historical temporal data from measurements with feature augmentations from the physics-based model to create an ensemble of Random Forests for anomaly detection. The authors demonstrated the increase in anomaly detection accuracy of the integration of physics in their proposed framework by applying the framework to a real-world dataset. \\

\begin{figure}[h!]
\begin{center}
\includegraphics[width=\textwidth]{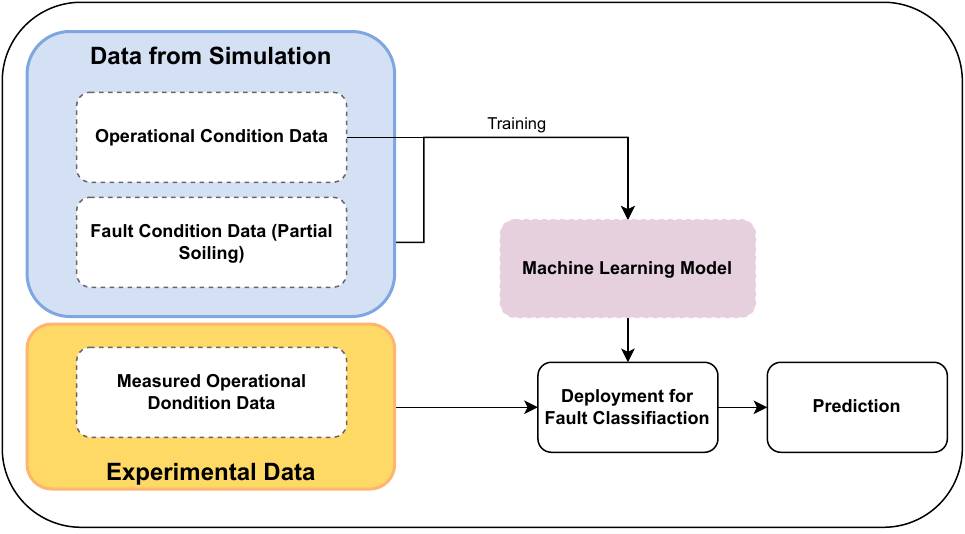}
\end{center}
\caption{Data augmentation employed to incorporate simulated fault and operation data for the training process of a machine learning fault classification algorithm, adapted from \cite{hopwood2022physics}.}\label{simdataaug}
\end{figure}

With respect to monitoring the state of health in electrochemical applications, \cite{li2021physics} employed a high-fidelity electrochemical-thermal physical model for the generation of non-observable data regarding the electrochemical states in batteries. Variables generated such as lithium-ion concentrations and electric potentials were used in the training process of a neural network which learns the nonlinear relationship between observable data and data which cannot be measured physically. In another study by \cite{hopwood2022physics}, the authors primarily employed physical modeling to overcome cost issues associated with the high-fidelity condition monitoring of photovoltaic arrays. The group proposed a fully synthetic training dataset based on physical simulations of photovoltaic arrays in the healthy state, partial soiling fault state, and cell crack fault state, whereby the framework is illustrated in figure \ref{simdataaug}. 
Data generated were employed to train a 1-dimensional CNN for the classification of fault states, and the effectiveness of this approach was validated with observational data. From experimentation, the accuracy of the ML model trained on the synthetic dataset was identical to that of the observed data. A similar strategy is employed by \cite{zgraggen2022physics}, utilizing synthetically generated data to supplement available labeled fault data. Due to the scarcity of labeled data for fault scenarios, the authors proposed a fault generation strategy via physics-informed corruption of available normal operational data, based on a model of the correlating irradiance and power produced given the tilt angle of tracking sensors. Through the physical model, the group augmented training data for a CNN model in diagnosing anomalous conditions of tracking sensors in a fleet of solar panels in a photovoltaic power plant. \\

\begin{figure}[h!]
\begin{center}
\includegraphics[width=\textwidth]{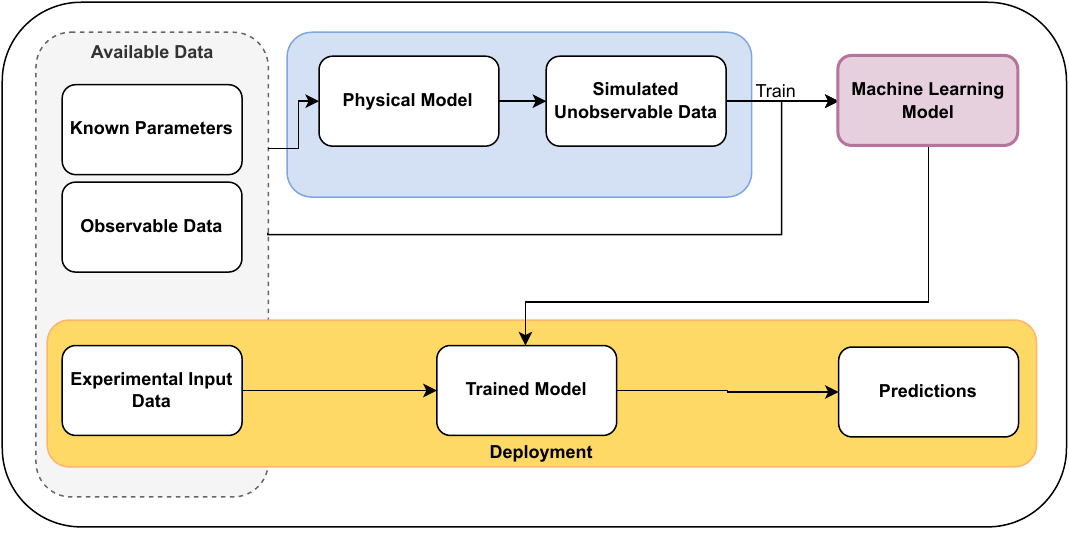}
\end{center}
\caption{The hybrid model, featuring physics-based modeling as a basis to map the observable parameters to unobservable parameters, for input to the machine learning algorithm.}\label{hybridPhysMl}
\end{figure}

With respect to applications in health monitoring in structural components, \cite{tsai2023physics} further automated the process of data generation through their proposed LSTM for monitoring and response prediction of a structure subjected to excitation by wind. The authors employed a mathematical model based on optimized aerodynamic and aeroelastic parameters to generate synthetic data on the response of the structure. To further facilitate data generation and avoid the computational cost that is associated with the mathematical models, the mathematical model was employed to train an intermediary LSTM network to automate the generation of large quantities of data while maintaining relative adherence to physical principles of structural response. Data generated from the simulated response was further employed to train an LSTM classifier, in conjunction with monitoring data to predict structural response. Similarly, \cite{kaneko2022hybrid} employed a physics-informed data generation scheme for the estimation of non-observable parameters in offshore drilling systems. Input data for the model are generated through a physical model of the system, whereby various input parameters are fed into the system to obtain the measurable data and identify the unmeasurable data. The general process of which is illustrated in figure \ref{hybridPhysMl}. 

Subsequently, a Gated Recurrent Unit (GRU) type recurrent neural network (RNN) is trained to derive the relation between the various inputs, outputs, parameters, and measurable data from the physical model, and the unmeasurable data. \cite{liu2021physics} proposed a novel generalizable physics-informed model for the monitoring and prediction of porosity during the additive manufacturing process. Rather than directly correlating machine operating parameters to porosity buildup within the part, the authors instead derived the direct physical effects of machine operating parameters such as energy density and pressure distribution. Using physical interpretations as the input feature rather than the machine parameters allows for a generalizable, machine-independent diagnostics framework yielding superior predictive capabilities. \\

\begin{figure}[h!]
\begin{center}
\includegraphics[width=\textwidth]{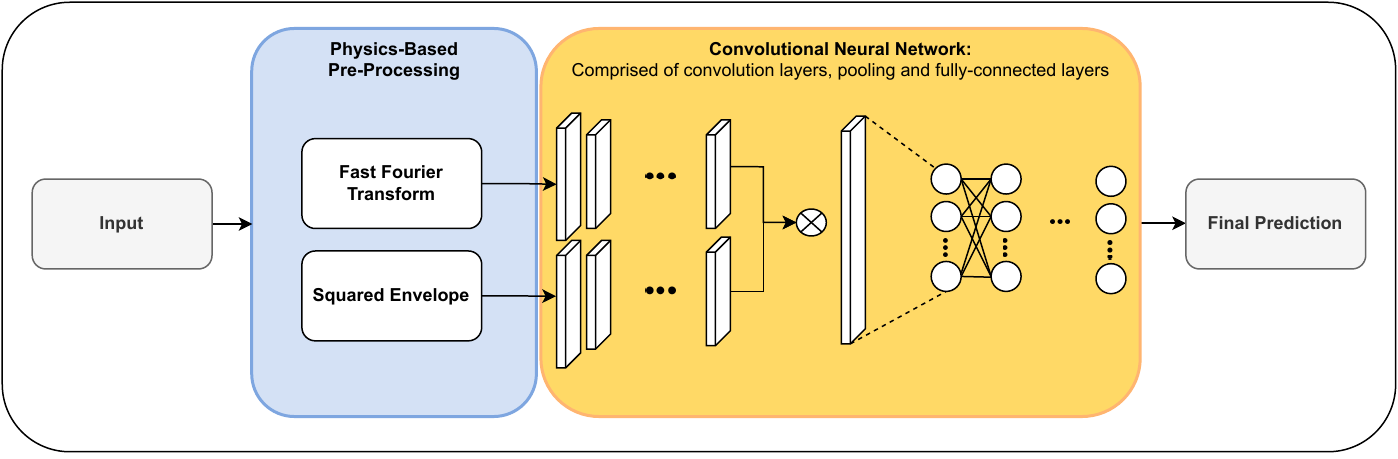}
\end{center}
\caption{}\label{PhysPreprocess}
\end{figure}

In addition to augmenting the input feature space, physics-guided methods have also been employed for feature selection and feature engineering. Through the integration of physical constraints, equations, or relationships into the feature selection algorithm, practitioners are better capable of identifying essential features that align with underlying physical mechanisms, providing a more robust and interpretable model for data analysis, prediction, and decision-making. In the work by \cite{li2019real}, the authors proposed a feature vector with physical interpretations based on the principles of substitution theorem for fault localization in a power grid system. The feature vector was parsed through a CNN to drastically lower the required network complexity for effective fault localization. Another example of this is apparent in the work of \cite{boushaba2022comparative}, whereby the authors compared the effectiveness of a physics-informed CNN approach for the detection of faults within induction motors. Of note in their study, prior to classification with the network devised, measurements from the motor current signature analysis were pre-processed in the frequency domain through Fourier transforms to form the input to the network, as illustrated in figure \ref{PhysPreprocess}. 

Here, the pre-processing step mainly serves as a method for feature selection, extracting certain sub-bands from the signal spectrum correlating to faulty components. \cite{de2020physics} automated the process of sensor fault detection in a system with multiple fault classes. Due to the complexity and high dimensionality of the system, Dynamic mode decomposition with control (DMDc), as defined by \cite{proctor2016dynamic}, was employed to identify a linear time-invariant model of sensor readings with respect to time. Although DMDc is data-driven, the methodology itself allows practitioners to identify and extract underlying coherent structures, or modes, from complex data. From this, DMDc may reveal the dominant patterns of behavior in a system and provide insights into its underlying physics. The model is applied with a Kalman observer, which provides an estimate of sensor measurement variables of the healthy state in real-time. For the classification of anomalies, features are in part derived from the DMDc procedure. During validation, features expected by the decision tree may be computed with the linear-time invariant system, and measurement anomalies are classified.  \cite{ma2023physics} investigated the degradation mechanism of an electro-hydro-static Actuator system by employing a physics-informed Long Short Term Memory (LSTM) network. Due to the complexity of the degradation mechanism, the authors performed a physics-informed selection of features, and model hyperparameters were performed based on the failure mechanism of the system. In their study, the physical state of the system is represented by a physical parameter indicator: the rise time. Based on the physical state of the system, the monitoring dataset is selected and split into a training and test dataset, which is employed to train and evaluate an LSTM network. Evaluation of network performance with different hyperparameters is performed through the selected dataset, and the parameters corresponding to the most accurate predictions are selected. Finally, in the work of \cite{jakubowski2022roll}, the authors proposed a physics-informed autoencoder model for the estimation of roll wear in equipment during the process of cold-rolling.  Similar to the above cases, input space augmentation was performed employing physics-based simulation models. In this case, information for parameters relevant to wear from cold-rolling, such as friction coefficients and forward slip, was generated with the prior knowledge available. The roll wear prediction is performed with an autoencoder, whereby data extracted from early stages of degradation, in conjunction with physically derived features, were utilized to train the autoencoder. Predictions for roll wear were performed based on deviations from the established nominal state. Furthermore, through counterfactual explanations methods, the authors sought to improve the interpretability of predictions from the network. \\

Authors have also proposed methodologies for the selection of feature space subject to evolution over time. \cite{green2022physics} presents a strategy for a physics-informed feature space evaluation in monitoring of electro-mechanical loads. Features were curated based on a load separability verification, in which the reliability of past training data for future classifications is evaluated. The underlying physics of the deviation over time is represented via the geometry of hyper-ellipsoidal regions generated by principle component analysis. Through this, the authors addressed the issue of separability in a system under a multi-load scenario subject to operational or degradation drift. The authors have demonstrated the effectiveness of their approach through both linear and non-linear classifiers, namely an SVM and a neural network.

Overall, feature augmentation by means of previously known physical principles represents an easy-to-implement approach to enforce soft constraints to machine learning algorithms. By tailoring the feature space in which the algorithm specifically is consistent with physics, the predictive capabilities of the algorithms are more likely to fall within the domain of physical feasibility. Conversely, while the models may be built upon physically consistent training data, the method by which they arrive at their predictions remains an enigma to practitioners. Furthermore, the loose constraints to the feature space of the model, rather than the model itself makes these type of algorithms especially prone to occasional predictions that are inconsistent with physical laws.

\subsubsection{Transfer Learning}
Another method of integration for ML algorithms may be through the Transfer learning (TL) procedure. TL is a technique commonly employed in machine learning and deep learning applications, whereby a model trained to perform a certain task is adapted to perform alternate tasks sharing similarities to the original. It has become prominent due to its ability to improve performance and reduce training requirements and has seen a great deal of use in applications such as image analysis, natural language processing, and speech recognition for its time and data efficiency. With transfer learning, the pre-trained model effectively acts as a vessel for feature extraction, leveraging learned features from the source domain and re-purposing for the target domain. Through this process, training time and resources required are drastically reduced, making TL suitable for mitigating the cost of complex deep learning architectures. A summary of compiled works may be found in Table \ref{table:2}. 

\begin{singlespacing}

\begin{tiny}

\begin{longtable}[h!]{p{0.25\textwidth}  p{0.11\textwidth} p{0.25\textwidth} p{0.25\textwidth}}

    \hline \hline \\
    
    \textbf{Article Title} & \textbf{Citation} & \textbf{Description} & \textbf{Application} \\ 
    \hline \hline \\[0.05cm]
    \endhead

    \hline
    \endfoot
    \caption{Literature compiled for studies employing transfer learning algorithms to learn physically relevant features.}
    \label{table:2}
    \endlastfoot
    
    \RaggedRight{Fault Cause Assignment with Physics Informed Transfer Learning} & \centering{\cite{guc2021fault}} & \RaggedRight{Dynamic mode decomposition with control extracts features representing physics of dynamics, continuous wavelet transforms represents modes in the time-frequency domain, classification with pre-trained GoogLeNet CNN} &  \RaggedRight{Fault diagnosis in fault source separation in sensor-actuator system} \\ \hline \\

    \RaggedRight{Sensor Fault Diagnostics Using Physics-Informed Transfer Learning Framework} & \centering{\cite{guc2022sensor}} & \RaggedRight{Dynamic Mode Decomposition with control extracts features representing physics of dynamics, Continuous Wavelet Transforms represents modes in the time-frequency domain, classification with pre-trained GoogLeNet CNN} & \RaggedRight{Fault diagnosis and fault source separation in sensor-actuator system}\\ \hline \\

    \RaggedRight{A physics-informed transfer learning approach for anomaly detection of aerospace cmg with limited telemetry data} & \centering{\cite{gong2021physics}} & \RaggedRight{Neural network established to represent the system in the healthy state, based on power balance equations, parameters of the network defined as degradation features fixed for a healthy state, with fine-tuning to account for degradation conditions. Anomaly detection via kernel density estimation} & \RaggedRight{Anomalous behaviour detection in aerospace control}\\ \hline \\

    \RaggedRight{Physics-guided, data-refined modeling of granular material-filled particle dampers by deep transfer learning} & \centering{\cite{ye2022physics}} & \RaggedRight{Artificial neural network trained on physical model based on governing and constitutive equations of particle dampers, re-calibrated on high-fidelity observational data} & \RaggedRight{State of health monitoring for particle dampers} \\ \hline \\

    \RaggedRight{Using Transfer Learning to Build Physics-Informed Machine Learning Models for Improved Wind Farm Monitoring} & \centering{\cite{schroder2022using}} & \RaggedRight{Artificial neural network pre-trained on Monte-Carlo simulation database of turbines operation data, re-calibration with available data} & \RaggedRight{Anomalous behavior detection in wind turbine sensor data}\\ \hline \\
 
    \RaggedRight{Multi-fidelity physics-informed machine learning for probabilistic damage diagnosis} & \centering{\cite{miele2023multi}} & \RaggedRight{Artificial neural network trained on low-fidelity finite element simulation, transference of low fidelity trained layers and re-calibration with high-fidelity finite element simulation data} & \RaggedRight{Structural health monitoring in concrete structures}\\ \hline \\

    \RaggedRight{Intelligent fault diagnosis of machinery using digital twin-assisted deep transfer learning} & \centering{\cite{xia2021intelligent}} & \RaggedRight{Sparse de-noising autoencoder trained on fault conditions produced by a digital twin of asset} & \RaggedRight{Fault detection and diagnosis in pump system}\\ \hline \\

    \RaggedRight{Digital-twin assisted: Fault diagnosis using deep transfer learning for machining tool condition} & \centering{\cite{deebak2022digital}} & \RaggedRight{Stacked sparse autoencoder trained on simulated dataset} & \RaggedRight{Condition monitoring for machine tools}\\ \hline \\
    
    \RaggedRight{Structural damage detection based on transfer learning strategy using digital twins of bridges} & \centering{\cite{teng2023structural}} & \RaggedRight{Convolutional neural network trained on a digital twin of asset} & \RaggedRight{Structural health monitoring of bridge Structures}\\ \hline \\

    \RaggedRight{Digital twin-driven intelligent assessment of gear surface degradation} & \centering{\cite{feng2023digital}} & \RaggedRight{Convolutional neural network trained on digital twin of asset} & \RaggedRight{Condition monitoring for gear surface degradation}\\ \hline \\
    
    \hline \hline \\[0.01cm]
    
\end{longtable}
\end{tiny}
\end{singlespacing}

\begin{figure}[h!]
\begin{center}
\includegraphics[width=\textwidth]{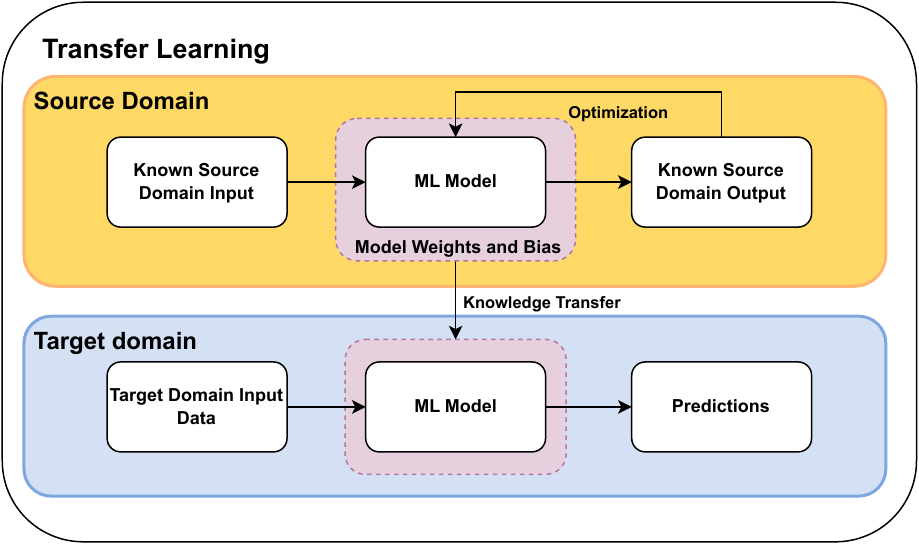}
\end{center}
\caption{Illustration depicting the principles and functioning of Transfer Learning in machine learning, a technique in machine learning that leverages knowledge gained from one task to improve performance on another related task.}\label{fig:2}
\end{figure}

According to the definition by \cite{pan2010survey}, the transfer learning framework operates on the source domain \(D\) whereby \(D = \{\mathcal{X}, P(X)\}\), defined by input features space \(\mathcal{X}\) and marginal probability \(P(X)\). Here, \(X\) represents a sample data, comprised of vectors from the feature space: \(X = \{x_{1},...,x_{n}\}, x_{i}\in\mathcal{X}\). Similarly, a label space may be defined for the data as \(\mathcal{Y}\). Thus, for a defined domain, a task \(T\) may be defined as \(T = \{\mathcal{Y}, P(Y|X)\} = \{\mathcal{Y}, \eta\}, Y = \{y_{1},...y_{n}\}, y_{i}\in\mathcal{Y}\), whereby the predictive function \(\eta\) is learned from labeled data pairing of \(\left(x_{i},y_{i}\right)\), such that \(\eta\left(x_{i}\right)=y_{i}\). For a given target domain \(D_{t}\) and unknown learning task \(T_{t}\), the objective of a transfer learning framework is to employ a learned predictive function \(\eta\) based on latent knowledge gained from source domain \(D_{s}\) and known learning task \(T_{t}\). Currently, TL frameworks have been used extensively in deep learning applications. Due to the universal approximation capabilities of neural networks, the predictive function may easily be approximated by the non-linear feed-forward function. A general scheme of the operations in a typical TL framework may be seen in Figure \ref{fig:2}.

In literature, there are two main methodologies by which transfer learning may be incorporated into the PIML framework:  Leveraging the source domain, the trained model may be transferred to the target domain in various engineering applications. Source domains may be modeled based on physical models, or are defined such that the model is physically sound and consistent with physical principles. In traditional machine learning approaches, models are trained from scratch on large datasets, which may be a resource-intensive and time-consuming process. Physics-based models can provide a more accurate representation of the underlying system dynamics than purely data-driven models, which can be limited by the quality and quantity of available training data. By incorporating prior knowledge regarding underlying system dynamics in the form of physics-based models, transfer learning can reduce the computational complexity of the model and enable more efficient training and inference \citep{torrey2010transfer, zhuang2020comprehensive}. Alternatively, physics-based or physics-informed data may be parsed in as the target domain training data. The model is fine-tuned using a smaller dataset specific to the target problem, containing features related to the target problem in the physics domain. By initializing the model with the pre-trained parameters, the model already has a degree of knowledge regarding features to be learned, enabling faster convergence during fine-tuning. Fine-tuning allows the model to adapt its representations to the specific features of the target problem, thus customizing the pre-trained model for the new task. In this approach, the source domain acts more as a supporting library of learned features, allowing the TL framework to leverage said features to significantly relax target domain data requirements and expedite the training process.

Examples of the TL methodology may be seen in the work of \cite{guc2021fault}, who proposed a method of fault source identification of complex and dynamic systems through their physics-informed CNN. Through dynamic mode decomposition, a physical representation of the system may be constructed in the form of linear reduced-order spatial-temporal modes. Dynamic mode decomposition modes are then formulated into images in the time-frequency domain by means of continuous wavelet transforms. Fault conditions are classified via a CNN image classifier, leveraging a pre-trained network structure known as GoogLeNet \citep{szegedy2015going} to take advantage of learned features from other domains. The Googlenet architecture is composed of 22 main layers and employs the inception architecture with weighted Gabor filters. The authors later extended this proposed framework to perform diagnostics on the various faults that are prevalent in sensors. The effectiveness of their proposed framework is demonstrated experimentally with the real-time velocity control of the target system \citep{guc2022sensor}. \\

Through pre-trained models that have already learned relevant features, transfer learning can reduce the amount of labeled data required for training. Instead of training a model from scratch on the new dataset, the pre-trained model can be adapted to the new task by updating select layers in the network. Specific to the field of condition monitoring and anomaly detection, physics-based models have been employed to alleviate the issue of limited labeled data. In many cases, labeled data for a specific machine or failure mode may be scarce, making it challenging to train accurate models. By generating synthetic data using physics-based models, practitioners can augment the training dataset and improve the model's ability to generalize to new data. For instance, \cite{gong2021physics} facilitated the process of anomaly detection on an aerospace control moment gyro through a physics-informed transfer learning neural network. Through this framework, they were able to overcome the limitations in data with regard to the telemetry signals monitored. The non-linear relationships between telemetry signals are captured with an ANN approximating the power consumption behavior. Subsequently, the degradation of the system is captured through a transfer learning approach, whereby the last layer of the neural network model for the healthy state is fine-tuned to represent the new degradation state. A performance index was constructed based on the Mahalanobis distance, and anomaly detection was performed with the kernel density estimation approach. Similarly, \cite{ye2022physics}  employed a multi-fidelity framework, physics-based low-fidelity data generated is employed to pre-train a neural network, such that when used with the limited amount of high-fidelity experimental data available, the network demonstrated robust characterization of granular material-filled particle dampers.  \cite{schroder2022using} applied the transfer learning paradigm for anomaly detection based on the operating behavior of a wind turbine through a physics-constrained artificial neural network. The network was pre-trained on data generated by a physics-based Monte-Carlo simulation, and the transfer learning of data from the physical simulation was validated in the detection of anomalies in turbine blade angles through monitoring data. With limited data, the model demonstrated superior capabilities in both prediction accuracy and robustness due to the incorporation of physical constraints from the pre-trained network. More recently, \cite{miele2023multi} proposed a transfer learning-inspired neural network framework for structural health monitoring applications. Due to computational restrictions of high-fidelity models, the authors elected to train the network initially on low-fidelity physical model derived from a 2-dimensional finite element simulation. Model weights are held constant, and an additional layer is added to the neural network structure to re-calibrate the model with high-fidelity 3-dimensional finite element simulation. The resultant model is validated in producing the probabilistic classification in a sample concrete specimen.\\

\begin{figure}[h!]
\begin{center}
\includegraphics[width=\textwidth]{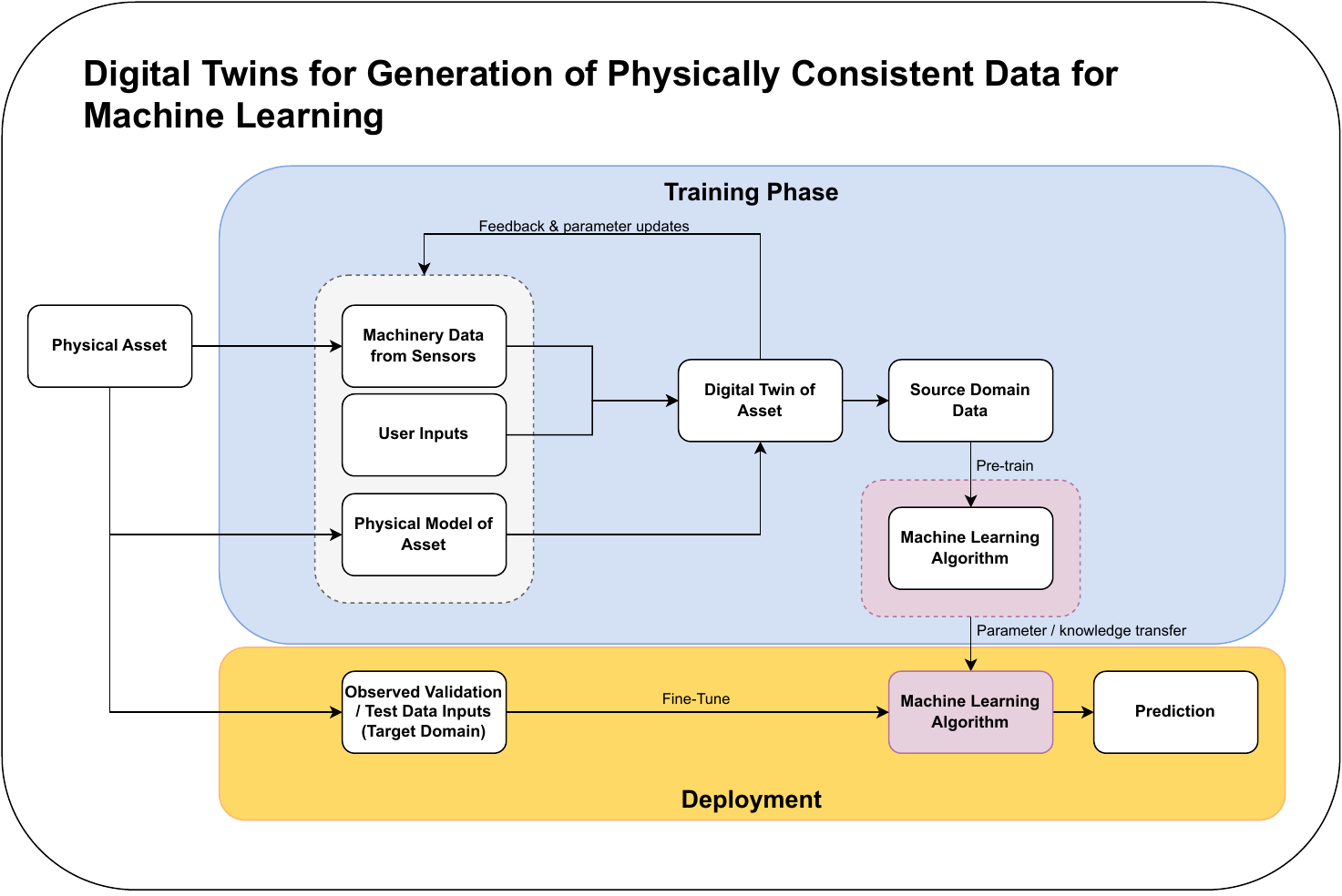}
\end{center}
\caption{Representation of the application of Transfer Learning in the context of Digital Twins, a virtual representation of a physical entity or system, showcasing the transfer of knowledge from a pre-existing Digital Twin.}\label{DT}
\end{figure}

A common form of physically constrained data of the source domain comes from digital twins (DTs). DTs are virtual replicas of physical systems or processes that mimic their real-world behavior in a digital environment. They are created by using a combination of sensor data, physics-based models, and ML algorithms to create a digital representation of the system or asset. DTs have been extensively utilized in applications such as predictive maintenance and condition monitoring, and have gained significant attention in recent years due to their ability to improve the efficiency and effectiveness of various engineering tasks \citep{liu2022digital}. A generalized formulation for employing the DT framework in conjunction with machine learning may be seen in figure \ref{DT}.

One of the key benefits of DTs is that they allow for real-time monitoring, analysis, and optimization of physical systems, enabling users to identify potential problems and make informed decisions to improve performance and efficiency. In literature, there has been growing interest in using DTs in conjunction with machine learning algorithms to create PIML frameworks. The idea behind this approach is to use data from DTs to train machine learning models that can then be applied to real-world systems to predict their behavior and optimize their performance. One challenge in using DTs for condition monitoring is that they often require significant computational resources to simulate the physical system accurately. This is where transfer learning, with its capabilities for domain adaptation, proves valuable. In the context of DTs for condition monitoring, transfer learning can be used to build a framework that leverages pre-existing DTs models to accelerate the development of PIML models. In general, the framework for deploying DTs in conjunction with TL is as follows: 

\begin{enumerate}
    \item A high-fidelity digital twin model of the physical system or process of interest is developed, capable of simulating the system's behavior under various conditions.
    \item Through the DT model, a large dataset of simulated data may be generated by varying the system's input parameters and monitoring the system's output variables.
    \item This dataset is then used to train an ML model to predict the system's behavior. Knowledge is transferred from the source domain, the DT model, to the target domain of a specific condition monitoring task with transfer learning. 
    \item Subsequently, models are fine-tuned on a smaller amount of real-world data to improve their performance on the target system.
\end{enumerate}

 Real-world data is used to adjust the model's parameters to better fit the specific system's behavior. Once trained, the adapted machine learning model may be deployed to predict the system's behavior and or to detect anomalies or deviations from normal operation. Several examples of the above framework have been utilized for various engineering tasks, for example: \cite{xia2021intelligent} proposed a transfer learning framework for diagnosing faults of a triplex pump system. A Digital twin of the physical asset was constructed to generate data that is consistent with underlying physical constraints on the system. Along with this, the authors also proposed a novel deep de-noising auto-encoder. In conjunction with the healthy state data generated by the digital twin, the autoencoder is pre-trained. Subsequently, the architecture may then be employed for anomaly detection in the physical machine. On the same topic, \cite{deebak2022digital} proposed a similar transfer learning framework featuring DT-assisted fault diagnosis, focusing mainly on condition monitoring for machine tools and equipment. The authors resolved the issue with the lack of real-world data through their proposed stacked sparse autoencoder structure, reducing the amount of physical data required for accurate predictions by the network, and improving the overall robustness of the model. \cite{teng2023structural} applied a digital twin for the diagnosis of structural fault in bridges, whereby generated data is employed in the training process of a CNN. Knowledge transfer from the simulated results was proven to be effective, as the model demonstrated superior convergence rate and accuracy, in comparison to physically naive transfer learning classification techniques. \cite{feng2023digital} applied the framework to gear surface degradation monitoring for predictive maintenance. Digital twin models were developed and fined tuned based on the governing equations for the dynamic and degradation behavior of their spur gearbox system. A CNN is established and trained on data from the DT model to assess surface pitting and tooth profile change. 

 Through the effective transfer of domain knowledge, the TL algorithms discussed above were capable of effectively utilizing physically relevant knowledge to aid in the predictive capabilities of automated learning. Through this process, several advantages present themselves. In addition to the reduced training time and data requirements discussed above, TL algorithms are also capable of improved generalization to data, dependent on the training dataset employed. Furthermore, the pre-trained model allows for improved interpretability within the overall predictive process of the model, through insights into the learned representations and the features that influence the model's decision. By their nature, TL algorithms are also designed with a particular aim to be fine-tuned to adapt to specific tasks. This property allows practitioners an added layer of flexibility in devising the final learning pipeline and its constituent components, whether those components are more-so physically derived, or data-driven. 

\subsection{Data-Enhanced Refinement of Physical Models} \label{32}
Another archetype common in literature is the use of ML models as correctional mechanisms for known inaccuracies or deficiencies between predicted and observed data. In current applications, physical models are based on simplifications and assumptions that may not accurately capture the complexity of real-world phenomena. As a result, physical models produce errors or inaccuracies in their predictions. Several works of literature focus on developing data-driven models to address these errors by learning to account for observed deviations, and subsequently, using physics-based and ML models in conjunction for the resultant predictions. In the works discussed in this section, ML models have been shown to work concurrently with physics-based models to fine-tune results based on outputs from both models. A summary of compiled works employing this strategy of integration is presented in Table \ref{table:3}. 

\begin{singlespacing}
\begin{scriptsize}
\begin{longtable}[h!]{p{0.25\textwidth}  p{0.11\textwidth} p{0.25\textwidth} p{0.25\textwidth}} \hline \hline \\
    
    \textbf{Article Title} & \textbf{Citation} & \textbf{Description} & \textbf{Application} \\ 
    \hline \hline \\[0.05cm]
    \endhead

    \hline
    \endfoot

    \caption{Literature compiled leveraging data-driven models working in tandem with physics-based models}
    \label{table:3}
    \endlastfoot

    \RaggedRight{Battery health management using physics-informed machine learning: Online degradation modeling and remaining useful life prediction} & \centering{\cite{shi2022battery}} & \RaggedRight{Recurrent neural network to model the deviation from physics-based aging model and observed aging} & \RaggedRight{State of health monitoring and degradation modeling for batteries} \\ \hline \\
    
    \RaggedRight{Probabilistic physics-informed machine learning for dynamic systems} & \centering{\cite{subramanian2023probabilistic}} & \RaggedRight{Augmentation of physics-based model with a machine learning model, Bayesian state estimation of model form error is learned by probabilistic ML structure} & \RaggedRight{Prognosis and structural response prediction under dynamic loads} \\ \hline \\
    
    \RaggedRight{Fusing physics-inferred information from stochastic model with machine learning approaches for degradation prediction} & \centering{\cite{li2023fusing}} & \RaggedRight{Bi-directional LSTM to model residual between observed and stochastic degradation model} & \RaggedRight{Structural health monitoring}\\ \hline \\

    \hline \hline \\[0.01cm]
    
\end{longtable}
\end{scriptsize}
\end{singlespacing}

\begin{figure}[h!]
\begin{center}
\includegraphics[width=\textwidth]{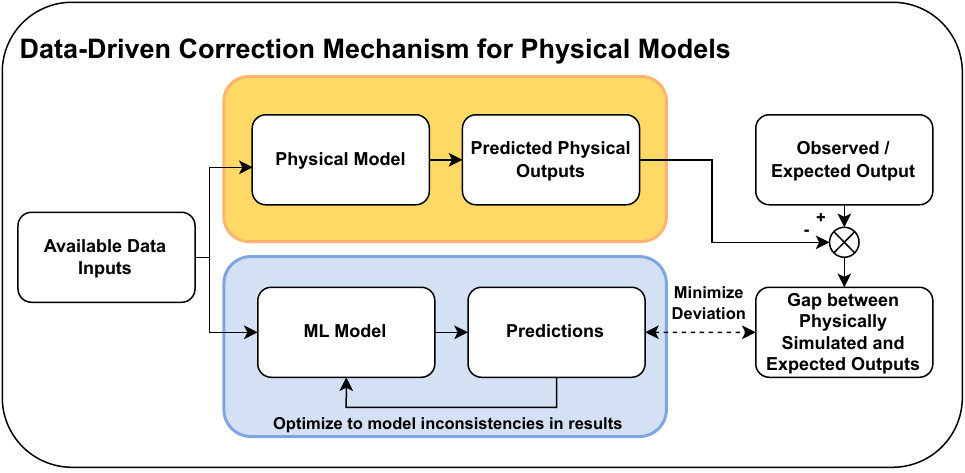}
\end{center}
\caption{General outline of correction to physics-based modeling via data-driven solutions}\label{fig:3}
\end{figure}

In this approach, a physical model is used to generate initial predictions, which are adjusted in tandem employing the predictive capabilities of an ML algorithm. The algorithm learns from the set of training data that includes both the input features used by the physical model and the corresponding ground-truth outcomes and applies this learning to generate corrections to the physical model's predictions. In literature, this strategy has often been referred to as hybrid-modeling or residual modeling. The general process by which this integration takes place is illustrated in Figure \ref{fig:3}.

Examples of this implementation are illustrated in various works, such as \cite{shi2022battery} combined a physics-based degradation model with a deep learning network to estimate the state of health in lithium-ion batteries. The physics-informed deep learning model is a combination of the physics-based calendar and cycle aging model and a Long Short-Term Memory (LSTM) model. Through parameters governing stress during operation, an initial estimate of calendar aging and cyclic aging of the battery is calculated. Thereafter, the LSTM learns the deviation between observed conditions and the predictions of the physics-based aging model over time. In conjunction, the physics-informed LSTM model was capable of accurately capturing the overall degradation trend of the asset.\cite{subramanian2023probabilistic} proposed a data-driven correction mechanism for a structure subjected to dynamic loading. Error resulting from the physics-based model were determined via Bayesian state estimation, whereby a probabilistic ML structure learns the discrepancies. In conjunction, the combined pipeline demonstrated robust predictions for linear and nonlinear systems with both Gaussian and non-Gaussian noise. Finally, \cite{li2023fusing} employed a Bi-directional LSTM to estimate the residuals between the observed degradation behavior and the degradation tendency from a two-stage stochastic degradation model. The estimated residual is used in conjunction with the outputs of the physics-driven stochastic degradation model to predict degradation in a bridge deck rebar structure. 

Though the introduction of residual learners has seen success in the above cases, limitations incurred by this architecture render it difficult to provide insightful and interpretive predictions. As the ML model learns to model the discrepancy, rather than the system itself. While studies discussed above have had success in utilizing this combination of physics-based modeling and machine learning, this key drawback severely limits its use cases, as well as its explainability and interpretability, in comparison to other architectures.  \\

\subsection{Physics-Informed Regularization} \label{33}

Regularization techniques have been fundamental in training ML models since their inception. Conventional regularization, such as Lasso (L1) or Ridge (L2) regularizations, incorporates an additional penalty term to reduce the model's capacity to overfit to data that may not be reflective of the general behavior of the system, resulting in simpler and more robust solutions.  While this has been used extensively, a new trend involves the usage of physics-based regularization with machine learning. This approach seeks to combine the advantages of physics-based models to enhance the accuracy, interpretability, and robustness of conventional data-driven solutions. Prior knowledge regarding the physical system is integrated as a part of the learning process, either as constraints or regularizers, effectively encoding the physical constraints to aid in guiding the optimization process in producing physically meaningful solutions. Past implementations of physics-based regularization involved solving the physical equations and incorporating them as constraints in the optimization problem \citep{ruhnau2007variational, oware2013physically}. However, this approach is computationally expensive and limited to physical systems that are mostly well-understood. With the recent advancements in deep learning and the availability of large amounts of data, new techniques have been developed that combine physics-based modeling and ML to be more efficient and scalable. For instance, in recent studies such as the work of \cite{raissi2019physics}, a novel regularization approach was proposed that leverages the structure of the physical system to learn more efficient representations. The proposed method, termed physics-informed neural networks (PINNs), incorporates the governing equations of the physical system as regularizers in the loss function. A summary of compiled literature employing this technique is provided in Table \ref{table:4}.

\begin{singlespacing}
\begin{tiny}
\begin{longtable}[h!]{p{0.25\textwidth}  p{0.11\textwidth} p{0.25\textwidth} p{0.25\textwidth}} \hline \hline \\
    
    \textbf{Article Title} & \textbf{Citation} & \textbf{Description} & \textbf{Application} \\ 
    \hline \hline \\[0.05cm]
    \endhead

    \hline
    \endfoot

    \captionsetup{justification=centering}
    \caption{Literature Compiled for physics-guided or physics-informed regularisation technique employed}

    \label{table:5}
    \endlastfoot
    
    \RaggedRight{Microcrack Defect Quantification Using a Focusing High-Order SH Guided Wave EMAT: The Physics-Informed Deep Neural Network GuwNet} & \centering{\cite{sun2021microcrack}} & \RaggedRight{Quantification of microcrack defects with hybrid physics-informed architecture design based on various deep learning frameworks, regularized via network structure and hybrid feed-forward and back-propagation loss} & \RaggedRight{structural health monitoring for detection of micro-crack defects} \\ \hline \\ 

    \RaggedRight{Physics-informed turbulence intensity infusion: A new hybrid approach for marine current turbine rotor blade fault detection} & \centering{\cite{freeman2022physics}} & \RaggedRight{Feature extraction via continuous wavelet transform from vibrational data. The classification was performed with a neural network, with physics-informed loss function to obtain turbulence intensity features} & \RaggedRight{Anomalous behavior detection and fault diagnosis in turbine rotor blades} \\ \hline \\

    \RaggedRight{A physics-informed neural network for creep-fatigue life prediction of components at elevated temperatures} & \centering{\cite{zhang2021physics}} & \RaggedRight{Neural network regularized via physics-informed loss function, penalizing the model for unrealistic predictions (negative or extreme values) of fatigue life} & \RaggedRight{Structural health monitoring for creep-fatigue life in steel specimen}\\ \hline \\

    \RaggedRight{Data-driven prognostics with low-fidelity physical information for digital twin: physics-informed neural network} & \centering{\cite{kim2022data}} & \RaggedRight{Physics-informed loss function penalizing deviations from expected values, determined by low-fidelity physical model} & \RaggedRight{Structural health monitoring for crack propagation} \\ \hline \\

    \RaggedRight{Long-term fatigue estimation on offshore wind turbines interface loads through loss function physics-guided learning of neural networks} & \centering{\cite{de2023long}} & \RaggedRight{Features selected through recursive feature elimination from sensors and monitoring data. Estimation of fatigue via neural network regularized by novel physics-informed loss function, reflective of priority given to long-term estimation} &  \RaggedRight{Structural health monitoring for wind turbines fatigue life} \\ \hline \\

    \RaggedRight{Physics-informed meta learning for machining tool wear prediction} & \centering{\cite{li2022physics}} & \RaggedRight{Parameters of dynamic relationships governing tool wear used to establish input space for individual models at different stages of degradation via cross physics-data fusion. Meta-learning model is employed to learn the experiences of ML models and optimized via physics-informed loss} & \RaggedRight{Tool life predictions} \\ \hline \\

    \RaggedRight{A physics-informed deep learning framework for inversion and surrogate modeling in solid mechanics} & \centering{\cite{haghighat2021physics}} & \RaggedRight{Physics-informed neural network for solving differential equations governing linear elasticity and non-linear von Mises elastoplasticity } & \RaggedRight{Elastostatics modelling in solid mechanics}\\ \hline \\

    \RaggedRight{Identification of Material Parameters from Full-Field Displacement Data Using Physics-Informed Neural Networks} & \centering{\cite{anton2021identification}} & \RaggedRight{Material parameter estimation via solution of momentum equation and governing equations of linear elasticity} & \RaggedRight{Structural health monitoring}\\ \hline \\

    \RaggedRight{Inferring vortex induced vibrations of flexible cylinders using physics-informed neural networks} & \centering{\cite{kharazmi2021inferring}} & \RaggedRight{Approximation of the linear beam-string equations via PINN for simulation of a cylindrical structure in uniform flow} & \RaggedRight{Structural health monitoring} \\ \hline \\

    \RaggedRight{Physics-Informed Machine Learning and Uncertainty Quantification for Mechanics of Heterogeneous Materials} & \centering{\cite{bharadwaja2022physics}} & \RaggedRight{Solution of PDE governing momentum balance and constitutive equations of elasticity, optimized via physics-informed loss function penalizing deviations from PDE and boundary conditions} & \RaggedRight{Surrogate modeling of elastic deformations}\\ \hline \\

    \RaggedRight{Simulation of guided waves for structural health monitoring using physics-informed neural networks} & \centering{\citep{rautela2021simulation}} & \RaggedRight{Solving PDEs governing wave propagation with PINNs, regularized by physics-informed loss function based on deviations from PDEs and boundary conditions} & \RaggedRight{Structural health monitoring in aerospace structures}\\ \hline \\

    \RaggedRight{A physically consistent framework for fatigue life prediction using probabilistic physics-informed neural network} & \centering{\cite{zhou2023physically}} & \RaggedRight{Probabilistic PINN optimized via hybrid loss function based on fatigue life distributions with respect to stress experienced} & \RaggedRight{State of health monitoring and fatigue life estimation} \\ \hline \\

    \RaggedRight{A robust physics-informed neural network approach for predicting structural instability} & \centering{\cite{mai2023robust}} & \RaggedRight{Feed-forward PINN optimized based on deviation from data, instability information, and boundary conditions} & \RaggedRight{Structural  health monitoring via estimation of structural instability} \\ \hline \\

    \RaggedRight{Machine Fault Classification using Hamiltonian Neural Networks} & \centering{\cite{shen2023machine}} & \RaggedRight{PINN encoding the laws of Hamiltonian mechanics to learn operating state of system from vibrational data, machinery state identification using network parameters as features} & \RaggedRight{Machinery fault diagnosis for rotating machinery} \\ \hline \\

    \RaggedRight{Physics-informed machine learning for surrogate modeling of wind pressure and optimization of pressure sensor placement} & \centering{\cite{zhu2022physics}} & \RaggedRight{Finite element based computational fluid dynamics model for the generation of input features, PINN employed for the solution to Navier–Stokes equations of incompressible flows, with Dirichlet and Neumann boundary conditions} & \RaggedRight{Structural health monitoring in buildings}\\ \hline \\

    \RaggedRight{Physics informed neural network for health monitoring of an air preheater} & \centering{\cite{jadhav2022physics}} & \RaggedRight{Stacked PINNs for solving non-denationalized governing equations for heat transfer between the fluids and metal interface, regularized by physics-informed loss function based on deviation from PDEs, boundary and interface conditions} & \RaggedRight{Condition monitoring and health monitoring in air heating system } \\ \hline \\

    \RaggedRight{Robust Regression with Highly Corrupted Data via Physics Informed Neural Networks} & \centering{\cite{peng2022robust}} & \RaggedRight{Feed-forward PINN based on the least absolute deviation method to reconstruct PDE solutions and parameters from highly corrupt sensor data}  & \RaggedRight{Corrupt data and parameter reconstruction}\\ \hline \\
    
    \RaggedRight{A generic physics-informed neural network-based framework for reliability assessment of multi-state systems} & \centering{\cite{zhou2023generic}} & \RaggedRight{Feed-Forward PINN regularized by deviations from ODE of system state transition and initial conditions. Individual element of the loss parse through projecting conflicting gradients to establish continuous latent function for reliability assessment} & \RaggedRight{Reliability assessment}\\ \hline \\

    \RaggedRight{Physics-guided convolutional neural network (PhyCNN) for data-driven seismic response modeling} & \centering{\cite{zhang2020physics}} & \RaggedRight{Physics-Informed Loss (Dynamic System with Ground Excitation)} & \RaggedRight{Structural Health Monitoring} \\ \hline \\
     
    \RaggedRight{A physics-informed deep learning approach for bearing fault detection} & \centering{\cite{shen2021physics}} & \RaggedRight{Physics-Informed Loss (Deviation from Physics-Based Threshold Model Penalized)} & \RaggedRight{Machinery fault detection and diagnosis in bearings} \\ \hline \\

    \RaggedRight{Physics-guided deep neural network for structural damage identification} & \centering{\cite{huang2022physics}} & \RaggedRight{CNN employed as feature extraction for both the physics and data domain. The network was regularized in accordance to labelled data as well as the objective of minimizing feature discrepancy between domains} & \RaggedRight{Structural health monitoring in bridge structures} \\ \hline \\

    \RaggedRight{Bridge damage identification under the moving vehicle loads based on the method of physics-guided deep neural networks} & \centering{\cite{yin2023bridge}} & \RaggedRight{Physics-informed loss function for feature fusion between the physics-based numerical model and data-driven model)} & \RaggedRight{Structural health monitoring in bridge structures} \\ \hline \\

    \RaggedRight{A physics-informed convolutional neural network with custom loss functions for porosity prediction in laser metal deposition} & \centering{\cite{mcgowan2022physics}} & \RaggedRight{Physics-informed CNN with loss function penalizing deviations from ideal simulated parameters)} & \RaggedRight{Process monitoring in additive manufacturing for porosity buildup)} \\ \hline \\

    \RaggedRight{Physics-Informed Learning for High Impedance Faults Detection} & \centering{\cite{lideka2021physics}} & \RaggedRight{Physics-informed convolutional autoencoder, with physics-informed regularization based on elliptical relation characteristics of voltage and current plots} & \RaggedRight{Fault detection in power grids} \\ \hline \\

    \RaggedRight{Physics-informed deep learning for signal compression and reconstruction of big data in industrial condition monitoring} & \centering{\citep{russell2022physics}} & \RaggedRight{Physics-informed convolutional autoencoder, featuring loss term incorporating auto-correlation and Fast Fourier Transform metrics} & \RaggedRight{Data compression for collected monitoring signatures in machinery fault detection and diagnosis}\\ \hline \\

    \RaggedRight{Physics guided neural network for machining tool wear prediction} & \centering{\cite{wang2020physics}} & \RaggedRight{Cross physics-data fusion for the integration of physical parameters within model input. Physics-informed loss function employed to enforce relationship between tool degradation with respect to operation progress} & \RaggedRight{Condition monitoring for tool wear}\\ \hline \\

    \RaggedRight{A Novel Physics-Informed Framework for Real-Time Adaptive Monitoring of Offshore Structures} & \centering{\cite{liu2023novel}} & \RaggedRight{Employed a physics-informed RNN for solution to governing equations of eigensystem, representative of the modal identification process} & \RaggedRight{Structural health monitoring}\\ \hline \\

    \RaggedRight{Physics-Informed LSTM hyperparameters selection for gearbox fault detection} & \centering{\cite{chen2022hyperparameter}} & \RaggedRight{Maximization of Mahalanobis distance between healthy state and established physics-informed fault state for LSTM optimization process} & \RaggedRight{Machinery fault diagnosis in gear boxes} \\ \hline \\ 

    \hline \hline \\[0.01cm]
    
\end{longtable}
\end{tiny}
\end{singlespacing}

Physics-guided regularizations consist primarily of tailoring constraints that directly alter the data-driven model in the training phase to favor predictions that are consistent with underlying physics. Constraints of this type are also known as learning biases, as characterized by \cite{karniadakis2021physics}, and implemented through physics-informed loss functions. These loss functions penalize deviations from physical laws, making the model more likely to produce physically plausible solutions. Conventionally, the loss function employed in ML algorithms is a measure of the empirical difference between the model prediction and ground truth, with the objective of minimizing the loss function through an iterative process. Model loss is optimized by adjusting the parameters of the model to reduce the aforementioned difference in model predictive capabilities versus ground truth. In contrast, a physics-informed loss function incorporates additional information about the system being modeled, such as physical constraints, conservation laws, and other known properties of the system in tandem with the penalization of deviations from ground-truth observations. Through this framework, the ML algorithm may more effectively constrain the prediction space to avoid violations of physical principles. \\

Algorithms introduced in this format aim to simultaneously minimize errors to both the labeled data and physical constraints. This is reflected in the structure of the loss functions implemented, whereby the physics-informed loss function is comprised of a data-driven loss term and a physics-based loss term. The data-driven loss term measures the error between the predicted output of the model and the ground truth, or observational data. In contrast, the physics-based loss term ensures that the solution satisfies the underlying physics of the problem through adherence to governing equations specific to the problem. Conventionally, compliance with observed data (data-driven loss) is achieved by minimizing the residual between predictions of the network and true state and is performed with a variety of distance evaluators such as the mean squared error (MSE) or cross-entropy error (CSE). Compliance with known physical laws is case specific and varies in implementation, however, the aforementioned methods for evaluation have seen many implementations in literature. The general form of the loss function then, may be represented as:

\begin{equation}
\ Loss_{total} = \lambda_{1}Loss_{empirical}(Y_{prediction},Y_{target}) + \lambda_{2}Loss_{physical}(Y_{prediction})\label{eq:01}
\end{equation}
    
Where the parameter \(\lambda_{1}\) and \(\lambda_{2}\) is the regularization factor to adjust loss terms to best-fit system characteristics. Thus in this format, authors have introduced a methodology for the incorporation of governing equations to influence the direction of loss minimization in networks. In literature, physics-informed regularization has been employed to incorporate knowledge of the expected fault signatures of the system under different failure modes, in an effort to ensure that the model is able to accurately detect and classify faults, even in the presence of noise or other confounding factors. For instance: \citep{sun2021microcrack} proposed a methodology for the non-destructive detection and quantification of micro-crack defects, a framework based on the electromagnetic acoustic transducer, which functions by exciting guided waves for crack detection. The group develops a novel physics-informed architecture that they have termed \emph{GuwNet}. The proposed network seeks to employ various deep learning modules such as convolutional layers, dense layers, and GRU layers in conjunction with the introduction of physical parameters for the approximation of variables of crack propagation. The physical process is represented through various connections within the data-driven and physics-based layers and parameters within the network. The network is optimized by hybrid feed-forward and feedback loss functions, comprised of empirical and physics-informed error terms to integrate the physics of ultrasonic wave testing into the training process of the network. Physics-informed terms are derived from the relationship of defect depth, and quantified by transmitted wave intensity and reflected wave intensity of the ultrasonic guided wave nondestructive testing method. The method demonstrated great promise in the detection of length, depth, and direction of crack propagation, and was shown to have significant improvements in accuracy in comparison to conventional deep learning approaches. \cite{freeman2022physics} proposed a hybrid approach for anomaly and fault detection in turbine rotor blades, whereby fault features acquired from turbine power signals are combined with environmental data to ensure conformity to the dynamics of the hydro-kinematic rotor. The framework extracts statistical features by means of continuous wavelet transforms, and categorized via multi-nomial regression. The time domain features selected were proven by the authors to be physically significant, accurately reflecting the high-frequency fluctuation behavior in signals with respect to turbulence intensity. Turbulence intensity is classified with a neural network, based on time-domain features extracted from the reduced feature space and physically constrained through a hybrid loss function, whereby deviation from the dynamics of turbulence intensity is penalized. \\

Regularization has also been applied with respect to applications in fatigue stress and life monitoring.  \cite{zhang2021physics} constrained the process of creep-fatigue life estimation in a stainless steel specimen with physics-augmented feature engineering and physics-informed regularization. The developed feed-forward model introduces two physics-informed loss terms that take into account and penalize physical violations with regard to fatigue life. From the expected behavior of creep-fatigue in the specimen, the authors added physical constraints in the form of penalization for negative values, as well as extreme values of creep-fatigue life within the loss function. The model constructed boasted superior performance when compared with benchmark empirical and purely data-driven methods. \cite{kim2022data} adopted a data-driven prognostics model that incorporates low-fidelity physical features in the optimization process. The authors presented an innovative methodology for obtaining training parameters for unlabelled extrapolation data. In general, the process for obtaining the extrapolated region, that is, the target of the prognostics framework, involves the physics-based regularization term that penalizes deviation from the low-fidelity physical model. To this effect, the model is optimized to minimize interpolation error with available data, as well as extrapolation error, from the embedded physical model. The authors validated their approach with their verification of fatigue crack growth with respect to Paris's law. \cite{de2023long} built upon conventional frameworks for monitoring the progression of fatigue on off-shore wind turbines by extending the monitoring time period. Conventional evaluation of damage monitoring models is based upon the model's ability in ten-minute fatigue damage estimations, whereas Santos et al. have extended this methodology for monitoring long-term fatigue accumulation. The PINN model proposed focuses on minimizing the Minkowski logarithmic error, providing a more conservative estimation of fatigue damage in the form of damage estimation moments. The loss function was derived such that accuracy between the model's ability to predict short-term and long-term damage is not compromised.\\

\citep{li2022physics} further extended the physics-informed loss function to meta-learning, in their proposed strategy for estimating tool wear. The method integrates both physically derived model inputs, as well as physics-informed loss terms with data-driven models over a series of ML models for the purposes of meta-learning. Meta-learning is defined as the systematic observation and learning of learning from meta-data or the observed experience accrued by ML models and their performance on various tasks. Meta-learning may be classified as a sub-field of machine learning, whereby artificial intelligence models are trained to automatically solve tasks or problems more efficiently and effectively. In their work, the inherent principles of tool wear are learned for applications in tool wear predictions under varying tool wear rates. Through the various parameters derived from the dynamic relationships governing tool wear, the authors derived the input feature space of the various deep learning and machine learning algorithms tested, for enhanced interpretability and robustness. Individual ML models are constructed with the basis of physics-informed data-driven modeling with \emph{cross physics-data fusion}. Initially conceptualized by \cite{wang2020physics}, the model represents a methodology to fuse data from both the physics and data-driven features. The meta-learning model is employed to learn the experiences of three machine learning models and their predictions of the degradation state of the asset at different stages of wear. The algorithms tested were optimized via the physics-informed loss function, whereby constraints to the tool wear rate are imposed based on inherent attributes of tool wear and relations governing tool wear and cutting force.

\subsubsection{Physics-Informed Neural Networks}
\label{PINNs}
Physics-informed neural networks (PINNs) are a rapidly growing field that leverages the power of neural networks to learn complex patterns and relationships from data, while also incorporating the underlying physical principles such as partial differential equations (PDEs) or ordinary differential equations (ODEs) that govern the system. This specific implementation of physics-informed regularization enables the development of predictive models that can not only make accurate predictions but also provide physical insights into the system's behavior. PINNs are referred to as physics-informed in that they incorporate physics-based knowledge or constraints into the model training process, whereby the neural network is employed to make predictions on the solutions space of governing PDEs. Through the introduction of learning biases, PINN significantly relaxes restrictions in terms of the quantity of data required to properly train deep learning algorithms \citep{xu2023practical}. PINNs are known for their ability to generate accurate predictions with small amounts of data, which is especially important in cases where data acquisition is expensive or challenging. Furthermore, PINNs are designed in accordance with the physical laws and constraints of the system and produce predictions that boast superior accuracy and that are physically meaningful. These factors make PINNs particularly well-suited for applications in which the underlying physics of the system is well-understood. \\

The concept of leveraging the computational capabilities of neural networks for solutions to differential equations was initially presented by \cite{lagaris1998artificial}, however, its reach was limited due to limitations of computational power at the time. More recently, \cite{raissi2019physics} popularized the concept through their study, where they demonstrated the effectiveness of PINNs in solving forward and inverse problems pertaining to governing differential equations of a physical system. The effectiveness of PINNs, as defined in the work of \cite{raissi2019physics}, is derived, in part, from their usage of the universal approximation capability of neural networks \citep{hornik1989multilayer}, which states that a neural network with a single-layered feed-forward network with an activation function may approximate any function, provided that it is comprised of a sufficient number of neurons. Naturally, researchers have extended this property to the solution com complex, non-linear differential equations, in which numerical or empirical solutions are difficult or impossible. In these scenarios, PINNs have been leveraged to learn the mapping between the input data and the output variables while enforcing the physical constraints of the system. In addition to their ability to incorporate prior knowledge, PINNs are capable of learning the solution to ODES or PDEs from incomplete data or data with noise, while simultaneously satisfying the governing equations of the system, making them particularly useful for applications in which data is scarce or expensive to collect \citep{raissi2019physics}.  Through this framework, researchers can build accurate models that provide insights into the underlying physical processes, making them a valuable tool in many scientific and engineering applications \citep{raymond2021applying}.\\

The original PINN architecture by \cite{raissi2019physics} is based on the feed-forward structure, and employed to solve the first-order non-linear PDE. Various names exist for this structure in literature such as Feed-Forward Neural Networks, Artificial Neural Networks, Multi-layer Perceptron Neural Networks, and Deep Neural Networks.  The feed-forward neural network is a type of artificial neural network that consists of multiple layers of interconnected nodes, or neurons, that transmit information through weighted connections. In the context of PINNs, the input layer of the network corresponds to the physical domain, while the output layer represents the solution to the problem of interest. The intermediate layers, also known as hidden layers, provide the necessary computational power to map the input to the output. \\
 
An artificial neural network may be described as a series of non-linear transformations. In terms of a mathematical definition of the network: For a given input layer of \(N\) neurons, and may be denoted as \(X = \{x_{1},...,x_{n}]\), whereby \(x_{i}\) represents a feature within the input space \(X\). The network may be defined to host \(H\) hidden layers, with each layer containing \(M\) neurons. From this, the output of the \(I\)-th hidden layer, \(i \in \left[ 1, H\right] \) may be represented as \(A^{I} = \{a^{I}_{1},...,a^{I}_{m}\}\), where \(a^{I}_{j}\) represents the \(j\)-th neuron in the \(I\)-th hidden layer. For each hidden layer, the output \(A^{I}\) is computed through an element-wise applications of non-linear activation function \(\Theta_{i}\) to the weighted sum of inputs from the prior layer \(I-1\), which may be written as:

\begin{equation}
    \ z^{I}_{j} = \sum{\left(w^{I}_{ji} * a^{I-1}_{i}\right) + b^{l}_{j}} \label{eq:02}
\end{equation}

Where \(w_{ji}\) represents the weight connecting the \(i\)-th neuron in the prior layer \(I-1\) to the \(j\)-th neuron in the current layer  \(I\), \(a^{I-1}_{i}\) represent the output of the \(i\)-th neuron in the prior layer, and \(B_{i}\) representing the bias term associated with the \(j\)-th neuron in the  \(I\)-th hidden layer. The output of the \(i\)-th hidden layer is computed as:

\begin{equation}
    \ a^{I}_{j} = \Theta^{I}\left(z^{I}_{j}\right) \label{eq:04}
\end{equation}

The output layer is comprised of \(K\) neurons, with predicted output denoted as as \(Y = \{y_{1},...,y_{k}\)). Thus, the output of the neural network may be computed as:

\begin{equation}
    \ z^{H+1}_{j} = \sum{ (w^{H+1}_{ji} * a^{H}_{i}) + b^{H+1}_{j} }\label{eq:05}
\end{equation}

Where \(w^{H+1}_{ji}\) represents the weight connecting the \(i\)-th neuron in the H-th hidden layer to the \(j\)-th neuron in the output layer, \(a^{H}_{i}\) is the output of the \(i\)-th neuron in the \(H\)-th hidden layer, and \(b^{H+1}_{j}\) is the bias term associated with the \(j\)-th neuron in the output layer. Collectively, this may be referred to as:

\begin{equation} 
    \label{eq:05}
    \ \textbf{z}^{H+1} = \textbf{w}^{H+1} * \textbf{a}^{H} + \textbf{b}^{H+1} \\
\end{equation}

The PINN employs this existing framework to be an approximator of the solution to the PDE. In the general case, the non-linear PDE parameterized by \(\gamma\), as well as its initial and boundary conditions may be represented by the form:

\begin{gather} \label{eq:06}
    \mathcal{F}\left(x, t, u, \nabla u,...; \frac{\delta u}{\delta t}...;\gamma\right) = 0, x \in \Omega, t \in [0,t] \\
    u(x, t=t_{0}) = g(x),  x \in \Omega \\
    u(x, t) = h(x, t), x \in \delta\Omega, t \in [0,t]
\end{gather}

Defined in the domain \(\Omega\), where \(\Omega \in {R}^{d}\) with boundaries \(\delta\Omega\). \(\mathcal{F}\) represents the non-linear function that defines the relationship between unknown function \(u\), its derivatives, and its parameters. The PDE defined has hidden solution \(u\left(x_{1}...x_{n}, t\right)\), with input space that may be composed of spatial variables \(x\) and temporal variables \(t\). For some subsequent literary works discussed, the system in question may be time-independent, therefore, terms in the above equations pertaining to time would not be relevant. The PDE has initial conditions\(g\) and boundary conditions \(h\). The neural network seeks to make a computational approximation of the solution \(u_{NN}\) from input space \citep{raissi2019physics, karandikar2021physics}. The approximation of solution space by the neural network is denoted as:

\begin{equation}
    \ u_{NN}\left(x_{1}...x_{n}, t\right) \approx \textbf{z}^{H+1}\label{eq:07}
\end{equation}

The derivatives of this approximation may then be calculated by automatic differentiation, employing the chain rule of calculus to compute the exact derivatives of a function with respect to its input variables \citep{baydin2018automatic}. Utilizing the predicted solution \(u_{NN}\) and its derivatives, it is possible to then reconstruct the PDE and its initial and boundary conditions. This reconstruction is then evaluated with respect to any labeled data provided, the residual to the differential equation itself, and any boundary or initial condition provided for deviations to any of the aforementioned terms, represented as:

\begin{equation}
\ Loss_{total} = \lambda_{1}Loss_{Data} + \lambda_{2}Loss_{PDE} + \lambda_{3}Loss_{BC} + \lambda_{4}Loss_{IC}\label{eq:01}
\end{equation}

With parameters \(\lambda_{1}, \lambda_{2}, \lambda_{3}, \lambda_{4}\) representing weights for the adjustment of each loss term.  Deviations, typically evaluated as mean squared error (MSE) are minimized during the back-propagation process, whereby neural network parameters, such as weight and biases, are adjusted accordingly in accordance with the governing equations, as represented in \ref{fig:4}. Minimization of the total deviation through the optimization algorithms such as gradient descent allows the network to learn the mapping between the input and output space, while simultaneously complying with known physical laws and constraints.

\begin{figure}[h!]
\begin{center}
\includegraphics[width=\textwidth]{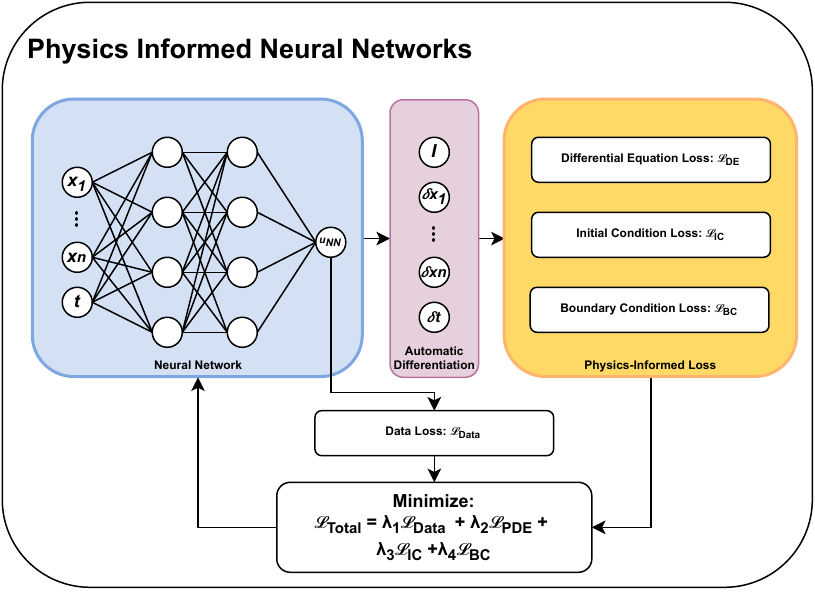}
\end{center}
\caption{Physics-informed Neural Network structure}\label{fig:4}
\end{figure}

In the context of condition monitoring, PINNs allow for accurate predictions by incorporating both data-driven and physics-based approaches. PINNs can handle sparse and noisy data, extrapolate beyond training data \citep{kim2022data}, and provide interpretable results. They also enable early fault detection, reduce false alarms, and can be used for online monitoring. Since their initial popularization by \cite{raissi2019physics}, a plethora of subsequent implementations that followed their publication have employed the same feed-forward architecture. However, experimentation with other popular deep learning architectures, such as the CNN, RNN and its variants, encoder and decoder networks, as well as graph neural networks have been deployed in literature. The following sections will detail the integration of physics-based regularization with a variety of neural network architectures. 

\subsubsection{Data-Driven Solutions to Differential Equations}

\begin{figure}[h!]
\begin{center}
\includegraphics[width=\textwidth]{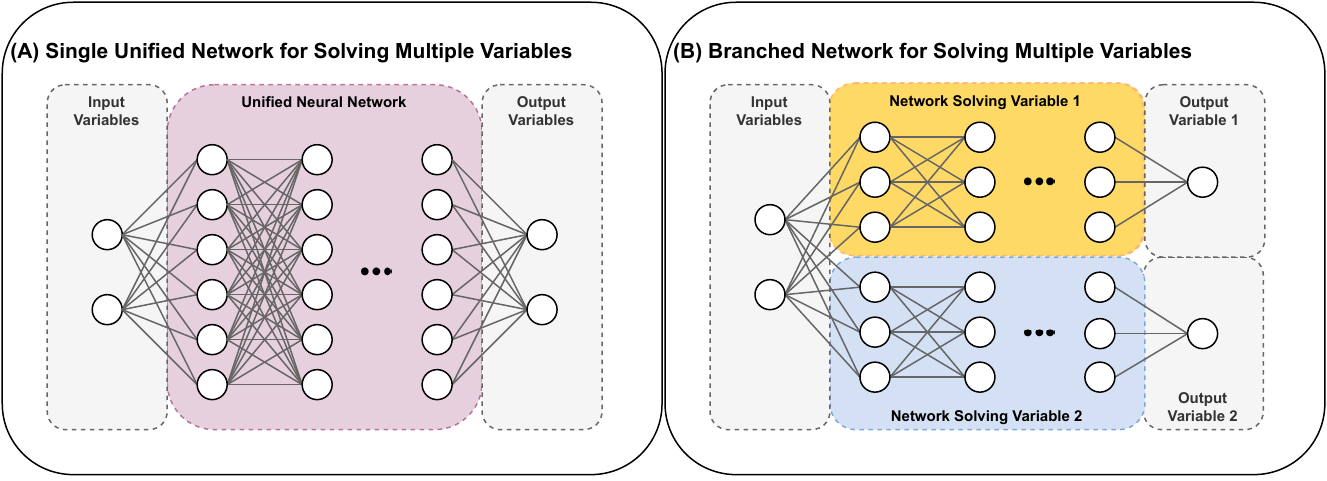}
\end{center}
\caption{Neural network architectures for the solutions of unknown variables (A) for a unified neural network, (B) for independent networks.} \label{U&SNN}
\end{figure}

Various current applications of the PINN framework have remained faithful to the initial PINN architecture, via the solution to governing differential equations of physical systems. Applications of such methods vary greatly across industries, and have been applied to numerous areas in which governing differential equations are known beforehand. For instance, within the domain of solid mechanics, PDEs of physical parameters such as elasticity, deformation, and structural response are determined with the purpose of continued structural health monitoring. One such example is evident in the work of \cite{haghighat2021physics}, who developed a method for surrogate modeling and model inversion with respect to behavior in structures defined by the principles of linear elasticity. This is performed through the incorporation of governing PDEs and various constitutive equations into a PINN for parameter estimations. Through their experimentation, the authors demonstrated the proof of concept through a model of the displacement field under elastic plane-strain conditions. For their use case, the authors compared the effects of a collective network with shared hidden layers \ref{U&SNN} (A), as opposed to utilizing the PINN framework to solve for individual outputs irrespective of the others \ref{U&SNN} (B), with each output being solved by a PINN drawing data from a collective input space. The authors have concluded that, while in principle, a wider network will allow individual associations to be made between sections of the network and output, it was more effective for each variable of the solution to be calculated separately. The authors attributed this to the hyperbolic tangent activation function used, being incapable of accurately representing the cross-dependencies of the network outputs in a manner faithful to kinematic relations.

\cite{anton2021identification} applied the PINN framework for material parameter estimation with inputs in the form of full-field displacement data. With respect to structural health monitoring on existing infrastructure, the estimation of material parameters of structural components may be a method of evaluation of degradation. To that effect, the authors derived the solutions to the momentum balance equation, as well as the constitutive equations for linear-elastic materials with the classic physics-informed neural network architecture. Physical regularization was implemented with respect to the PDE established, as well as labeled data available for boundary conditions and observed deformation. Similarly, \cite{kharazmi2021inferring} estimated the structural parameters of a flexible cylinder structure subjected to vortex-induced vibrations from the hydrodynamic force, with the objective of evaluating structural damage due to fatigue. Utilizing the PINN framework, the authors solved the linear beam-string equation, which governs the motion of the cylindrical structure in question. \cite{bharadwaja2022physics} utilizes a PINN to model and quantify uncertainty in the elastic deformation of heterogeneous solids. More specifically, isotropic linear elastic behavior is assumed to solve the governing differential equation for the approximation of momentum balance and constitutive equations governing elasticity. The proposed PINN is optimized via the physics-based loss function, representing model error to governing differential equation, as well as the Dirichlet, Neumann boundary conditions, the boundary conditions associated with fibers and voids, and initial conditions. From their analysis, the proposed physics-informed methodology returned results that are similar to that of the Monte Carlo finite element simulation model, designated as the benchmark model in this scenario. As another example: \cite{rautela2021simulation} simulated guided waves for monitoring structural health with applications in aerospace applications. The framework revolves around using a PINN to solve governing PDEs associated with wave propagation. In their study, the one-dimensional wave equation with Dirichlet boundary conditions is formulated as the target of the loss function, and predictions by the PINN are continuously optimized by the loss function to more accurately reflect the physical governing PDE. \cite{zhou2023physically} proposed a methodology for fatigue life estimation, physically constrained by a hybrid loss function within a probabilistic PINN framework. Through the feed-forward model, the stress-life relationship is approximated. Physical violations are determined through the evaluation of select collocation points, whereby the ground truths are approximated by the probability distribution out-putted by the feed-forward model. Finally, \cite{mai2023robust} employed the PINN architecture in predicting structural instability in truss structures. The network outlined is a representation of the displacement field of the structure, and analysis of parameters allows for the location of critical points susceptible, given the input load factors. Optimization is performed via the minimization of the physics-informed loss function, which represents, physically, the residual load and stiffness characteristics of the structure. In all, the method yields superior accuracy through the various example validations on several truss structures. \\

With applications to machinery fault detection and classification, \cite{shen2023machine} proposed a novel machine fault classification framework employing a unique PINN framework based on Hamiltonian mechanics, whereby the model is trained to represent the energy conservation of the system in healthy and abnormal states. Hamiltonian systems are those that obey Hamilton's equations of motion, which describe the time evolution of a system's state variables in terms of its energy. Based on the principle of Hamiltonian mechanics, the evolution of a physical system is described via the energy of the system as a function of its position and momentum. This network is termed \emph{Hamiltonian Neural Network} (HNN) and may be considered a class PINNs specifically tailored towards the modeling dynamical systems governed by Hamiltonian equations. This incorporation allows networks to predict the evolution of a system over time \citep{greydanus2019hamiltonian}. In their work, \cite{shen2023machine} applied this concept for the classification of faults in rotating machinery. Estimations of system energy signatures are derived from observed sensor measurements through the HNN. Subsequently, parameters of the HNN are extracted to form the total energy function, which is used as the input features for the classification algorithm based on the conventional RF algorithm.\\
 
An abundance of studies has also been performed in optimizing or complementing the available data from sensors for monitoring applications. Through optimization, the objective of designed systems is to maximize the relevant and informative data for monitoring the system. An example of this optimization process with PINNs may be seen in the work of \cite{zhu2022physics}, who optimized sensor placement locations for the monitoring of low-rise buildings in response to wind pressure. The ML model is trained on data generated from a physical simulation by means of a high-fidelity finite element computational fluid dynamics model. From the data provided, the ML model seeks to construct a surrogate model of pressure-field in real time. This surrogate model is further embedded within a neural network for the optimization of sensor placement locations. For inference of non-observable sensor data, \cite{jadhav2022physics} performed condition monitoring of fouling conditions on system health with respect to an air pre-heating system in thermal power plants. Issues arising from the lack of available sensors on the interior of the system are resolved with the proposed PINN architecture based on the non-dimensionalized governing equations for heat transfer for fluid and metal interfaces. The authors employed a series of multiple PINNs in parallel, operating from the same set of input features to resolve a plethora of equations governing heat transfer. PINNs are regularized via the physics-informed loss function, composed of the loss components of the governing equations, boundary conditions, and interface conditions. From the various applications listed, the accuracy of sensor data is critical for the collection of data faithful to the system. Decisions based on inaccurate or incomplete information may lead to sub-optimal outcomes or catastrophic consequences, and as such, one direction of this architecture has been the reconstruction of corrupt sensory data to allow users a holistic view of system operations. In particular, in the work by \cite{peng2022robust}, the authors proposed a PINN structure to reconstruct data with significant corruption from sensor errors. The networks proposed are based upon the Least Absolute Deviation and median absolute deviation, whereby the PINN architecture is continuously optimized by minimizing the residual between data-driven and physical models. The design of the architecture was validated on several classical problems involving PDEs, such as the Navier-Stokes equation, Poisson's equation, and wave equations, whereby the algorithm was capable of accurately recovering governing equations from corrupted observation data. \\

In other avenues of research, PINNs have been applied for the modeling of dynamic systems, as demonstrated in the work of \cite{zhou2023generic}. The authors applied the PINN framework for the evaluation of reliability in multi-state systems. Given that the governing equations for Markov processes take the form of differential equations, the computational efficiency of PINNs is leveraged. The group utilized the gradient surgery method for multi-task learning as outlined by \cite{yu2020gradient} to improve the PINN's precision in approximating solutions to differential equations by alleviating issues with imbalanced gradients during training phases. For multi-state system reliability evaluation, the PINN solves for the state estimates of systems with the input of time instant. As with the traditional PINN, the network is penalized based on loss with respect to boundary conditions, and with respect to approximation of governing equations. In addition to the  PINN architecture based on ANN, various works have incorporated alternate deep learning architectures to best optimize the network for the data structure of particular applications, which will be detailed below.

\subsubsection{Physics-Informed Regularization in Tandem with Other Deep Learning Architectures}
A plethora of literary works employs the inherent symmetries and invariances encoded by various conventional deep-learning architectures in compliance with the philosophy of physics-guided regularizations. Literary works presented in this section mainly utilize physics-informed regularizations as the primary methodology to encode physical knowledge into the system. Leveraging the unique computational efficacy and efficiency of certain architectures for specific data types, researchers have drastically innovated upon the structure of the original PINN and employed the framework in their own fields of specialization. \\

For instance, with respect to the CNN architecture, their unique convolutional layers are valued for their capabilities in automatically extracting features without the need for manual feature engineering, making them invaluable in complex applications whereby the relevant features are difficult to understand or quantify. Studies employing the CNN architecture can be seen in the works of \cite{mcgowan2022physics}, who monitored the porosity during the additive manufacturing process with their introduction of a set of loss functions. The regularization of the network comprises standard cross-entropy data loss, as well as losses informed by physical parameters that penalize deviations from ideal simulated melt pool temperature and length-to-width ratio and relative error prior to normalization. As another example: \cite{zhang2020physics} established a surrogate model for the estimation of structural seismic response, informed via equations of motion representing a dynamic system subjected to ground excitation. \\

Several instances of literature attempt to employ the physics-informed loss function as a methodology to minimize deviations between established physical and data-driven domains. For example: \cite{shen2021physics} adopted a hybrid approach in their development of a physics-informed CNN model for fault detection in bearings under varying rotational speeds. The proposed CNN model and the physics-based threshold model operated co-currently to evaluate the health class of bearings. The threshold model is established based on known limits with regard to the amplitude of envelope spectra of healthy and damaged bearings. Subsequently, a customized physics-informed loss function is implemented, which serves to penalize the model for predictions that deviates from known physics, as represented by the threshold model.  Through this format, however, the authors have made the simplifying assumption that predictions of physics-based models are correct, or rather the probability of predictions being correct is very high, due to the extreme thresholds set. \cite{huang2022physics} explored a similar approach for the combination of the physical and data domains. The authors trained a CNN employing a finite element model for applications in structural health monitoring. Through their designed framework, the authors sought to incorporate predictions from both the physics-based finite element model and data-driven methods. The CNN proposed functions as a set of feature extractors that operates simultaneously based on inputs from the finite element model-driven physics domain, and the data domain. Physical constraints are encoded in a classifier through a novel cross-physics-data domain loss function, whereby predictions of the classifier are evaluated with respect to the labeled data, as well as the discrepancy of features between the physical domain and the data domain. On a similar note, \cite{yin2023bridge} monitored structural damage localization in bridge structures due to loads applied by vehicles. The authors developed a numerical simulation of the structure and, using the physics-informed loss function sought to fuse features from the physics and data domains. Processed data from both domains are fed through the Visual Geometry Group 16 architecture \citep{simonyan2014very}, whereby damage features are extracted from the time-frequency map of acceleration signals. The optimization was carried out with a hybrid loss function comprised of data-driven cross-entropy loss and physics-informed loss penalizing deviations from the physical domain established via numerical simulations. Effectively, the network seeks to minimize discrepancies between the physical and numerical models. 

\begin{figure}[h!]
\begin{center}
\includegraphics[width=\textwidth]{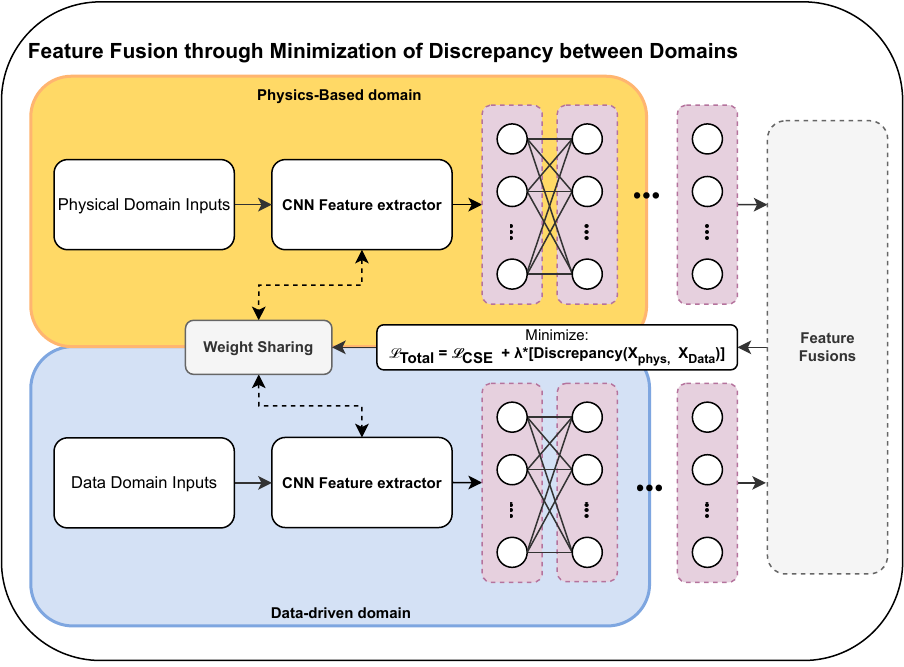}
\end{center}
\caption{Integration of physics-based and data-driven domains through feature fusion: The CNN architecture is employed as a feature extractor. Adapted from \cite{huang2022physics} and \cite{yin2023bridge}} \label{fig:6}
\end{figure}

Another implementation of physics-informed regularization is with structures involving the encoder-decoder style networks, or autoencoders. The structure of networks of this style may be described as two components working in tandem: an encoder and a decoder network. Through the encoder network, input data is compressed through multiple transformations to a low-dimensional representation. This representation is subsequently decompressed and transformed back into the original representation through various transforms in the decoder, with the objective of accurate reconstruction of input data. Intermediate layers typically consist of lower quantities of neurons, which in effect force the network to learn a compressed representation. In general, AEs are particularly well-suited for condition monitoring tasks as they are able to learn the representations of the normal operating state of a system and detect anomalies or deviations from that state \cite{zhou2017anomaly}. Implementations of the autoencoder learn to identify these changes by encoding the normal behavior of the system into a lower-dimensional representation, and then detecting anomalies in the reconstruction error when the system deviates from this normal behavior.  This strategy has been employed in subsequent literary works for the effective detection of deviant behavior without the need for additional labeled data. For example; \cite{lideka2021physics} designed a physics-informed convolutional autoencoder for the detection of high impedance faults in power distribution grids to overcome the issue of the lack of labeled data from conventional approaches. The physics hybrid physics-informed loss term featured in the network serves to regularize the prediction of voltage, taking advantage of the physical relationship, the elliptical trajectory between measured voltage and current. As another example: \cite{russell2022physics} proposed a framework for signal compression and reconstruction of large quantities of data in the setting of industrial condition monitoring through a physics-informed deep convolutional autoencoder. A hybrid loss function was developed comprised of the traditional MSE, Pearson's correlation coefficient loss, and a physics-informed loss term. As the primary objective of an autoencoder is to reconstruct a given signal, dominant frequencies in the signals must be preserved post-reconstruction. This fact is leveraged by the authors to impose a physical constraint on the data-driven solution through a loss term sensitive to frequency. The authors also selected to learn latent representations of operating conditions individually, effectively isolating the compressed representations, with the objective of optimal representation for individual faults.\\

\begin{figure}[h!]
\begin{center}
\includegraphics[width=\textwidth]{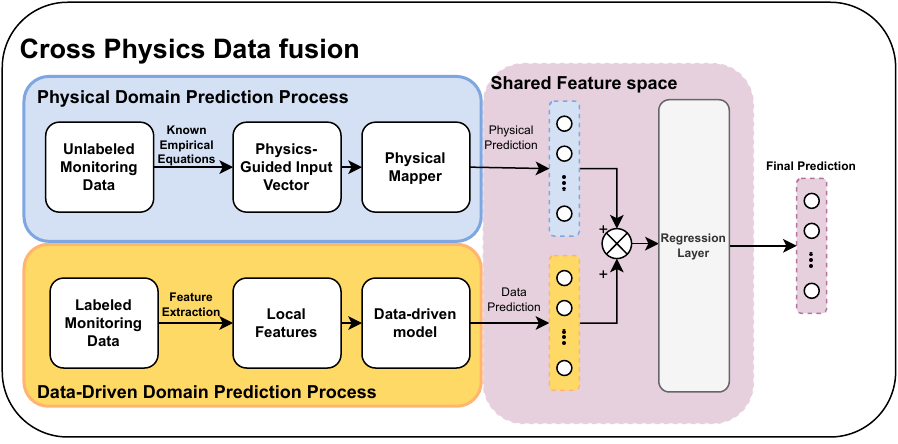}
\end{center}
\caption{Cross Data-Physics Fusion, as presented by \cite{wang2020physics} predictions based on information from both the data domain (comprised of features derived from labeled monitoring data), and physics domain (comprised of features derived from unlabeled data) are simultaneously mapped to a shared space, and concatenated. Both are processed through a regression layer for the final prediction.} \label{CDF}
\end{figure}

Several examples in literature also take advantage of the RNNs' ability to extract temporally invariant data, for use in applications involving time-domain monitoring. For example, \cite{wang2020physics} fused features from the data-driven and physics domain through their applications of the \emph{cross physics-data fusion}, with application in modeling damage accumulation in tools. Features from the data domain and physics domain are extracted separately, and subsequently mapped to a shared feature space, representing tool wear. Predictions from both domains are concatenated, and evaluated in the final regression layer of the network whereby a physics-informed loss function is employed to minimize discrepancies between the data-driven Bi-directional Gated Recurrent Unit model and empirical equations. \cite{liu2023novel} proposed a physics-informed RNN for offshore structural monitoring. The methodology proposed employs an optimal singular value decomposition procedure for modal identification of the structure. Through their study, the authors formulated the physics-informed modal identification process into an eigensystem and employed an RNN for the solution of the governing differential equations of the eigensystem.  Through their proposed framework, the authors improved upon conventional monitoring methods to devise an efficient strategy for modal identification and monitoring in real-time, and under dynamic environmental conditions.\\

Researchers have also innovated upon the methodology by which the loss is evaluated. Traditionally, the vast majority of literature explores the minimization of deviations from a target value. \cite{chen2022hyperparameter} instead proposed an LSTM differentiation strategy for the state of health focusing on maximizing deviations between known states. In their developed strategy for the selection of LSTM hyperparameters in the detection of gearbox faults, rather than the conventional minimization of mean squared error of the labeled data, the selection strategy proposed maximizes the discrepancy, in this case, evaluated by the Mahalanobis distance, between healthy and physics-informed faulty states. Data of vibration signatures correlating to the fault state are generated based on prior knowledge of the system and used to establish the target of evaluation.\\

In all, physics-informed regularization techniques represent a powerful tool for the introduction of constraints within the training process of deep learning networks. Unlike the previously detailed models, physics-informed regularization presents a guided process by which the algorithm is able to acclimate to the domain of physical feasibility, as illustrated in the numerous works discussed in this particular section. Though effective, the main limitations of this approach are primarily regarding the increased complexity of the loss landscape, and difficulties in achieving generalization. Various authors have devised methodologies to circumvent this issue, with several further exploring the idea of physical constraints to network optimization, through various alterations to the architecture itself, as will be discussed in the following section.

\subsection{Physics-Guided Design of Architectures} \label{34}
In addition to the loss function, the architecture of the ML algorithm itself can be designed to incorporate physics-based constraints. From the literature, this area of development primarily focuses on the design of appropriate neural network architectures that can efficiently encode biases and learn the underlying physics of a system. A number of specialized neural network architectures have been proposed to tackle the unique challenges in engineering applications.\\

One such approach is to leverage the information available to encode some physical meaning to hidden values within the black-box structure. Particularly with deep learning architectures, physical meaning may be assigned to intermediary nodes or outputs to facilitate physically-guided and interpretable information flow throughout the network. Depending on the application, through specialized operations and/or transformations of data retained in intermediary nodes in the form of network layers and connections, the physical relevance of the node may be propagated.  Another approach commonly employed is to ascribe physical significance to the connections between nodes. Through this node connection, a fixed physical operation or transformation may be specified between layers of the network, also accomplishing the task of the preservation of physical principles within information flow, albeit with a different methodology. \\

Subsequent subsections will detail some applications of the aforementioned architecture design, with respect to a selection of popular deep learning frameworks. In addition, this section will detail the workings of conventional deep learning architectures within the frame of physics-informed architecture design, with details regarding their architecture and their suitability for specific applications pertaining to data type and physics encoded.

\paragraph{Feed-Forward Neural Networks}
Various examples of this adjustment to architecture exist in literature. As the feed-forward structure has already been discussed in section \ref{PINNs}: Physics-Informed Neural Networks, this section will not feature the description of the network itself. Despite recent innovations in architectures, feed-forward neural networks are still commonly employed for their simplicity, relatively efficient computation, and capabilities for universal approximation of continuous functions. Their structure itself makes feed-forward networks comparatively easier to analyze, and subsequently encode physical relevance to sections of the network. As such many authors have taken to the development of interpretable and physics-informed architectures based on the feed-forward structure. Table \ref{table:4} provides a brief summary of literary works compiled for the embedding of physics within the feed-forward architecture:

\begin{singlespacing}
\begin{tiny}
\begin{longtable}[h!]{p{0.25\textwidth}  p{0.11\textwidth} p{0.25\textwidth} p{0.25\textwidth}} \hline \hline \\
    \\
    \textbf{Article Title} & \textbf{Citation} & \textbf{Description} & \textbf{Application} \\ 
    \hline \hline \\[0.05cm]
    \endhead

    \hline
    \endfoot

    \caption{A summary of literature compiled for the design of physics-informed architecture, with innovations to the feed-forward neural network architecture primarily.}
    \label{table:4}
    \endlastfoot

    \RaggedRight{Probabilistic physics-guided machine learning for fatigue data analysis} &  \centering{\cite{chen2021probabilistic}} & \RaggedRight{Probabilistic feed-forward neural network with physically constrained weights and or bias optimization to model fatigue life curve} & \RaggedRight{Condition monitoring and fatigue life estimation} \\ \hline \\

    \RaggedRight{Integration of a novel knowledge-guided loss function with an architecturally explainable network for machine degradation modeling} & \centering{\cite{yan2022integration}} & \RaggedRight{feed-forward network with physically interpretable layers based on signal processing techniques, optimized via knowledge-guided loss function} & \RaggedRight{Machine condition monitoring for bearings}\\ \hline \\

    \RaggedRight{Fully interpretable neural network for locating resonance frequency bands for machine condition monitoring} &  \centering{\cite{wang2022fully}} & \RaggedRight{Extreme learning machines, with physically interpretable signal processing algorithms and physical feature extraction encoded as additional layers in the network} & \RaggedRight{Machine condition monitoring} \\ \hline \\
    
    \RaggedRight{A physics-informed neural network approach to fatigue life prediction using small quantity of samples} &  \centering{\cite{chen2023physics}} & \RaggedRight{Feed-forward network, with physical meaning ascribed to certain nodes, enforced by physics-based activation functions based on the  Walker mean stress model and Basquin relation model} & \RaggedRight{Fatigue Life Estimation} \\ \hline \\ 

    \hline \hline \\[0.01cm]

\end{longtable}
\end{tiny}
\end{singlespacing}

 Much of the literature developed in this section sought to provide interpretability and explainability to the neural network model by imposing physical constraints on the feed-forward and back-propagation process of the neural network itself. One such example of assigning physical parameters as nodes to enforce information flow consistent with underlying physics may be found in the work of \cite{chen2021probabilistic}, who proposed a probabilistic approach, whereby a feed-forward model is employed to learn the mean and standard deviations for stress to fatigue life distribution relation. Prior knowledge is imposed through a constrained optimization process, whereby physical parameters such as the fatigue stress applied, fatigue life, and an index indicating if the sample failed or is sustained through the trial are assigned as input nodes. Output nodes involve parameters to define the probabilistic distribution of fatigue life, with mean and standard deviation. The network is constrained via its weights and/or bias restrictions based on known physical relations between parameters, enforcing the intermediary values to be consistent in terms of the sign. As another example \cite{yan2022integration} employed physics-based signal processing techniques in conjunction with physics-informed regularization for a fully architecturally interpretable neural network. The resultant feed-forward neural network developed was designed with three hidden layers, representative of a data-driven formulation of signal processing techniques such as the Hilbert transform, squared envelope, and Fourier transform respectively. The network was regularized by a hybrid loss function, whereby desired characteristics of the health indicator constructed, such as the sensitivity of early fault detection, are optimized. The authors applied this framework to directly construct health indicators from vibrational signals for applications in degradation modeling in machines. 

\begin{figure}[h!]
\begin{center}
\includegraphics[width=\textwidth]{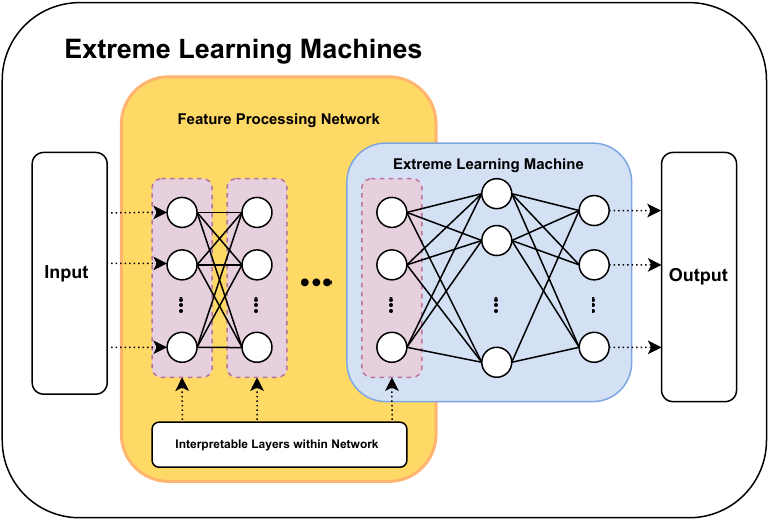}
\end{center}
\caption{Incorporation of physically interpretable feature extraction for use in conjunction with the Extreme Learning Machine: adapted from \cite{wang2022fully}} \label{fig:5}
\end{figure}

 Similar to the above work, \cite{wang2022fully} developed an interpretable framework through the assignment of appropriate physical meanings to layers within the network. The authors applied their proposed extreme learning machine framework for applications in machine health monitoring. Extreme learning machine may be defined as a subset of conventional neural networks that emphasizes the use of simple models to enable efficient and scalable learning. Initially introduced by \cite{huang2006extreme}, rather than the multiple hidden layers of conventional neural networks, an extreme learning machine framework is typically composed of a single hidden layer that maps inputs to outputs based on a set of fixed weights. These models are much easier to train and require much less data and computational resources than standard neural nets. To compensate for the simplicity of the models, extreme learning machines emphasize the use of advanced techniques for feature extraction, data pre-processing, and data fusion to enable the models to learn complex patterns in the data. Such is the case in this study, whereby the authors employed additional feed-forward layers for the purposes of applying the wavelet transform, square envelope and Fourier transform to the sampled input features as illustrated in \ref{fig:5}, similar to the work of \cite{yan2022integration}. Traditionally, hidden nodes within the extreme learning machines are randomly initialized, with random input weights and random biases. Due to this structure, extreme learning machine models only require the accurate learning of the output layer, thereby directly bypassing much of the time and computational required in comparison to a traditional back-propagation optimization approach. \cite{wang2022fully} further innovated upon this structure by introducing specific sparsity measures as a replacement for the randomly initialized hidden layers, greatly increasing the interpretability of the network. Novel transformations and indices of evaluation employed by the authors include the Gini index, kurtosis, smoothness index, and negative entropy.

\begin{figure}[h!]
\begin{center}
\includegraphics[width=\textwidth]{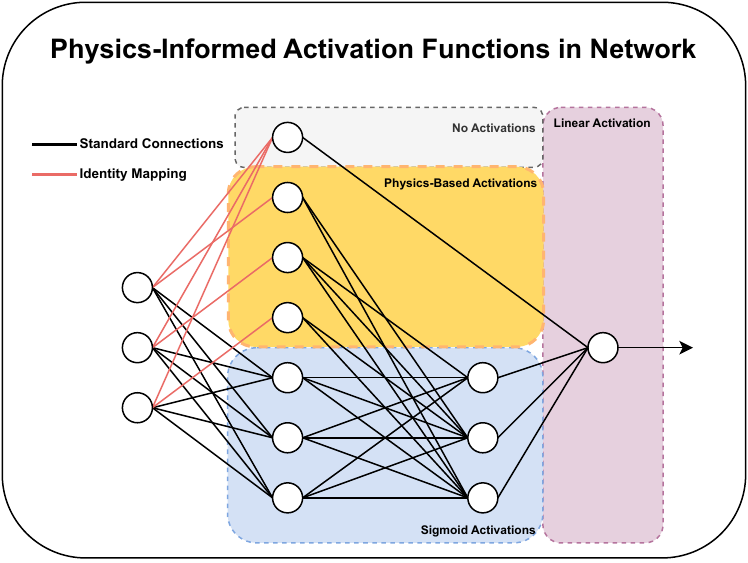}
\end{center}
\caption{Integration of physics-based and conventional sigmoid activation functions in neural networks: adapted from \cite{chen2023physics}} \label{fig:6}
\end{figure}

In contrast to the above works, \cite{chen2023physics} proposed an alternate approach in the integration of physics through architecture, with applications in fatigue life estimation. The author employed a multi-fidelity model, whereby physics governing fatigue life is embedded in the system through a combination of data-driven and novel physics-informed neurons. Interestingly, the authors chose to apply physics-based activation functions to certain nodes within the model, based on purely physical models such as the Walker mean stress model and Basquin relation model. The resultant model structure features certain physical neurons operating in conjunction with data-driven neurons, as illustrated in \ref{fig:6}. This, in effect, enforce the physical relevance of the node itself via its relations with other nodes in the network. 

Due to their simplicity, there exists a wide variety of research available for the application of this particular architecture. Feed-forward neural networks have been employed to great effect in a variety of novel alterations, as seen in the works discussed above. In the past years however, numerous research in this area have improved upon the base neural network structure to be more suitable and specialized for the specific data types and structures, which will be detailed in the following sections.

\subsubsection{Convolutional Neural Networks}
In addition to direct feed-forward models, CNNs have also enjoyed great popularity in the research community. Through their innate architecture, CNNs have the ability to encode certain invariances or symmetries that are inherent in the data they are trained on, making them useful in encoding certain biases based on prior knowledge. By design, CNNs innately take into account spatial invariance through their use of convolutional layers and pooling layers. The unique convolutional layers of the CNN offer an effective and automatic methodology for the extraction of physical meaning from data. These layers serve to extract spatial features from input data, and are employed in tandem with physics-informed layers, which impose physical constraints on the predictions. \\

More specifically, regarding the working of the convolutional layer: Within the convolutional layer, the network applies a set of filters to the input data, with each filter detecting a particular feature or pattern in the input, thereby allowing the network to detect local patterns in different regions of the input regardless of their location in data. In each convolutional layer, the filter is convolved across the entire range of input data in accordance with stride size. The output of this action is known as feature maps, tensors of locally weighted sum. For a typical 2-D convolution operation, the action may be given mathematically as:

\begin{equation}
    \ S\left(i,j\right) = \left(I*K\right)\left(i,j\right)=\sum_{m}{\sum_{n}{I\left(m,n\right) K\left(i-m, k-n\right)}} \label{eq:04}
\end{equation}

Whereby the input data \(I\) is convolved with filter kernel \(K\). From this convolution action, the CNN is capable of accounting for local connectivity, allowing for the capability to detect features invariant of location \cite{lecun1998gradient, lecun2015deep}. A nonlinear activation function is typically applied after convolutions to introduce non-linearities to the system. Pooling layers are generally inserted between convolutional layers to reduce dimensionality while maintaining descriptions of features. In the pooling layer, a sub-sample of each region in the resultant feature map is taken, Instead of the precise feature locations outputted by the convolutional layer, subsequent operations are performed on the summarized features from the pooling layer, allowing for the network to be more robust to variations in feature locations. Pooling layers also help to introduce spatial invariance into the network by reducing the spatial resolution of the input, typically by taking the maximum or average value in each local region. This has the effect of making the network more robust to small variations in the input, such as translations or distortions. Other invariances that may be represented may be rotational, scale, or permutation invariances, depending on the application. This property makes CNN an important asset in condition monitoring tasks where fault signatures may vary. A summary of compiled studies using the physics-informed CNN framework is presented in Table \ref{table:5}.

\begin{singlespacing}
\begin{scriptsize}
\begin{longtable}[h!]{p{0.25\textwidth}  p{0.11\textwidth} p{0.25\textwidth} p{0.25\textwidth}} \hline \hline \\
    
    \textbf{Article Title} & \textbf{Citation} & \textbf{Description} & \textbf{Application} \\ 
    \hline \hline \\[0.05cm]
    \endhead

    \hline
    \endfoot

    \caption{A summary of literature compiled for the design of physics-informed architecture, with innovations to the convolutional neural network architecture primarily.}
    \label{table:5}
    \endlastfoot

    \RaggedRight{Physics-based convolutional neural network for fault diagnosis of rolling element bearings} & \centering{\cite{sadoughi2019physics}} & \RaggedRight{Spectral kurtosis and envelope analysis embedded within layers of CNN for informed feature extraction} & \RaggedRight{Machinery fault detection and diagnosis in bearings} \\ \hline \\

    \RaggedRight{WaveletKernelNet: An interpretable deep neural network for industrial intelligent diagnosis} & \centering{\cite{li2021waveletkernelnet}} & \RaggedRight{Continuous wavelet convolutional layer as the initial layer for effective extraction of bearing fault features} & \RaggedRight{Machinery fault detection and diagnosis in bearings}\\ \hline \\

    \RaggedRight{A health-adaptive time-scale representation (HTSR) embedded convolutional neural network for gearbox fault diagnostics} & \centering{\cite{kim2022health}} &  \RaggedRight{Input signals mapped to health adaptive time scale representation as initial feature map of CNN} & \RaggedRight{Machinery fault detection and diagnosis in gearboxes}\\ \hline \\

    \RaggedRight{Fault Diagnosis of Rolling Element Bearings on Low-Cost and Scalable IIoT Platform} & \centering{\cite{lu2019fault}} & \RaggedRight{Physics-based feature weighting based on fault characteristic frequencies for evaluation of fault information carried by features} & \RaggedRight{Machinery fault diagnosis in bearings} \\ \hline \\

    \RaggedRight{A physics-informed feature weighting method for bearing fault diagnostics} & \centering{\cite{lu2023physics}} & \RaggedRight{Feature weighing layer for evaluation of discrepancy between monitored signals and physics of fault, for construction of input feature map of CNN classifier} & \RaggedRight{Machinery fault diagnosis in bearings} \\ \hline \\

    \RaggedRight{Fleet-based early fault detection of wind turbine gearboxes using physics-informed deep learning based on cyclic spectral coherence} & \centering{\cite{perez2023fleet}} & \RaggedRight{Spectral coherence map established based on vibration signals. Convolutional autoencoder employed for fault detection based on spectral coherence maps of fault conditions} & \RaggedRight{Machinery fault detection and diagnosis in gearboxes}\\ \hline \\ 
 
    \RaggedRight{Physics-informed lightweight Temporal Convolution Networks for fault prognostics associated to bearing stiffness degradation} & \centering{\cite{deng2022physics}} & \RaggedRight {Developed temporal CNN based on the relationship between stiffness and vibration amplitudes to construct physics-informed health indicator} & \RaggedRight{State of health monitoring for bearing stiffness} \\ \hline \\ 

    \RaggedRight{Traffic-induced bridge displacement reconstruction using a physics-informed convolutional neural network} & \centering{\cite{ni2022traffic}} & \RaggedRight{Branched network design based on separate analysis from acceleration-based and strain-based methods, optimized via physics-informed-loss function} &  \RaggedRight{Prediction of displacement in infrastructure for structural health monitoring } \\ \hline \\
    
    \RaggedRight{On-line chatter detection in milling with hybrid machine learning and physics-based model} & \centering{\cite{rahimi2021line}} & \RaggedRight{Energy-based chatter detection model, supplemented by data-driven estimation of the operational state of machine} & \RaggedRight{anomaly detection during process monitoring for milling} \\ \hline \\
    
    \hline \hline \\[0.01cm]
    
\end{longtable}
\end{scriptsize}
\end{singlespacing}

The use of specially designed layers or architectures enables the networks to capture the underlying physics while still leveraging the power of deep learning to make accurate predictions. A common approach employed in current literature is to incorporate physics-inspired layers such as Fourier features, tailored to the physical problem being addressed, with the overall architecture of the CNN itself \citep{jing2017convolutional}. The fundamental concept behind the network design is to integrate physics-based techniques such as signal processing within the network layers, allowing for the visualization of fault features related to physics, and providing a physical perspective on the impact of physics-based, interpretable features in the decision-making process. In many such studies, \citep{lu2019fault, li2021waveletkernelnet, kim2022health, lu2023physics}, layers within a physics-informed CNN can be specifically designed to prompt the network to extract features that are related to the specific fault types of interest. These layers produce a physically relevant feature, map which may then be propagated through various abstractions within the CNN architecture. Through this constraint, subsequent layers are more capable of focusing on more complex feature extraction and classification, improving the accuracy and robustness of the monitoring system, as demonstrated in the works of \cite{wang2022attention} and \cite{li2019understanding}.\\

\begin{figure}[h!]
\begin{center}
\includegraphics[width=\textwidth]{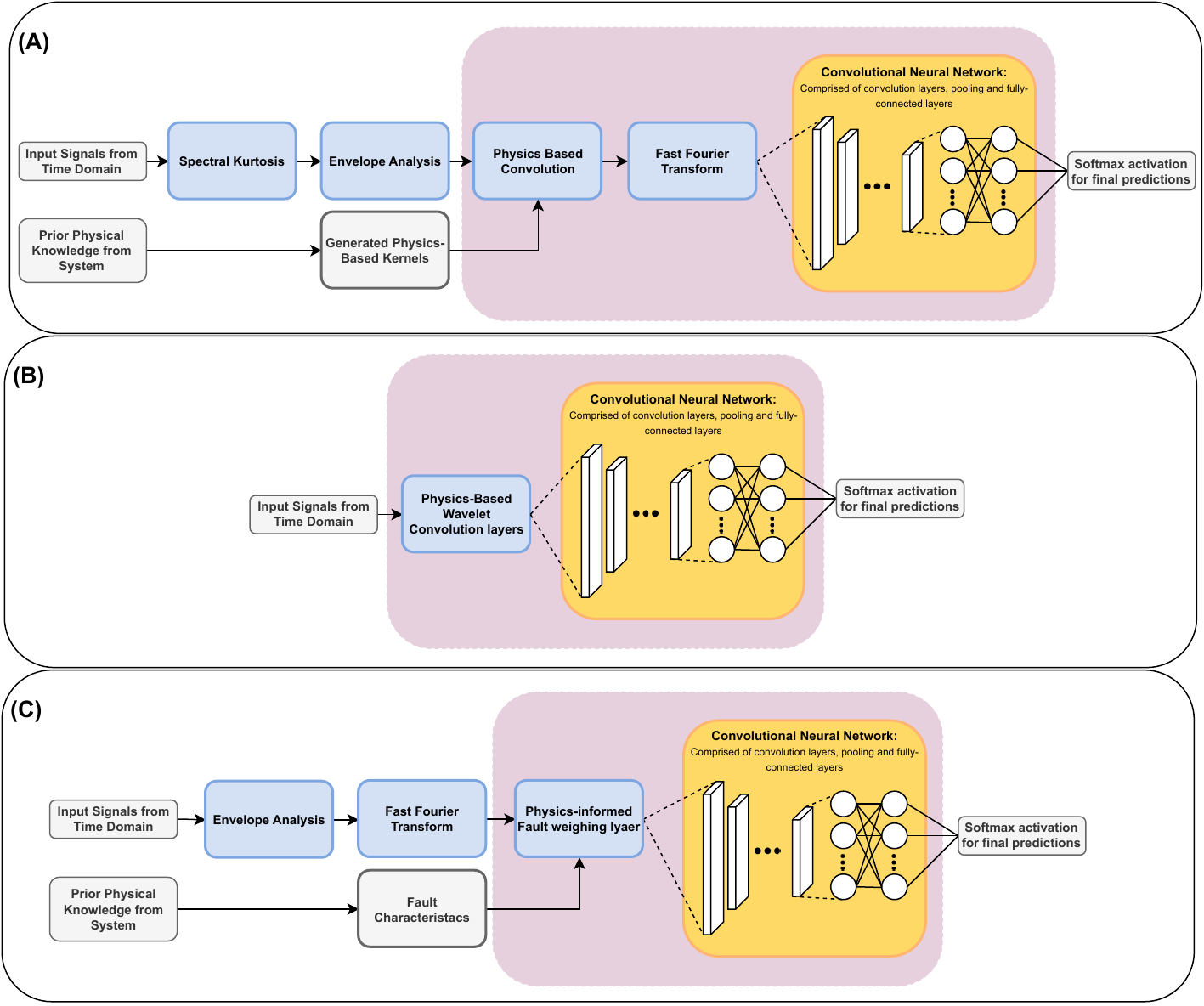}
\end{center}
\caption{Design of Physics-Informed layers for CNN networks, including example architectures adapted from: (A) \cite{sadoughi2019physics} who employed a physics-based kernel generation scheme to generate convolved filters for physics-informed convolutions, (B)\cite{li2021waveletkernelnet} utilizing a convolutional layer to process Continuous Wavelet Transform. (C) \cite{lu2023physics} employing a physics-informed feature selection layer.} \label{PICNN}
\end{figure}

Physics-informed CNN architectures have seen prominent use in analyzing time-frequency type data due to their inherent structure and the symmetries and invariances encoded. In many such applications, the metric by which the state of the system is evaluated is often the vibrations of the asset in operation. Deviations from the standard operation may be determined based on the evaluation of processed operational vibration signals through either one-dimensional CNN for vibration signals or two-dimensional CNN for images mapped in the time-frequency domain. Authors such as \cite{sadoughi2019physics} have also taken to representing physical processes within the CNN through modifications to convolutional filters, or kernels. In their work, a physics-informed CNN framework is established for the diagnostics of faults in rolling element bearings. To process signals from the frequency domain, the authors modified the conventional CNN classification scheme, whereby additional processes are included to enhance fault features. Additional layers consist of a spectral kurtosis layer, an envelope analysis layer for pre-processing information, as well as a Fast Fourier Transform layer for the post-processing transformation of the predicted feature map to the frequency domain. For the network itself, the kernels convolved are generated with reference to the shaft rotation speed and characteristic frequencies of the bearing. The architecture may be seen in figure \ref{PICNN} (A). The authors noted the efficacy of this approach, which may be attributed to its non-reliance on hyper-parameters due to the physics-based nature of kernels. Through this methodology, the authors have shown that the framework is capable of constraining the faults consistently with higher accuracy than conventional deep learning approaches. Further examples involving the use of signal processing techniques embedded within layers of the network are apparent in the work of \cite{li2021waveletkernelnet}, who introduced a novel physics-informed CNN architecture that they have termed WaveletKernelNet, as illustrated in figure \ref{PICNN} (B). The authors presented modification to the conventional CNN architecture through a novel continuous wavelet convolutional layer, allowing the network to more effectively extract impulses embedded in vibrational signals representing bearing faults. Similarly, A similar approach was taken by \cite{kim2022health}, who developed a health-adaptive time-scale representation model, physically informed by characteristic time and frequency domain fault signatures, and embedded within a CNN for analysis of time-frequency images. The authors adapted the physics-informed CNN framework introduced for the monitoring of gearbox faults from vibrational signals with a similar structure as specified in the above work, employing a health-adaptive time-scale representative module for the generation of indicators.

As an extension of \cite{sadoughi2019physics}'s work, \cite{lu2019fault} constructed a physics-informed CNN based on their proposed physics-based feature weighting mechanism, whereby prior knowledge regarding characteristic fault frequencies are employed in weighing vibrational features of rolling element bearings, as seen in figure \ref{PICNN} (C). Inspired by the above works, in \cite{lu2023physics}, the authors further built upon their initial model with the introduction of a physics-informed CNN framework, whereby prior to classification with the CNN, the features are pre-processed in accordance to an initial feature weighing layer and signal processing layers. The proposed layers function to assign greater importance to features with minimal discrepancy to the bearing fault characteristic frequencies. In comparison to \cite{sadoughi2019physics}'s work, \cite{lu2023physics} has elected to directly operate in the frequency domain when constructing the input space of the CNN classifier, with notably lower requirements in terms of computational complexity and similar accuracy. \cite{perez2023fleet} presented an alternate method for vibrational signal processing based on cyclo-stationary analysis. Physical information from vibration signals obtained via a 2-dimensional cyclic spectral coherence map is incorporated with ML for anomaly detection. Through the cyclic spectral coherence maps, physical insights are indirectly integrated through the assumption of the vibration model. A convolutional autoencoder is leveraged for its ability to process spatial data, and employed to reconstruct cyclic spectral coherence maps based on machine data collected in the healthy state. Evaluation of anomalies is performed on physical components subject to rotary motion, with the evaluation creation being the motion producing or exacerbating the cyclo-stationary signal if deviating from nominal operation behaviors. Another implementation of physically relevant layers is demonstrated in the work of \cite{deng2022physics}, who proposed a series of physics-informed temporal CNN for the estimation of bearing stiffness degradation. The authors have presented several frameworks implementing the CNN with physics-informed integration, as discussed in prior sections. These strategies involve a physics-augmented input feature space, a physics-informed loss function, and network architecture design based on physical principles. Of note, the authors sought to emulate the mapping between the remaining useful life of the bearing with respect to features extracted from vibrational signals through a custom physics-informed layer in the network. The layer is designed to ensure that the process of neural network computations adheres to that dictated by prior physical knowledge.

\begin{figure}[h!]
\begin{center}
\includegraphics[width=\textwidth]{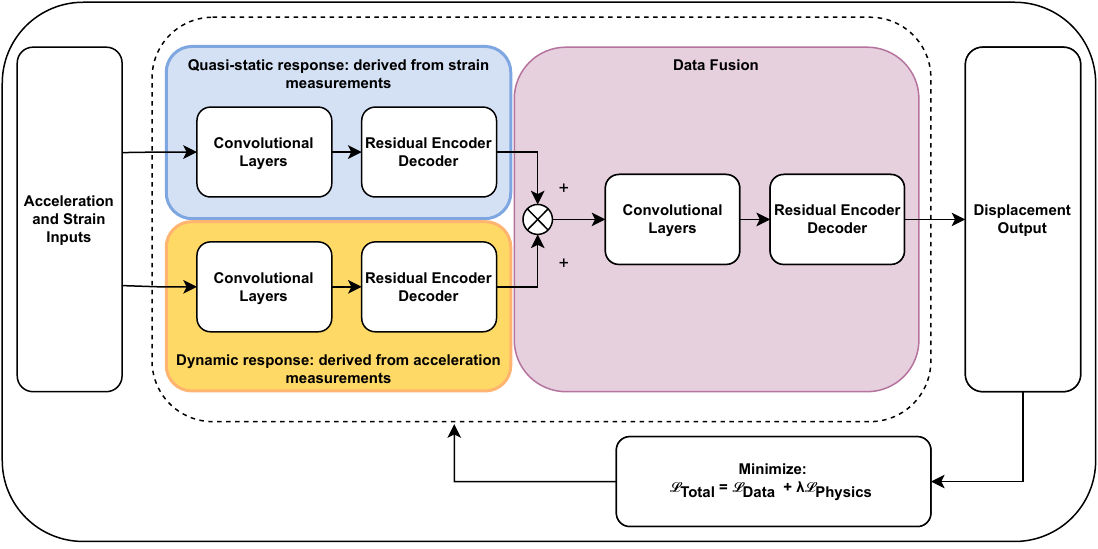}
\end{center}
\caption{Design of a multi-branch CNN, for individual modeling of displacement form strain and acceleration measurements respectively; adapted from \cite{ni2022traffic}} \label{NiCNN}
\end{figure}

An alternate implementation of physically inspired architecture design is demonstrated in the work of \cite{ni2022traffic}, who implemented a multi-branch structure of the CNN for the monitoring of deflection in bridge structures. Through the architecture illustrated in \ref{NiCNN}, the authors fuse analysis approaches for displacement reconstruction with respect to strain-based and acceleration-based methods. Due to the shortcomings of each method: in that acceleration-based methods are less capable of reconstructing quasi-static displacement, and pure strain-based methods are inaccurate with respect to the reconstruction of dynamic components in displacement,  the authors proposed a two-branch CNN to construct individual components of the expected displacements. In this fashion, relations between each component with respect to displacement may be learned independently of the other. Similar to the feed-forward network proposed by \cite{haghighat2021physics}, the individualistic modeling of physical parameters within the network is more efficient with regard to optimization. Feature maps, in this scenario, are also independent of each other, allowing for each branch of the network to exclusively focus on defining features characteristic of quasi-static, or dynamic response, with minimal "false" or spurious interference from feature maps depicting another type of behavior. A further residual encoder-decoder block was employed following convolution layers for enhanced information transmission. Components are aggregated, and further processes through convolution layers and residual encoder-decoder layers for enhanced accuracy and robustness to noise. The process is also supervised by a physics-informed loss function based on the minimization of the residual between predicted displacements through time states, formulated as acceleration term from calculus, and observed acceleration. A similar idea is illustrated by \cite{rahimi2021line}, who introduced a decision-making algorithm capable of alerting operators to abnormal conditions such as chatter in the milling process. The framework combines results from a physics-based vibration analysis, as well as spectral features from a CNN to determine probabilistically, the presence of chatter during operations. Through this design, the authors circumvented the issues with existing physics-based monitoring methods, in which false alarms are often produced due to the transient vibrations from the excitation of the machine under dynamic operating conditions. Based on the energy-based chatter detection model, the hybrid framework trains a CNN in parallel during the machining process to ascertain the specific state of operation, with assistance from the physics-based model. In conjunction with the physics-based model, the probability of chatter is updated with the operating state for an accurate and robust prediction.

\subsubsection{Recurrent Neural Networks}

Another popular deep-learning architecture popular within the community is the RNN. RNNs have been prevalent since their inception due to their capabilities in processing sequential data: taking into account the context of the previous inputs in a sequence. Information from the previous time state is parsed as the inputs to a new time state, along with the conventional input data, allowing the network to incorporate information from previous inputs into its current processing. As a direct result, RNNs are inherently designed to encode temporal invariance and have been proven to be invaluable in tasks that involve understanding temporal dynamics and relationships.

\begin{figure}[h!]
\begin{center}
\includegraphics[width=\textwidth]{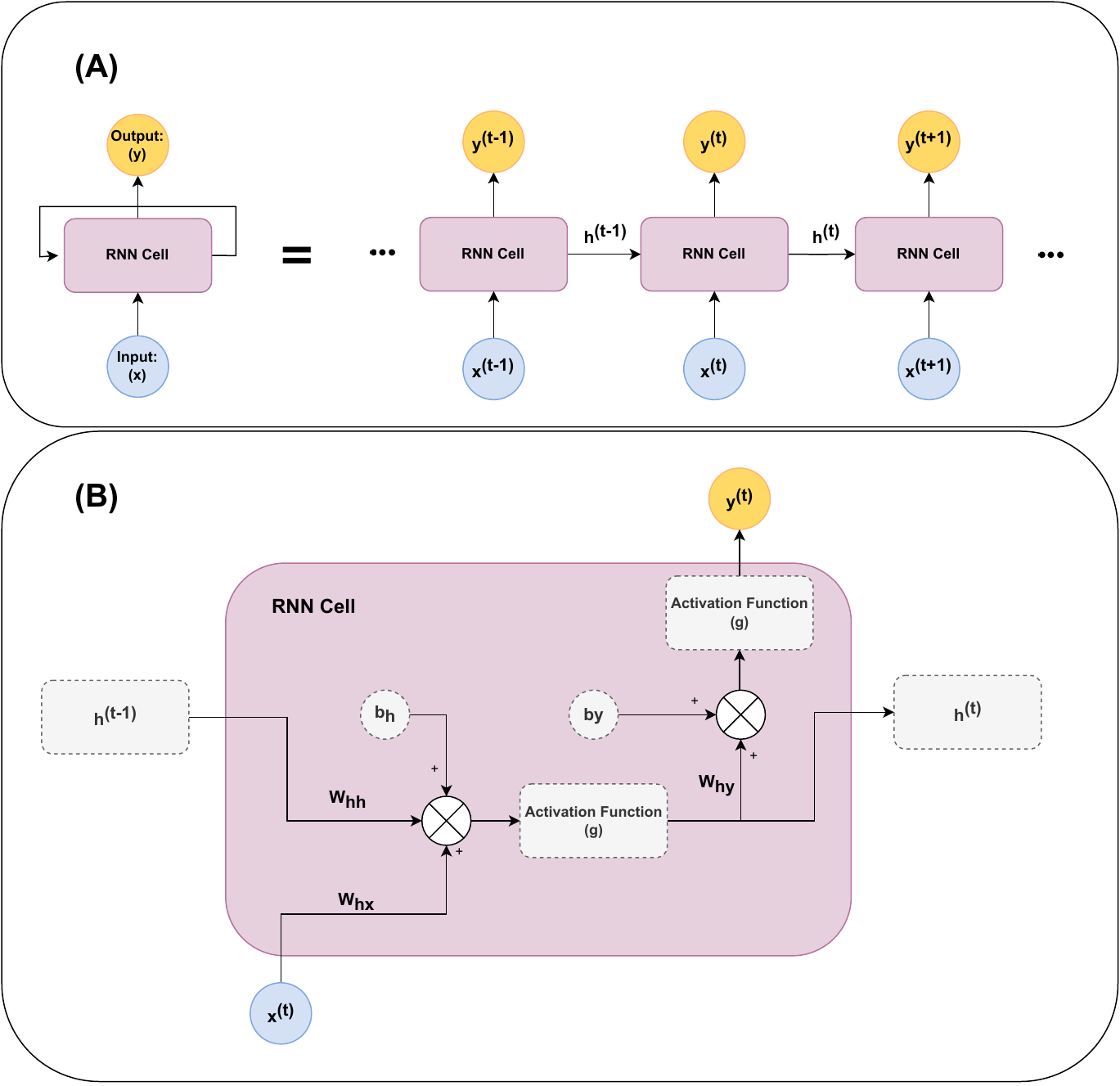}
\end{center}
\caption{An illustration of (A) the general Recurrent Neural Network architecture, and (B) the inner computational processes within each RNN cell.} \label{RNN}
\end{figure}

Conventional RNNs maps some input \(x^{\left(t\right)}\) at time \(t\) to an output \(y^{\left(t\right)}\) through possessing information form both the input space \(x^{(t)}\), and prior time state \(h^{(t-1)}\), also known as the hidden state. An illustration of the RNN architecture may be seen in figure \ref{RNN} (A).

The mathematical representation of an RNN may be written as follows, for the given input and prior hidden state, the hidden state of a cell may be represented as:
\begin{equation}
    \ z^{\left(t\right)} = W_{hh}h^{\left(t-1\right)} + W_{hx}x^{\left(t\right)} + b_{h} \label{eq:06}
\end{equation}
 Where \(W_{hh}\) and \(W_{hx}\) represents the weight matrix associated with the prior temporal state, and current input state respectively, and \(b_{h}\) represents the associated bias for the current hidden state. A non-linear activation function \(g\left(.\right)\) is applied element-wise to produce the hidden state of the cell:
\begin{equation}
    \ h^{\left(t\right)} = g\left(z^{\left(t\right)}\right) \label{eq:07}
\end{equation}
Following this, the output at time \(t\), \(y^{\left(t\right)}\), may be represented as:
\begin{equation}
y^{\left(t\right)} = g(W_{hy} h^{\left(t\right)} + b_y)
\end{equation}
Whereby \(W_{hy}\) and \(b_{y}\) represents the associated weights and biases respectively. The activation function \(g\left(.\right)\), typically the soft-max or sigmoid function, is applied to a linear transformation of the hidden cell state to produce the final cell state output. The above computational process is visually represented in figure \ref{RNN} (B). From their feedback connections, RNNs are capable of maintaining hidden cell states that capture the information from prior time states, granting the ability to process sequential data and capture temporal dependencies. Additionally, unlike other structures like the CNN, RNNs and their variants have the flexibility in processing and outputting sequences of varying lengths, allowing them to be applied to processes involving data with dynamic lengths, a common property in real-world monitoring applications. 

Long Short-Term Memory (LSTM) and Gated Recurrent Unit (GRU) are two popular variants of RNNs developed to address the problem of vanishing gradients, a prevalent issue in the training process of traditional RNNs. LSTMs and GRUs both use gating mechanisms to selectively store or discard information within the internal memory. These mechanisms enable LSTMs and GRUs to capture long-term dependencies in the data, while simultaneously alleviating the issue with vanishing gradients. LSTMs, introduced in the work of \cite{hochreiter1997long}, has since become one of the most widely used variants of RNNs. LSTMs maintain an additional internal cell state representing long-term memory and employ three gating mechanisms to regulate the flow of information. The input gate selectively updates the memory cell with new information from the input of the cell network, while preventing irrelevant information from being added to the existing memory state. The forget gate allows for the selective removal of irrelevant information from the memory cell. Finally, the output gate selectively passes relevant information from the memory to the next hidden state and output, effectively controlling the flow of information through the network. A more recent variant of the RNN, the GRU, was introduced in the work of \cite{chung2014empirical}, and is a simpler variant of LSTMs that use two gating mechanisms: the update gate and the reset gate. The update gate determines how much of the new input should be stored in the memory cell, while the reset gate determines how much of the previous memory should be discarded. In addition to these variants, the introduction of bi-directionality in the RNN architecture has also been well-studied, whereby at the cost of increased computational resources, the hidden states of two RNNs processing information in forward and backward time steps are combined, allowing the network to capture information from both past and future contexts.

In spite of their popularity, RNNs and variants of the RNN model have major limitations in terms of their computational efficiency.  This limitation arises due to the sequential nature of the RNN computation \citep{kolengradient}. For sequential data processing tasks, the inefficiency of RNNs for parallel computation may be a major limitation, especially when dealing with large-scale datasets. Due to the nature of their computations involving sequential dependencies and hidden states, RNNs require a significant amount of time and computational resources to process each data point, especially for long sequences or deep architecture. This sequential dependency also makes it challenging to parallelize the computations across time steps, as the hidden states need to be computed in a sequential manner, severely limiting the ability of RNNs to take advantage of parallel processing architectures, such as GPUs or TPUs, and leading to further delays and inefficiencies in the monitoring process. In Table \ref{table:6}, an overview of the literature reviewed is provided.

\begin{singlespacing}
\begin{scriptsize}
\begin{longtable}[h!]{p{3.5cm} p{1.5cm} p{3.5cm} p{2cm}} \hline \hline 

    \textbf{Article Title} & \textbf{Citation} & \textbf{Description} & \textbf{Application} \\ 
    \hline \hline \\
    \endhead

    \hline
    \endfoot

    \caption{A summary of literature compiled for the design of physics-informed architecture, with innovations to the recurrent neural network architecture and its variants.}
    \label{table:6}
    \endlastfoot

     \RaggedRight{Structural dynamics simulation using a novel physics-guided machine learning method} & \centering{\cite{yu2020structural}} & \RaggedRight{Embedded residual block within RNN cell as a representation of prediction consistency with physics, iteratively optimized through a deep residual-based RNN} & \RaggedRight{Structural health monitoring through dynamic simulations} \\ \hline \\
     
    \RaggedRight{Physics-Informed Deep Neural Network for Bearing Prognosis with Multisensory Signals} & \centering{\cite{chen2022physics}} & \RaggedRight{Physical knowledge regarding monotonic degradation behavior integrated within LSTM cell, regularized by physics-informed loss function based on observed degradation} & \RaggedRight{Prognosis and remaining useful life estimation in bearings} \\ \hline \\

    \RaggedRight{Fleet prognosis with physics-informed recurrent neural networks} & \centering{\cite{nascimento2019fleet}} & \RaggedRight{Paris' law governing crack growth embedded within RNN cell as a physics-based module to capture cumulative damage employing the RNN architecture} &  \RaggedRight{Prognosis with respect to fatigue crack propagation in aircraft} \\ \hline \\

    \RaggedRight{Cumulative damage modeling with recurrent neural networks} & \centering{\cite{nascimento2020cumulative}} & \RaggedRight{Paris' law governing crack growth embedded within RNN cell as a physics-based module to capture cumulative damage employing the RNN architecture} &  \RaggedRight{Prognosis with respect to fatigue crack propagation in aircraft} \\ \hline \\

    \RaggedRight{Wind Turbine Main Bearing Fatigue Life Estimation with Physics informed Neural Networks} & \centering{\cite{yucesan2019wind}} & \RaggedRight{Data driven method to evaluate grease degradation. Utilizing parameters of characterized grease degradation, as well as physical modeling to characterize bearing fatigue, embedded within RNN cell} &  \RaggedRight{Prognosis with respect to bearing under fatigue and grease degradation} \\ \hline \\

     \RaggedRight{A hybrid model for main bearing fatigue prognosis based on physics and machine learning} & \centering{\cite{yucesan2021hybrid}} & \RaggedRight{Modified RNN cell for evaluation of grease degradation and bearing fatigue simultaneously} &  \RaggedRight{Prognosis with respect to bearing under fatigue and grease degradation} \\ \hline \\
    
    \RaggedRight{A hybrid physics-informed neural network for main bearing fatigue prognosis under grease quality variation} & \centering{\cite{yucesan2022hybrid}} & \RaggedRight{Physics of degradation embedded within RNN cell, with focus on a probabilistic methodology for the identification of grease quality and variation} & \RaggedRight{Prognosis with respect to bearing under fatigue and grease degradation} \\ \hline \\

    \RaggedRight{Hybrid physics-informed neural networks for main bearing fatigue prognosis with visual grease inspection} & \centering{\cite{yucesan2020hybrid}} & \RaggedRight{Modified RNN cell for evaluation of grease degradation and bearing fatigue simultaneously. A novel ordinal classifier that aids in calibrating model for grease degradation} &  \RaggedRight{Prognosis with respect to bearing under fatigue and grease degradation}\\ \hline \\
    
    \RaggedRight{A Hybrid Model for Wind Turbine Main Bearing Fatigue with Uncertainty in Grease Observations} & \centering{\cite{yucesan2020hybriduncertain}} & \RaggedRight{Modified RNN cell for evaluation of grease degradation and bearing fatigue simultaneously. A novel ordinal classifier that aids ins calibrating model for grease degradation} &  \RaggedRight{Prognosis with respect to bearing under fatigue and grease degradation} \\ \hline \\
    
    \RaggedRight{A Probabilistic Hybrid Model for Main Bearing Fatigue Considering Uncertainty in Grease Quality} & \centering{\cite{yucesan2021probabilistic}} & \RaggedRight{Graph implementation of physics-informed and data-driven components within RNN cell, for estimation of fatigue damage accumulation with consideration to bearing fatigue and grease degradation.} & \RaggedRight{Prognosis with respect to bearing under fatigue and grease degradation} \\ \hline \\

     \RaggedRight{Estimating model inadequacy in ordinary differential equations with physics-informed neural networks} & \centering{\cite{viana2021estimating}} & \RaggedRight{Utilised physics-based RNN as a method of numerical integration for the solution of ordinate differential equations. A data-driven term is introduced to correct for discrepancies in physics through embedding within the physics-based RNN} & \RaggedRight{Prognosis in various models subject to complex degradation mechanism} \\ \hline \\
    
    \RaggedRight{Physics-Informed Neural Networks for Corrosion-Fatigue Prognosis} & \centering{\cite{dourado2019physics}} & \RaggedRight{Integration of Walker's equation governing fatigue crack growth within RNN cell architecture to model corrosion fatigue stress. Parameters of the equation are solved via data-driven or physics-based components within the RNN cell.} &  \RaggedRight{Prognosis with respect to corrosion damage and fatigue in aircrafts}\\ \hline \\
      
    \RaggedRight{Physics-informed neural networks for bias compensation in corrosion-fatigue} & \centering{\cite{dourado2020physics}} & \RaggedRight{RNN with modified physics-based layers incorporates Walker's mean stress model for fatigue crack propagation. Employed data-driven layers to compensate for additional corrosion degradation} &  \RaggedRight{Prognosis with respect to corrosion damage and fatigue in aircrafts} \\ \hline \\

    \RaggedRight{Hybrid physics-informed neural networks for lithium-ion battery modeling and prognosis} &  \centering{\cite{nascimento2021hybrid}} & \RaggedRight{Integration of the Nernst and Butler-Volmer equations within RNN cell to represent battery discharge at each time state. Data-driven neural network module within RNN compensates between known physics and observed degradation behavior} & \RaggedRight{Prognosis of degradation of charge and aging within batteries} \\ \hline \\

    \RaggedRight{Li-ion Battery Aging with Hybrid Physics-Informed Neural Networks and Fleet-wide Data} & \centering{\cite{nascimento2021li}} & \RaggedRight{Fleet-wide prognosis with modified RNN cell structure, based on physical degradation behavior governed by the Nernst and Butler-Volner equations} & \RaggedRight{Prognosis of degradation of charge and aging within batteries} \\ \hline \\
    
    \RaggedRight{Novel informed deep learning-based prognostics framework for on-board health monitoring of lithium-ion batteries} & \centering{\cite{giorgiani2023quantifying}} & \RaggedRight{Cumulative damage model via modifications to the RNN cell, comprised of physics-informed modules and data-driven neural network module, for prediction of charge state. Neural network embedded further regularized via Monte-Carlo dropout} & \RaggedRight{State of Heath Monitoring and prognosis of degradation of charge and aging within batteries} \\ \hline \\

    \RaggedRight{Novel informed deep learning-based prognostics framework for on-board health monitoring of lithium-ion batteries} & \centering{\cite{kim2022novel}} & \RaggedRight{Cumulative damage model via modifications to the RNN cell, comprised of physics-informed modules and data-driven neural network module, for prediction of charge state. Neural network embedded further regularized via Monte-Carlo dropout} & \RaggedRight{State of Heath Monitoring and prognosis of degradation of charge and aging within batteries} \\ \hline \\

    \hline \hline \\[0.01cm]
    
\end{longtable}
\end{scriptsize}
\end{singlespacing}

\begin{figure}[h!]
\begin{center}
\includegraphics[width=\textwidth]{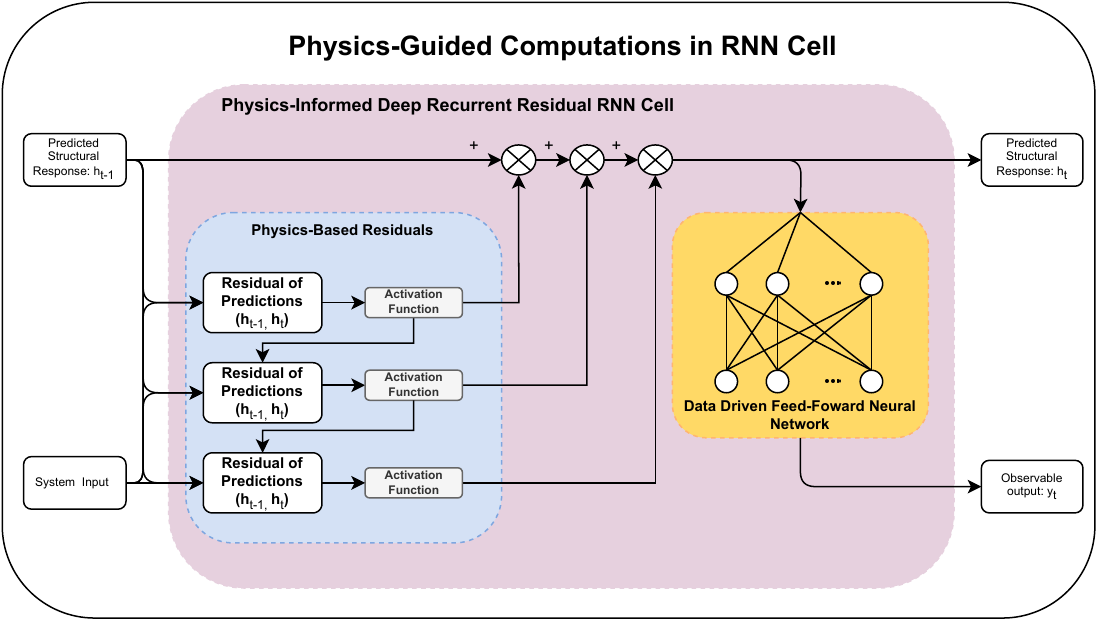}
\end{center}
\caption{Incorporation of physics within an RNN cell, via physics-informed Deep Residual Recurrent Neural Network: adapted from \cite{yu2020structural}} \label{DRRNN}
\end{figure}

An approach prevalent in literature is the incorporation of physics-based constraints directly into the RNN architecture, whereby the neural network architecture is designed to incorporate physical models as an integral part of the model's architecture. This can be achieved by including physical equations or constraints as additional layers in the neural network, which are trained alongside the traditional neural network layers. For example, \cite{yu2020structural} augmented the structure of RNNs by embedding physics-informed residual blocks within certain RNN cells for structural dynamic simulation. Residual values represent the deviation of predictions with the known physics. Within the context of their work, the residual block seeks to model exactly the inconsistencies in the dynamic system between each time state and is iteratively minimized through the proposed RNN, as illustrated in \ref{DRRNN}. \cite{chen2022physics} proposed an architecture involving the LSTM for the detection of faults and prognosis for bearings. The proposed method has been referenced as a degradation-consistent RNN. The network is physically informed through the integration of the monotonic degradation behavior of mechanical components.  The authors enforce the irreversible nature of the degradation behavior of bearings through the introduction of an intermediary variable within the network. The variable is a representation of degradation in time and is embedded within the cell of an LSTM network. The authors also implemented a physics-informed loss function whereby the performance of the training phase is evaluated against labeled data. A physics-informed term evaluates the observed degradation at any state, with the intermediary variable representing degradation, further re-enforcing the underlying physics represented by the LSTM. 

A recent popular methodology in literature comes in the form of cumulative damage modeling based on RNN. Initially employed by \cite{nascimento2019fleet}, an RNN to model was employed to capture the temporal dynamics of a machine fleet. The authors incorporated domain knowledge regarding the physics of the machines in question into the model through the inclusion of physics-based model elements directly within the RNN architecture in a format that they have termed the \emph{Euler Integration Cell}, as seen in figure \ref{fig:4}. Employing Euler's forward method, the authors formulated the discretized system state as a function of the previous system state and the input vector. In this particular instance, based on Paris' law governing crack growth, a novel RNN architecture was developed whereby a physics-informed layer was incorporated within the cell of a conventional RNN architecture to model mechanical factors affecting crack propagation. Working in tandem with the physics-based model, the traditional data-driven model estimates the stress intensity factor range. The combination of these two models with the RNN cell yields an accurate estimation of temporal dynamics and cumulative damage in the specimen. \\

\begin{figure}[h!]
\begin{center}
\includegraphics[width=\textwidth]{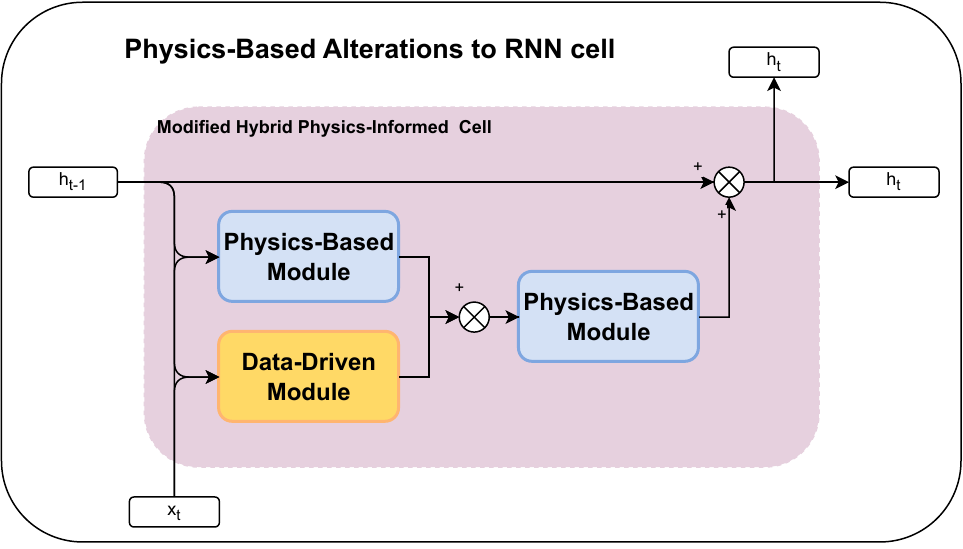}
\end{center}
\caption{Incorporation of physics within a recurrent neural network cell via Euler integration: adapted from \cite{viana2021estimating}} \label{fig:4}
\end{figure}

Following this publication, the authors also employed the same model to estimate fatigue crack length growth in aircraft with limited observations \cite{nascimento2020cumulative}. Subsequent works by other authors are based upon the modification and tailoring of the framework in alternative applications. For instance, studies by \cite{yucesan2019wind, yucesan2021hybrid, yucesan2022hybrid} applied a modified version of this framework to model bearing fatigue in wind turbines, whereby the data-driven model was employed in combination with known physics to estimate the unknown effects of lubrication on failure. Through a combination of data-driven elements within the cell, as well as physics-based layers such as the Palmgren-Miner’s rule, the authors sought to characterize the relationship between bearing fatigue and grease degradation through the combined network. To this effect, the structure of the network was designed to take into account parameters from both degradation behaviors, and accurately characterize each form of degradation with respect to the other. The authors further innovated on this model by extending its applications to used cases whereby elements of uncertainty are introduced to the grease degradation process \citep{yucesan2020hybrid, yucesan2020hybriduncertain, yucesan2021probabilistic}.\\

\cite{viana2021estimating} presented a method for estimation of missing physics utilizing the model, whereby a data-driven layer is employed to approximate the uncertain behavior of the physical model. Interestingly, \cite{viana2021estimating} chooses to employ the RNN architecture as a purely physics-based solution to ordinary differential equations, with the addition of a data-driven node to quantify the discrepancy between known physics and observed results. The authors have verified the approach with various case studies such as the modeling of fatigue governed by established physics-based models such as Paris' law for fatigue crack propagation, Walker's equation for fatigue crack propagation, and  Palmer-Miner’s rule for fatigue life estimation.\\

Another avenue for the application of this hybrid RNN architecture was explored by \cite{dourado2019physics} who employed a similar framework for the estimation of corrosion effects system cumulative damage. In their work, the structure of the RNN was designed to represent Paris's equation, with stress intensity factors being determined physically, and the rest of the parameters being determined by data-driven feed-forward modules within the cell. The authors later expanded on their work with the introduction of a data-driven compensator to correct for Walker's model for crack propagation, whereby data-driven layers are employed to model the bias in damage accumulation as a result of corrosion \citep{dourado2020physics}. The cumulative damage model has also seen much use in modeling degradation behaviors in lithium-ion batteries: for instance, based on their previous work, \cite{nascimento2021hybrid} modified the existing framework to be consistent with the Nernst and Butler–Volmer equations, with a multi-layer perceptron module within the cell to characterize the model-form uncertainty. The approach focuses on building a reduced-order model based on Nernst and Butler-Volmer equations.  Following a similar idea as \cite{viana2021estimating}, the authors employed multiple data-driven modules within their modified RNN cell structure to compensate for deviations between known physics, and observed degradation in the asset. The authors further expanded upon their work to extend the range of applications. In the work: \citep{nascimento2021li} to a fleet-wide dataset, allowing for the identification of assets deviating from fleet norms established. While in \cite{giorgiani2023quantifying}, the authors further extended the model for use with incomplete historical usage of assets through a Bayesian update strategy of revising the probability of a hypothesis or belief based on new evidence or information. \cite{kim2022novel} applied the cumulative damage framework to the estimation of lithium-ion battery state in a model that they have termed the knowledge-infused RNN. In their model, the recurrent cell is further modified via the addition of physics-informed modules based on a double-exponential model of battery capacity. Furthermore, the authors also introduce a Monte-Carlo dropout within the data-driven feed-forward network embedded within the RNN cell to secure robust and reliable probabilistic estimations of performance. 

\subsubsection{Graph Neural Networks}
\begin{figure}[h!]
\begin{center}
\includegraphics[width=\textwidth]{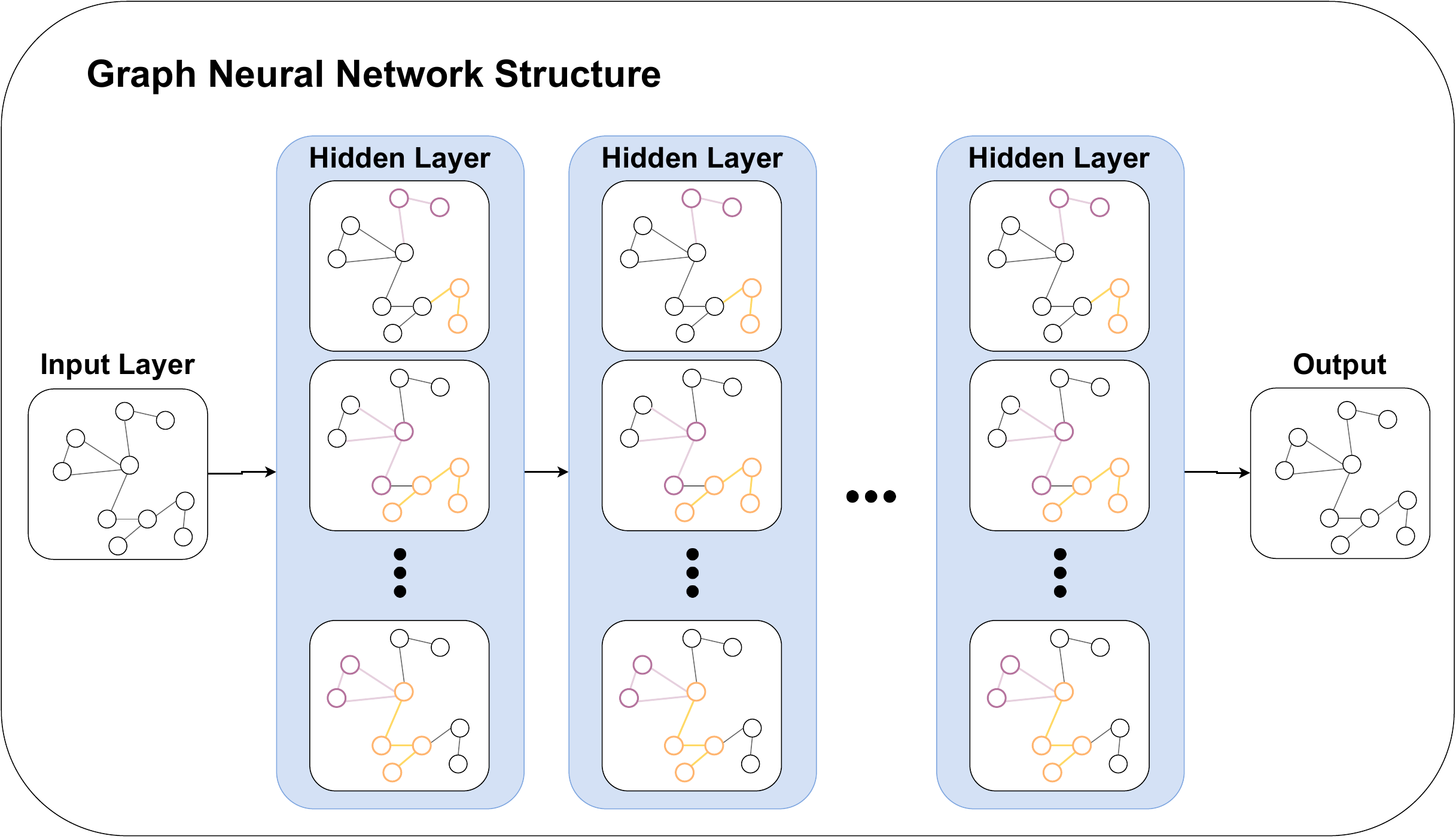}
\end{center}
\caption{A depiction of a graph neural network (GNN) architecture, showcasing its key components and information flow: GNNs operate on graph-structured data, enabling effective analysis, inference, and learning tasks within complex relational datasets.}\label{GNN}
\end{figure}

Another example of physics-informed architecture comes from the structural composition of Graph Neural Networks (GNNs). GNNs are a class of deep learning models capable of processing graph-structured data, initially conceptualized by \cite{scarselli2008graph}.  GNNs are comprised of nodes and edges, as defined in the work of \cite{scarselli2008graph}. In this representation, nodes within the network represent entities and edges represent the connection or relation between entities. An illustration of this architecture is shown in figure \ref{GNN}.

For graph \(G = \left(V, E\right)\) with nodes (also known as vertices) \(V\) and edges \(E\), each node may be represented as \(v \in V\) with feature vector \(h_{v}\). The operation of the GNNs may then be defined as the iterative process of updating the node feature vector representations by aggregating information from their neighboring nodes and then using these updated representations to make predictions or classifications. Employing a message-passing mechanism, nodes exchange information with neighboring nodes, enabling them to update the feature vector based on the information received. This operation is reminiscent of the convolution operation applied for CNNs, in the sense that both operations effectively aggregate and process neighboring entities to update the value of the entity in question. Each node aggregates information from its neighbors using a learnable function that takes into account both the node features and the edge weights. This information is then passed to each node's neighbors. This may be understood as follows:  for each node \(v\), compute the message vector \(m_{v}^{l}\) through aggregating information from its neighbors \(N\left(v\right)\) using a learnable function \(f_{e}\):

\begin{equation}
    \ m_{v}^{l} = \sum_{u \in N(v)}{f_{e}^{l}(h_{v}^{l-1}, h_{u}^{l-1}, e_{v,u}^{l})}
\end{equation} 

Where \(e_{v,u}^{l}\) is the edge feature vector between nodes \(v\) and \(u\), \(f_{e}^{(l)}\) is the learnable function that maps the features inputted of prior layer \(l-1\) to the resultant message at layer \(l\). Subsequently, each node updates its representation through the combination of information it received from its neighbors in the prior procedure, with its own original representation. This may be represented as the computation of the new feature vector \(h_v^{l}\) for each node \(v\) by combining the previous feature vector with the aggregated messages:

\begin{equation}
h_{v}^{t+1} = f_{n}^{l}(h_{v}^{l}, m_{v}^{l})
\end{equation}

The message passing and updating steps outlined above are repeated for a fixed number of layers until a final representation is obtained for each node in the graph. The final node features can then be used for downstream tasks such as node classification or link prediction. The functions \(f_{e}\) and \(f_{n}\) can be any differentiable functions and varies with respect to the application. Typically in a GNN, this is approximated with a deep learning structure such as the feed-forward neural network, or graph convolutional networks, and can be learned through back-propagation during training. Through this connected architecture, GNNs are capable of capturing complex relationships between entities in the graph, such as the local and global structure of the graph, enabling them to make accurate predictions and perform various tasks on graph-structured data. A summary of the literature compiled is presented in Table \ref{table:7}.

\begin{singlespacing}
\begin{scriptsize}
\begin{longtable}[h!]{p{0.25\textwidth}  p{0.11\textwidth} p{0.25\textwidth} p{0.25\textwidth}} \hline \hline \\
    \textbf{Article Title} & \textbf{Citation} & \textbf{Description} & \textbf{Application} \\ 
    \hline \hline \\[0.05cm]
    \endhead

    \hline
    \endfoot

    \caption{A summary of literature compiled for the applications of the physics-informed graph neural network architecture.}
    \label{table:7}
    \endlastfoot

    \RaggedRight{Physics-informed geometric deep learning for inference tasks in power systems} & \centering{\cite{deJongh2022physics}} & \RaggedRight{GNN with physics-informed loss function based on power flow}& \RaggedRight{State estimation and anomalous behaviour detection power systems} \\ \hline \\
    
    \RaggedRight{PPGN: Physics-Preserved Graph Networks for Real-Time Fault Location in Distribution Systems with Limited Observation and Labels} & \centering{\cite{li2021ppgn}} & \RaggedRight{Physics of power grid embedded in graph structure to train a GNN} & \RaggedRight{Fault detection and localization in power systems} \\ \hline \\
    
    \RaggedRight{Data-Driven Transmission Line Fault Location with Single-Ended Measurements and Knowledge-Aware Graph Neural Network} & \centering{\cite{xing2022data}} & \RaggedRight{Inherent relationship between observable parameters and fault location embedded in graph structure} & \RaggedRight{Fault detection and localization in power systems}\\ \hline \\
    
    \hline \hline \\[0.01cm]
    
\end{longtable}
\end{scriptsize}
\end{singlespacing}

 The inherent structure of GNNs, which allows them to operate on graph-structured data, makes them suitable for applications with various real-world systems, in which the behavior of the system is determined by complex interactions between various components and can be naturally represented as a graph. In particular, GNNs have emerged as a powerful approach in modeling power systems, for applications such as power system state estimation, load forecasting, fault detection and diagnosis, and optimal power flow estimation \citep{liao2021review, yu2022pidgeun, gao2020physics, zhu2022cascading}. The popularity of graph neural networks in modeling power systems may be attributed to the structure of power systems being inherently graph-like as well, consisting of interconnected nodes (such as power generators, transformers, and loads) and edges (such as transmission lines and cables) that represent the flow of power and information. For example, through a physics-informed GNN, \citep{deJongh2022physics} monitored and performed state estimations in their study. Power systems exhibit an underlying, irregular structure in the form of grid topology, which can be represented mathematically as a graph. Due to this structure, geometric deep learning methods such as GNNs are suitable due to their inherent structure. The group proposed a generic framework that uses geometric deep learning techniques and a physics-informed loss function to solve power flow calculation and state estimation tasks in power systems. The framework is shown to perform well on simulated medium voltage grid topologies with various sensor penetrations. \citep{li2021ppgn} further proposed a physics-preserved graph network for the estimation of the location of faults in a power grid system. The two-stage framework provided an accurate estimation of node fault location with limited data. Through a novel adjustable adjacency matrix by which sparse fault currents are aggregated, the first stage of the framework approximates the topology of the structure. Whereas the second stage of the framework learns the correlation between observed, and non-observable data samples. Finally, \cite{xing2022data} adapted the physics-informed GNN framework to improve fault location in transmission lines. The authors incorporated prior physical knowledge through their establishment of a graph structure of known fault types, whereby the inherent relation between fault types and locations is incorporated within measured mode voltages and measured mode currents.

\subsubsection{Generative Deep Learning Networks}
\label{Generative Networks}
Generative adversarial networks (GANs) are a class of ML models designed to automatically discover and learn regularities from training data, such that the model may be able to generate realistic samples of data that plausibly could have belonged to the dataset provided. GANs consist of two neural networks that are trained collectively in a competitive setting. A generator network primarily learns to generate samples that resemble the training data provided, and a discriminator network learns to distinguish between the generated samples and the real training data \citep{wang2017generative}. The generator network takes random noise or a latent vector as input and generates synthetic data samples. As the process of training progresses, the generator network learns to generate increasingly realistic samples that resemble the training data distribution. The discriminator is represented as a binary classifier that seeks to distinguish between real and synthetic sample data, with inputs from both real data samples from the training set and synthetic samples from the generator. As the generator network learns to generate more realistic samples, the discriminator network becomes better at distinguishing between the generated and real data and thus provides more informative feedback to the generator network. The training objective of GANs can be framed as a min-max game between the generator and the discriminator. The generator aims to minimize the discriminator's ability to distinguish between real and fake samples, while the discriminator aims to maximize its discriminative accuracy. This objective is typically expressed as the minimization of the Jensen-Shannon divergence or the Wasserstein distance between the real and generated data distributions. This iterative process continues until the generator network is able to produce samples that are indistinguishable from the real training data \citep{goodfellow2020generative}. A summary of compiled literature is provided in Table \ref{table:8}.

\begin{singlespacing}
\begin{scriptsize}
\begin{longtable}[h!]{p{0.25\textwidth}  p{0.11\textwidth} p{0.25\textwidth} p{0.25\textwidth}} \hline \hline \\
    
    \textbf{Article Title} & \textbf{Citation} & \textbf{Description} & \textbf{Application} \\ 
    \hline \hline \\[0.05cm]
    \endhead

    \hline
    \endfoot

    \caption{A summary of literature compiled for the applications of the physics-informed generative adversarial network architectures.}
    \label{table:8}
    \endlastfoot

    \RaggedRight{PhyMDAN: Physics-informed knowledge transfer between buildings for seismic damage diagnosis through adversarial learning} & \centering{\cite{xu2021phymdan}} & \RaggedRight{Multiple source domain adaptation framework, with physics-guided loss function based on similarities in domains} & \RaggedRight{Structural health monitoring in buildings}\\ \hline \\
   
    \RaggedRight{A new cyclical generative adversarial network-based data augmentation method for multi-axial fatigue life prediction} & \centering{\cite{sun2022new}} & \RaggedRight{Dynamic Time Warping equation to eliminate generated samples inconsistent with physical knowledge} & \RaggedRight{Fatigue life estimation for specimen under multi-axial loading} \\ \hline \\ 
    
    \RaggedRight{Deep convolutional generative adversarial network with semi-supervised learning enabled physics elucidation for extended gear fault diagnosis under data limitations} &  \centering{\cite{zhou2023deep}} & \RaggedRight{Deep Convolutional Generative Adversarial Network to establish implicit physical correlation between known and new faults} & \RaggedRight{Machinery health monitoring in gear transmissions} \\ \hline \\ 
    
    \RaggedRight{Adversarial uncertainty quantification in physics-informed neural networks} & 
    \centering{\cite{yang2019adversarial}} & \RaggedRight{Generative adversarial network for construction of surrogate models to physical systems, regularized via physics-informed loss function} & \RaggedRight{Uncertainty quantification and propagation in non-linear systems} \\ \hline \\
    
    \RaggedRight{Physics-informed deep learning: A promising technique for system reliability assessment} &  \centering{\cite{zhou2022physics}} & \RaggedRight{Generator network constrained by domain knowledge via physics-informed loss function, trained in an adversarial setting with discriminator to produce probabilistic estimates of system state and reliability for Markovian systems} & \RaggedRight{Reliability assessment and degradation monitoring} \\ \hline \\

    \hline \hline \\[0.01cm]
    
\end{longtable}
\end{scriptsize}
\end{singlespacing}

Leveraging the underlying physics of the system, physics-informed GANs have been employed to generate synthetic data that may be used to supplement the available observational or measured data, thus enabling more accurate modeling and prediction. Core to its functionality, physics-informed GANs constrain the generated samples through the application of physical laws. Some instances of this exist in works by \cite{xu2021phymdan}, who introduced a framework that they have termed Physics-Informed Multi-source Domain Adversarial Networks for the unsupervised identification of structural damage in buildings. The proposed method employs a multiple-source domain adaptation framework that seeks to extract domain-invariant features from a variety of source domains. The authors proposed a novel loss function whereby additional consideration is given to similarities between source and target domains, and the knowledge transfer from similar source domains are prioritized. Similarly, \cite{sun2022new} presented a cyclical GAN model that embeds the physics of the hysteretic behavior within the network for the augmentation of available data. More specifically, the authors aimed to capture the relation between cyclic hysteresis loops of the half-life cycle of a specimen under multi-axial loading and corresponding fatigue life.  The authors enabled the generation of synthetic data that obeys the distribution characteristics of real fatigue behavior, constrained by physical laws through various Fourier transforms and semi-empirical equations. Through the Dynamic Time Warping algorithm and various semi-empirical equations representing the relation between fatigue life and loading, strain loading, and stress response, samples with deviations from physical principles are eliminated. With the augmentation, multi-axial fatigue life data of a test specimen was employed to train several well-known ML models, including feed-forward networks, Random Forest, SVMs, and extreme gradient boosting algorithms, demonstrating a significant improvement in accuracy. Similarly, for applications in gear fault diagnosis, \cite{zhou2023deep} proposed a convolutional GAN model to extend available training data in the gear fault diagnosis process, due to the high-cost limitations of labeled fault data for specific gear fault failure modes. Through this framework, the authors leveraged fault features from large quantities of unlabelled training data to be representative of new fault data, with respect to labeled training data, effectively extending the prediction space of the deep convolutional GAN. Through this process, the physical correlation between known and unseen faults may be derived.

Various authors have also employed the framework for uncertainty quantification. This is typically performed by training the networks on data with known uncertainties, effectively allowing GANs to effectively generate synthetic data samples with associated uncertainties. The generated sample may be used to estimate and quantify the uncertainty in predictions made by machine learning models. GANs can also be used to generate diverse data samples that span the entire range of possible uncertainties, helping to improve the robustness and reliability of uncertainty quantification methods. Applications of the GAN architecture may be seen in works such as \citep{yang2019adversarial}, which employed the framework for the quantification and propagation of uncertainty pertaining to the non-linear PDEs in physics-informed neural networks. Due to limitations in data acquisition, the authors sought to produce a method of uncertainty propagation based on \emph{a-priori} knowledge by means of governing differential equations. Through latent variable models, the probabilistic representations of the system states were procured. In a latent variable model, the observed variables are typically considered to be influenced by one or more latent variables. These latent variables are not directly observed or measured but are assumed to underlie the relationships among the observed variables. The objective of the latent variable model is to estimate the values of the latent variables and understand their impact on the observed variables. An adversarial inference procedure was proposed for the training of models with respect to available data. The incorporation of physical constraints in the form of the physics-informed loss function during optimization phases of the deep adversarial generative network allows for training utilizing smaller datasets. Approximation of the solution was performed by the minimization of error with respect to minimizing the reverse Kullback-Leibler divergence. In this fashion, predictions are constrained to be consistent with known physics. Employing the physics-informed constraints as regularization mechanisms, the authors trained a deep generative model for the generation of surrogates for physical systems, effectively circumventing the issue with data acquisition through the characterization of uncertainty within the physical system outputs. This methodology has been validated with a series of experiments demonstrating uncertainty propagation in systems governed by non-linear PDEs. As another example, \cite{zhou2022physics} incorporated physics-informed GAN in their framework for system reliability analysis. The network configuration is modeled based on system state probability and encodes the governing equations of the reliability evolution model. The authors characterized the system performance at each time state problematically via derivations from the forward Kolmogorov equations, and subsequently, the system reliability as an aggregate of state probability where the system is considered to be functional. The authors further proposed a GAN network for uncertainty quantification with respect to reliability assessment, whereby the generator seeks to produce synthetic data based on the derivative of system state probability, or state transition defined. The generator model is also constrained by any observed data from initial conditions or the continued operation of the system, whereas the discriminator seeks to produce confidence estimates of the data. The two are regularized by competing loss functions and trained in an adversarial setting. In particular, physics-based regularization is employed for the generator as an additional loss term, in accordance with domain knowledge. The authors demonstrated the effectiveness of the proposed methodology with a variety of numerical examples. In all, the proposed method yielded similar results to that of conventional Runge-Kutta and Monte Carlo simulations.

\section{Discussion} \label{Discussion}
In total, a sample size of 105 literary works was explored in the survey, with the overall objective of discussing and summarizing popular implementations of PIML learning frameworks, with applications to the monitoring of assets for anomalous behavior and or operating conditions. Of the works of literature surveyed, the methods of integration between physics-based methods and data-driven models were subdivided into four distinct categories, as discussed in section \ref{PIML}. 

\subsection{Summary of Compiled Literature and Interpretations}

\begin{figure}[h!]
\begin{center}
\includegraphics[width=\textwidth]{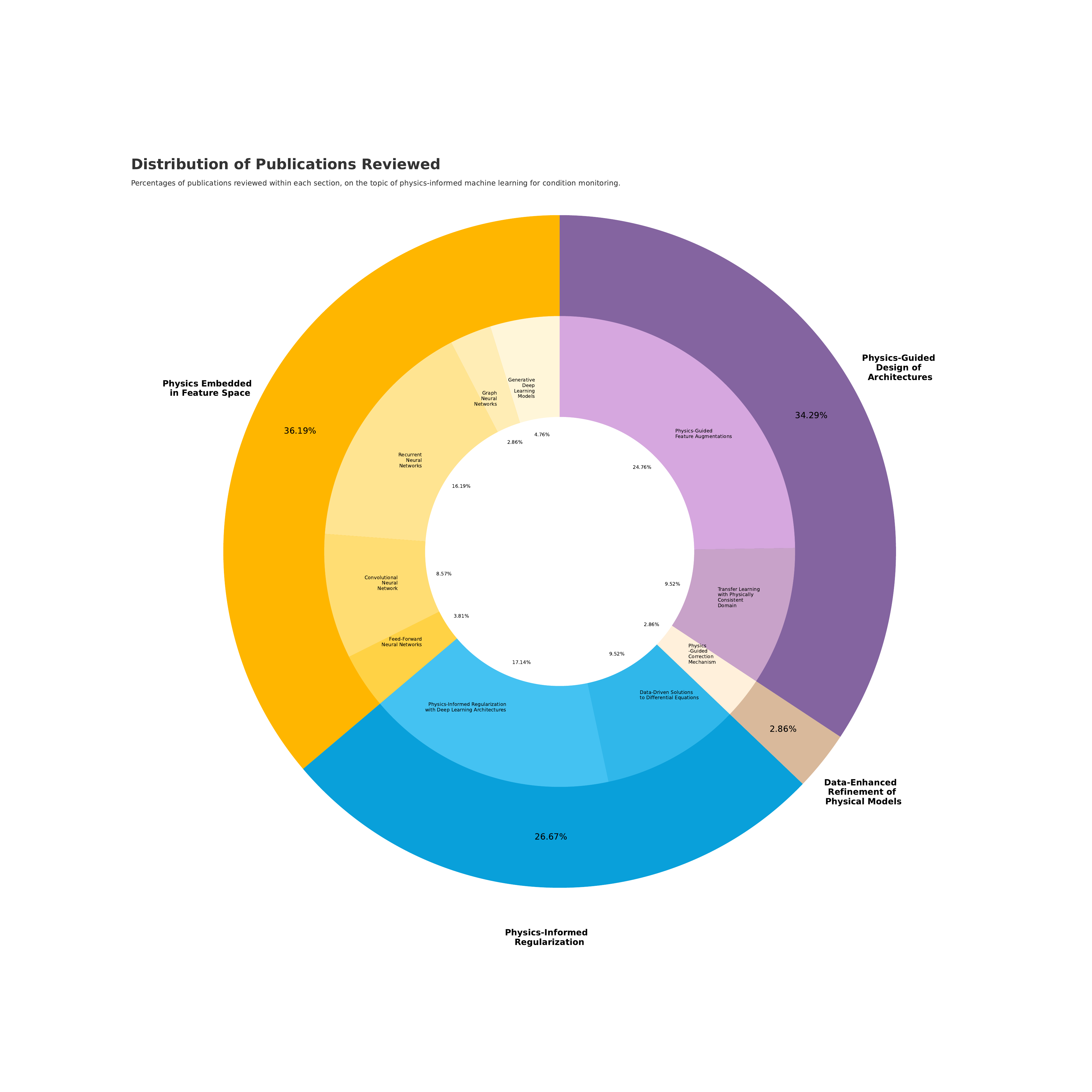}
\end{center}
\caption{Distribution of Publications Reviewed on Physics-Informed Machine Learning for Condition Monitoring} \label{distribution}
\end{figure}

An illustration of the distribution of literature reviewed may be seen in figure \ref{distribution}. The pie chart illustrates the distribution of publications reviewed, highlighting the different areas within the field of PIML for condition monitoring. 

Of the literature surveyed, a large sample of works (36) identified employed physics-based techniques to modify the input feature space of the ML model, introducing physical knowledge through observational biases. Alterations to the input space indirectly allow models to learn physically consistent relationships through restricting mappings that are not adherent to physical principles. This implementation has witnessed great popularity which may be attributed to its simplicity and ease of implementation. \\

From literature, this type of integration dealt primarily with direct physical-model-driven generation of input data or augmentation of feature space: with (21) studies reviewed aiming to generate synthetic data or used physics-based methods to create new physics-based features, and (5) studies using physics-based methods to select discriminating features. Despite the varied approaches, a commonality with the methods mentioned above is with respect to the custom tailoring of feature space for use with conventional ML and deep learning models. Incorporation of physics within the feature space of machine learning has also been performed in several works (10), utilizing the philosophy of transfer learning, whereby the model is pre-trained on a known source domain, and subsequently re-calibrated for a target domain. Almost all works in this area have designated the source domain as the known physical domain, and employed knowledge transfer to capture known physics to be re-purposed; Exceptions to this trend were with the work of \cite{guc2021fault} and \cite{guc2022sensor} who instead relied on the pre-trained features of the source domain, and incorporated physics through the fine-tuning phase. Various authors have employed this framework for the supplementation of available data and enhancement of ML learning space for improved performance and robustness. \\

With respect to the limitations of this technique, despite the ease of implementation and apparent efficacy, this type of implementation does not directly incorporate any physical constraints during the learning process, resulting in a naive black-box model, with minimal interpretability. While feature engineering may indirectly restrict the model's capabilities for physical violations, no constraints are enforced during the learning process. Furthermore, there exists a degree of dependency on the completeness and reliability of the physical model for a true-to-life or authentic generation of features. As such, the outlined approach may not be suitable for complex systems where the underlying physics is not well understood, due to the difficulties in capturing the intricacies and nuances of real-world phenomena in a set of predefined features. \\

Another formula for the incorporation of physical knowledge within ML models is the applications of data-driven modules in tandem with physics-based models, such that the data-driven model acts as a correctional mechanism to complement the decisions made based solely on physical principles. (3) of the works sampled have employed this format for their applications. While the technique has demonstrated some success as demonstrated in the above literature, the action of utilizing ML as a correction mechanism for physical models is not without limitations. \\

As with most purely data-driven models, a major limitation in this strategy is its inability to capture behaviors not present in the domain on which they are trained. In this format, ML models operate independently from the physics-based models, and as a direct result, in the case that the training data does not accurately capture the true physics of the system as is characteristic of the error of the physical system, the ML algorithm may learn to correct the errors in the physical model, but may not be able to accurately capture the underlying physical phenomena. With respect to the integration of physics with data-driven models, another major limitation results from the target learning space of the ML model. Due to the ML model learning the error of the system, rather than the system itself, it is difficult to ensure that the resulting corrections are physically meaningful. In some cases, machine learning algorithms may identify patterns or relationships in data provided that are not related to the underlying physics, leading to incorrect or spurious corrections. \\
  
More recently, through several defining contributions, physical knowledge of this system has been employed in conjunction with the powerful approximative capabilities of neural networks. Traditional methods in training a neural network involve an initial prediction made by the neural network, and its subsequent optimization in accordance with some form of distance evaluation of predictions of the neural network and prior knowledge, in the form of a loss function. Optimization in supervised learning methods has been carried out with respect to labeled data; the established methodology has remained unchanged since its inception. Recently, several authors have made innovations to this process through the introduction of physics-informed regularizations. Conventional regularization such as L1 or L2 regularization has been used extensively with ML models as a methodology for machine learning and statistical modeling to address overfitting and improve the generalization capability of the model. This effectively balances the trade-off between fitting the training data well and limiting the model complexity with additional penalty terms in the loss function. \\

With physics-informed regularization, rather than limiting model complexity, models are penalized based on their deviations from physical principles through the introduction of a physics-based loss term. (38) sampled works applied this format of regularization for their proposed methodologies. As characterized by \cite{karniadakis2021physics}, physics-based regularizations have been known to introduce knowledge of the underlying physical system through learning biases. Predictions from deep learning architectures a be iteratively guided via the loss functions over several optimization cycles to be consistent with known physics. In addition, (32) of the studies employing this methodology employ physics-based regularizations for the solution of governing ordinary or partial differential equations. Through independent variable inputs, the neural network seeks to predict the unknown variable. Leveraging automatic differentiation, the predictions of variables from the base neural network may be employed to reconstruct the differential equations, as well as initial or boundary conditions. These reconstructions are subsequently evaluated in the form of the loss function, with some studies electing to include loss with respect to labeled data as well. Initial works in this area by the likes of \cite{raissi2019physics} made use of conventional feed-forward networks, although the general framework of physics-based regularization has been quickly expanded to leverage other deep learning architectures as well as demonstrated in a sample of (6) works. Architectures such as convolutional and recurrent neural networks have been employed for their capabilities in capturing spatial and temporally invariant features respectively, and autoencoders for their unsupervised learning capabilities. Several advantages of this approach are apparent, as demonstrated by the above works. \\

The popularization of this format represents an effective methodology for the incorporation of prior knowledge of physical principles within the optimization process of neural networks, and its superiority over conventional "naive" methods has been demonstrated in various works \citep{raissi2019physics, haghighat2021physics}. Models constructed are also reliant to a lesser degree, on the available data for learning, enabling authors to reduce data requirements to train a deep learning architecture and improve the robustness of models to noisy or incomplete data. In fact, some studies (8) have employed purely physical loss terms in the optimization process. Training a model in this format may prove advantageous when limited data is available. In addition, it reduces sensitivity to noisy or inaccurate data due to the absence of dependence. In general, a data-driven loss term may also increase convergence and stability, as well as generalization to unseen data via additional guidance during training.\\ 

While the implementation of physics-informed loss functions has been proven to be effective in solving issues pertaining to condition monitoring as has been shown by many of the studies listed above, there exists some limitations to this approach. One such limitation is with respect to the method by which the physical constraints are imposed in the network. Through the physics-informed loss functions, physics-based loss terms act as a penalization for the network in the case of violations, however, they are not enforced as hard constraints. This may prove an issue in hybrid loss functions involving penalization terms with respect to labeled data in particular, as inaccuracies in the data may cause the corresponding loss term to dominate within the hybrid loss function. To a lesser extent, with respect to physics-based regularization and PINNs in general, as the physical loss is not strictly enforced, physical violations or deviations from expected physical behaviors may still be produced by the network. Another limitation of note is that the addition of a physical regularizer, depending on the problem being solved, may introduce additional degrees of complexity to the loss function overall. Current methods of optimization rely primarily on gradient descent and its variants, in which the network adjusts its parameters in steps toward the direction of minimal error with respect to loss. The increased complexity of the loss function landscape may further complicate or hinder the process of optimizations through the introduction of local minima, for example. This aspect of physics-based regularization has been noted in the work of \cite{krishnapriyan2021characterizing}, whereby the characteristic increase in model complexity has been noted with the introduction of soft regularization terms. \\

Alternatively, authors have also attempted to incorporate hard constraints through the design of neural network architecture. A total of (34) of the works sampled provided innovative solutions to incorporate physical principles as part of the computational processes of deep learning architectures itself. Overall, this format offers improved interpretability, as the computation process is designed within the framework of deep learning networks. The learned parameters and model outputs can be directly related to physical quantities, making it easier to understand and validate the predictions, allowing practitioners a deeper understanding of the process by which algorithms predict and ascertain a predicted result. Innovations have been made with respect to several popular network architectures such as the conventional feed-forward neural network (4), the CNN (9), RNN (17), GNN (3), and GANs (5). With respect to architecture design, the majority of studies examined either assigned physical meaning with respect to intermediary nodes or layers or alternatively, to the connections between nodes themselves in the form of constrained optimization \citep{chen2021probabilistic}. Of which, in addition to employing physics-informed layers, many such studies also employ a physics-informed regularisation as well, for additional guidance during the optimization process. Some studies, such as the work of \cite{chen2023physics}, alternatively employed informed activation functions for each node within the network. With respect to the feed-forward networks, several authors have proposed interpretable layers within the networks to elucidate the computational processes of data-driven models, with physical meaning being assigned to layers. This methodology represents an alternate form of physics-based feature extraction. With regards to applications with vibrational data from structural and machinery health monitoring specifically, conventional signal processing techniques such as Fourier transforms, envelope analysis, and wavelet transforms are embedded within neural network layers as a form of physics-informed feature extraction and processing. A similar technique is employed with CNNs, whereby layers within CNNs perform advanced feature selection or extraction with regard to a defined computational process that is adherent to known physics. \\

While the framework outlined above shares many similarities to simply tailoring the input feature space, as discussed in section \ref{31}, there exist several key advantages of incorporating the pre-processing stage within the network itself. For one, the framework outlined is effectively an end-to-end learning pipeline, whereby the entire network, including the pre-processing stage, is incorporated within the learning process. The advantages of this design lie in the fact that the network can adapt and optimize both the pre-processing and subsequent feature extractions simultaneously and eliminates the need for manual feature engineering. In addition, the resultant network architecture embeds physical knowledge, and is therefore more interpretable, due to the network's behavior being enforced to align with known physical principles. By explicitly modeling and accounting for factors that may be physically modeled during the feature extraction process, the network can learn to extract more reliable and invariant features, resulting in improved performance under challenging conditions. \\

The design of network layout has also been explored, as tabulated in the work of \cite{ni2022traffic}, whereby the branched network was introduced to solve for multiple pre-determinate physics-based relations independent of each other. As noted in both \cite{ni2022traffic} and \cite{haghighat2021physics}, while technically possible to solve for multiple physical variables with a wide enough network layer, in the case where the relations may be modeled independent of the other, it is often more efficient in terms of computational resources, and more accurate to model each variable individually through separate branch networks. Several studies also focus on the RNN structure, with the primary form of physical information being embedded in computational procedures within the RNN cell. Key contributors within this area include the works of \cite{nascimento2019fleet}, who initially made use of the Euler Integration cell to embed the physics of crack propagation within the RNN cell, as a representation of cumulative damage modeling. This model is later extended to various other applications in modeling the propagation of damage through time, as well as model form uncertainty \citep{viana2021estimating, yucesan2020hybriduncertain, yucesan2021probabilistic}.  Of the works covered, (14) made use of this format of integration. Other works such as the study by \cite{yu2020structural} also made alterations through the incorporation of the Deep Residual Recurrent Neural Network, as initially proposed by \cite{kani2017dr}. Utilizing the embedded physical dynamics of the system, practitioners were able to better capture dependencies and improve the model's ability to make accurate predictions over longer time horizons. (3) studies employed the graph neural network, in which the inherent structure of the network is leveraged to better model and process graph-structured data, with extensive applications in power systems. In contrast to conventional neural networks, GNNs are capable of handling non-Euclidean data via graph representations, whereby nodes in the graph structure represent entities and connections represent the relationships between them. Unique to their structure, GNNs do not assume spatial locality. This assumption is commonly used in CNNs, which are designed to operate on grid-like data, such as images. This property of GNNs allows for operation on data structures of arbitrary sizes and complex topologies. (5) samples of literature reviewed dealt mainly with the optimization of GANs, of which, (2) studies implement the network as an automatic framework for the synthetic generation of physically plausible data, while the remainder (3) employed the network to characterize and quantify the uncertainty in predictions made by machine learning models. \\

In all, through the embedding of physical models within network architecture, physical principles are enforced, leading to theory-adherent communication through the architecture itself. However, as with all learning algorithms, there exists a trade-off between the detail in which the model is designed to interpret, and computational demands of the model. In addition to the domain knowledge required, integrating physical principles within deep learning models increases their complexity. Depending on the implementation, physics-informed architectures may require more computational resources than conventional deep learning models, which could be a limiting factor in some applications in which computational speed is a requirement. \cite{viana2021estimating} has also noted this limitation in their study, wherein the complexity of the physical models embedded may prove unwieldy. By extension, an avenue of potential further research may be the adaptation of said complex models through guided simplifications or reduced-order models. The introduction of inductive biases through this format may also restrict the learning model, as it imposes strong assumptions on the data and learning process. While rigid constraints imposed by biases may be able to enhance efficiency through explicit guidance to the model, they may also serve to limit the model's flexibility to capture the underlying complexity of the data and the ability to generalize. Thus, the suitability design of the architecture with respect to its application must be carefully evaluated and tailored to ensure the efficacy of the algorithm. 

\subsection{Outlook}
Despite some limitations outlined above, the combination of physics-guided architecture design, in tandem with physics-informed regularisation techniques for optimization remains some of the cutting-edge and most sophisticated methodologies for integration of physical knowledge with data-driven techniques. Through a combination of hard and soft constraints, researchers have been able to tailor current machine learning algorithms to suit the need of several real-world condition monitoring applications. Current studies have already demonstrated great promise with regard to evaluation metrics such as accuracy, reliance on data, and robustness to noise and or incomplete data. With ongoing advancements in computational power, researchers can tackle more complex and realistic physical problems. The increased computational resources enable the exploration of larger and more comprehensive datasets, facilitating the discovery of intricate relationships and patterns that might have otherwise remained hidden. Additionally, higher computational capabilities allow for more sophisticated modeling techniques, enabling the consideration of complex physical phenomena and nonlinear dynamics that were previously challenging to capture accurately.

\subsection{Limitations}
A limitation of the findings in this survey was with regard to the sample size of literary works examined. Although the paradigm of PIML has been rapidly expanding since its inception, instances of literature implementing the PIML for applications within condition monitoring systems remain relatively low in comparison to other areas of development. Trends and literature outlined by this survey may be skewed towards authors or methodologies, and may not accurately capture the underlying trend of the technology, with respect to condition monitoring applications.

\section{Concluding Remarks} \label{Conclusion}
Physics-informed machine learning (PIML) methods offer a promising avenue for improving predictive modeling in physical systems, whereby the underlying physics-based constraints can be leveraged to further enhance conventional data-driven methods. By integrating the governing laws of physics into the learning algorithm, PIML is capable of effectively determining a non-naive, physically consistent representation of the system, thereby enabling accurate predictions and extrapolations beyond the training data. Furthermore, PIML methods facilitate data-efficient learning by guiding the learning algorithm to prioritize regions of interest and reduce the need for large training datasets. The incorporation of physics also enhances generalization capability, as the models can naturally handle extrapolation and capture the behavior under different conditions or perturbations. This work serves to provide an overview of such methods, with a focus on the methodology by which physical knowledge is integrated into conventional machine naive learning frameworks to formulate predictive models with a higher level of understanding and sophistication in relation to the underlying physical principles of the system. A total of 105 literary works have been sampled, with applications of PIML for condition monitoring in various fields of engineering. In the context of condition monitoring and fault detection, PIML methods leverage underlying known physical principles and domain knowledge to develop models capable of accurately predicting system behavior, detecting anomalies, and assessing the health status of critical components. Through this incorporation, models are able to more effectively capture the complex interactions between various system variables, enabling the identification of incipient faults and abnormalities with high sensitivity and specificity. A detailed exploration of current methodologies for the integration of known physics with machine learning methods is provided in this context, which is classified into primary categories with respect to the methodology by which physical knowledge of the system is integrated. Furthermore, this survey provides a generalized overview of some of the most popular deep learning algorithms employed, with brief explanations regarding their workings, their inherent advantages as well as limitations. Leveraging the initial understanding provided, the work seeks to detail recent innovations in the incorporation of physical knowledge by various authors in their respective studies. In all, several avenues of research were identified, including physics-guided augmentation or feature space, data-driven correctional mechanism, physics-informed regularization, and finally, physic-guided design of deep learning architectures. An interpretation of the various strengths, weaknesses, and limitations of each avenue of research is provided, and recommendations are made regarding nurturing areas of research with respect to the integration of the PIML paradigm with applications to condition monitoring of assets.

\section*{CRediT authorship contribution statement}
\textbf{Y. Wu:} Conceptualization, Methodology, Investigation, Writing - Original Draft, Visualization\\ 

\textbf{B. Sicard:} Writing - Review and Editing

\textbf{S. A. Gadsden:} Methodology, Supervision, Funding Acquisition, Writing - Review and Editing

\section*{Declaration of Competing Interest}
The authors declare that they have no known competing financial interests or personal relationships that could have appeared to influence the work reported in this paper.

\section*{Acknowledgements}
The authors would like to give their sincerest thanks and appreciation to the postdoctoral fellows, research associates, graduate students, and undergraduate researchers in the Intelligent and Cognitive Engineering (ICE) Laboratory at McMaster University for their continued support and encouragement.

\bibliography{sample}

\end{singlespacing}
\end{document}